\documentclass{article}
\usepackage{verbatim}
\usepackage{amsmath}
\usepackage{bm}
\usepackage{caption}
\usepackage{subcaption}
\usepackage{enumerate}
\usepackage{float}
\usepackage{mathtools}
\usepackage[section]{placeins}
\usepackage{url}
\usepackage[ruled]{algorithm2e}
\usepackage{hyperref}
\usepackage[left=3cm,right=3cm,top=3cm,bottom=3cm]{geometry}
\usepackage{authblk}

\usepackage[utf8]{inputenc} 
\usepackage{graphicx}
\usepackage[T1]{fontenc}    
\usepackage{booktabs}      
\usepackage{amsfonts}       
\usepackage{nicefrac}       
\usepackage{microtype}

\newtheorem{theorem}{Theorem}
\newtheorem{remark}{Remark}

\DeclareMathOperator{\WF}{W-F}
\DeclareMathOperator{\BetaDist}{Beta}
\DeclareMathOperator{\PRF}{PRF}
\DeclareMathOperator{\WFBP}{WFBP}
\DeclareMathOperator{\BeP}{BeP}
\date{}

\begin{document}

\title{Poisson Random Fields for Dynamic Feature Models}

\author[1]{Valerio Perrone}
\author[1,2]{Paul A. Jenkins}
\author[1]{Dario Span\'{o}}
\author[3]{Yee Whye Teh}
\affil[1]{Department of Statistics, University of Warwick}
\affil[2]{Department of Computer Science, University of Warwick}
\affil[3]{Department of Statistics, University of Oxford}
\affil[ ]{\texttt {\{v.perrone,p.jenkins,d.spano\}@warwick.ac.uk}}
\affil[ ]{\texttt {y.w.teh@stats.ox.ac.uk}}

\maketitle

\begin{abstract}
We present the \emph{Wright-Fisher Indian buffet process} (WF-IBP), a probabilistic model for time-dependent data assumed to have been generated by an unknown number of latent features. This model is suitable as a prior in Bayesian nonparametric feature allocation models in which the features underlying the observed data exhibit a dependency structure over time. More specifically, we establish a new framework for generating dependent Indian buffet processes, where the Poisson random field model from population genetics is used as a way of constructing dependent beta processes. Inference in the model is complex, and we describe a sophisticated Markov Chain Monte Carlo algorithm for exact posterior simulation.  We apply our construction to develop a nonparametric focused topic model for collections of time-stamped text documents and test it on the full corpus of NIPS papers published from 1987 to 2015.
\end{abstract}

\section{Introduction}
The Indian buffet process (IBP) \cite{ibp} is a distribution for sampling binary matrices with any finite number of rows and an unbounded number of columns, such that rows are exchangeable while columns are independent. It is used as a prior in Bayesian nonparametric models where rows represent objects and columns represent an unbounded array of features. In many settings the prevalence of features exhibits some sort of dependency structure over time and modeling data via a set of independent IBPs may not be appropriate. There has been previous work dedicated to extending the IBP to dependent settings \cite{williamson,depbeta,philo,ddep}. In this paper we present a novel approach that achieves this by means of a particular time-evolving beta process, which has a number of desirable properties. 

For each discrete time $t=t_0 = 0 <\cdots<t_T$ at which the data is observed, denote by $Z_t$ the feature allocation matrix whose entries are binary random variables such that $Z_{ikt}=1$ if object $i$ possesses feature $k$ at time $t$ and $0$ otherwise. Denote by $X_k(t)$ the probability that $Z_{ikt}=1$, namely the probability that feature $k$ is active at time $t$, and by $X(t)$ the collection of these probabilities at time $t$. 
The idea is to define a prior over the stochastic process $\{X(t)\}_{t\geq 0}$ which governs its evolution in continuous time.  In particular, for each feature $k$, $X_k(t)$ evolves independently, while features are born and die over time.  This is a desirable property in several applications such as in topic modeling, in which a new topic may be discovered (birth) or may stop being relevant (death) at some point in time. Our model benefits from these properties while retaining a very simple prior where sample paths are continuous and Markovian. Finally, we show that our construction defines a time-dependent beta process from which the two-parameter generalization of the IBP is marginally recovered for every fixed time $t$ \cite{twopar}.

The stochastic process we use is a modification of the so-called Poisson Random Field (PRF), a model widely used in population genetics \cite{sawyer1992,hartl94,bustamante,bustamante2,williamson2005,boyko,guten,amei2010,amei2012}. In this setting, new features can arise over time and each of them evolves via an independent Wright-Fisher (W-F) diffusion. 
The PRF model describes the evolution of feature probabilities within the interval $[0,1]$, allows for flexible boundary behaviours and gives access to several off-the-shelf results from population genetics about quantities of interest, such as the expected lifetime of features or the expected time feature probabilities spend in a given subset of $[0,1]$ \cite{EWENS}.

We apply the WF-IBP to a topic modeling setting, where a set of time-stamped documents is described using a collection of latent topics whose probabilities evolve over time. The WF-IBP prior allows us to incorporate time dependency into the focused topic model construction described in \cite{ibpc}, where the IBP is used as a prior on the topic allocation matrix determining which topics underlie each observed document. As opposed to several existing approaches to topic modeling, which require specifying the total number of topics in the corpus in advance \cite{lda}, adopting a nonparametric approach saves expensive model selection procedures such as the one described in \cite{scientific}. This is also reasonable in view of the fact that the total number of topics in a corpus is expected to grow as new documents accrue. Most existing nonparametric approaches to topic modeling are not designed to capture the evolution of the popularity of topics over time and may thus not be suitable for corpora that span large time periods. On the other hand, existing nonparametric and time-dependent topic models are mostly based on the Hierarchical Dirichlet Process (HDP) \cite{hdp}, which implicitly assumes a coupling between the probability of topics and the proportion of words that topics explain within each document. This assumption is undesirable since rare topics may account for a large proportion of words in the few documents in which they appear. Our construction inherits from the static model presented in \cite{ibpc} the advantage of eliminating this coupling. Moreover, it keeps inference straightforward while using an unbounded number of topics and flexibly capturing the evolution of their popularity continuously over time. 

Section \ref{sec:PRF} introduces the PRF with its application to the modeling of feature probabilities over time and shows that the mean measure of its stationary distribution coincides with the distribution of the atom masses in the beta process.
Section \ref{sec:model} presents the full feature allocation model and Section \ref{sec:mcmcf} describes a novel MCMC algorithm for posterior simulation with the model. 
Section \ref{sec:gauss} combines the model with a linear-Gaussian likelihood model and evaluates it on a synthetic dataset. Finally, Section \ref{sec:topic} illustrates the application of the WF-IBP to topic modeling and presents results obtained on both synthetic data and on the real-world data set consisting of the full text of papers from the NIPS conferences between the years 1987 and 2015.

\section{The Poisson random field model} \label{sec:PRF}
\subsection{The Wright-Fisher model}
\label{sec:WF}
Our starting point is the Wright-Fisher (W-F) model from population genetics \cite{EWENS}, which we briefly summarize here.
Consider a finite population of organisms of size $G$ such that $i \in \{0,1,\dots,G\}$ individuals have the mutant version of a gene at generation $k$, while the rest has the non-mutant variant. Assume that each individual produces an infinite number of gametes such that the gametes yielded by a non-mutant become mutant with probability $\mu_G$ and, conversely, those yielded by a mutant become non-mutant with probability $\beta_G$.
Finally, assume that the next generation of $G$ individuals is formed by simple random sampling from this infinite pool of gametes. 
The evolution of the number $Y^G(k)$ of mutant genes at time $k$ is described by a Markov chain on the discrete space $\{0,\dots,G\}$. The transition probability $p_{ij}$ of switching from $i$ mutants (at time $k$) to $j$ mutants (at time $k+1$) is given by the following binomial sampling formula:
\begin{align*}
p_{ij}&=  \binom {G}{j}  (\Psi_i)^j (1-\Psi_i)^{G-j}, \\ \Psi_i  &= \frac{i(1-\beta_G)+(G-i)\mu_G}{G}.
\end{align*}
Assume the initial state is $Y^G(0)=y_0$ and
denote the resulting Markov chain by $Y^G = (Y^G(k))_{k=1,2,\ldots} \sim \WF^G(\mu_G, \beta_G)$. Notice that, if $\mu_G=0$ and/or $\beta_G=0$, the states $0$ and/or $G$ are absorbing states that respectively correspond to the extinction and fixation of the mutation. 

A continuous-time diffusion limit of the W-F model can be obtained, by
rescaling time as $t=k/G$, and taking $G \to \infty$.  The Markov chain $Y^G(\lfloor Gt \rfloor) /G$ converges to a diffusion process on $[0,1]$ \cite{opac1986,sawyer1992} which obeys the one-dimensional stochastic differential equation
\begin{align*}
dX(t)= & \gamma(X(t))dt+ \sigma(X(t)) dB(t), 
\end{align*}
where
\begin{align}
\gamma(x) &= \frac{1}{2} [\mu (1-x) - \beta x], \label{drift} \\
\sigma(x) &= \sqrt{x(1-x)},  \label{diffusion}
\end{align}
with some initial state $X(0)=x_0$, over the time interval $t \in [0,T]$, with rescaled parameters $\mu = \lim_{G \to \infty}2G\mu_G$, $\beta = \lim_{G \to \infty}2G\beta_{G}$, and with $B(t)$ denoting a standard Brownian motion. The terms $\gamma(x)$ and $\sigma(x)$ are respectively referred to as the drift term and the diffusion term. Denote the diffusion process as $X \sim$ $\WF(\mu, \beta)$.

Notice that when $x(t)\to 0$ (respectively $x(t)\to 1$), then the diffusion term tends to $0$ while the drift term tends to $\frac{\mu}{2}$ (respectively $-\frac{\beta}{2}$), preventing absorption at 0 or 1 provided that $\mu>0$ (respectively $\beta>0$). Otherwise, 0 is an absorbing extinction state (respectively, 1 is an absorbing fixation state).  Moreover, if both $\mu,\beta>0$ then the diffusion is ergodic and has a stationary distribution that is a $\BetaDist(\mu,\beta)$.  

As there exists no closed form expression for its transition function, simulating from the W-F diffusion requires non-trivial computational techniques. The method we used is outlined in \cite{dangerfield}, a stochastic Taylor scheme tailored to the W-F diffusion. A novel alternative approach that allows for exact simulation has very recently been proposed by \cite{dandp}.

\subsection{The Poisson random field} \label{sec:prf}

The W-F model describes the evolution of a gene at one particular site. The Poisson random field (PRF) generalizes it to modeling an infinite collection of sites, each of which evolves independently according to the W-F model. As before, we start with a model with a population of finite size $G$, before taking the diffusion limit as $G\rightarrow \infty$.  For a site $i$ at which some individuals carry the mutant gene, denote by $X_{i}(k)$ the fraction of mutants in generation $k$. Each site evolves independently according to the W-F$^G$($0, 0$) model.  Further suppose that at each generation $k$ a number of mutations $M \sim \text{Poisson}(\nu_G)$ arise in new sites with indices $j_1, j_2,\dots j_M$. $\nu_G$ will be referred to as the immigration parameter of the PRF. Assume that each of the new mutations occurs at a new site in a single individual, with initial frequency $X_{j_m}(k)=1/G$. Subsequently, each new process $X_{j_m}(k+1),X_{j_m}(k+2),\ldots$ evolves independently according the W-F$^G$($0, 0$) model as well. As with pre-existing mutant sites, each process eventually hits one of the boundaries $\{0,1\}$ and stays there (we say that the mutation is extinct/has been fixed).

Consider the limit $G \to \infty$, so that after the same rescaling $t = k/G$ of time as in Section \ref{sec:WF} each site evolves as an independent W-F diffusion $X_i \sim$ W-F($0, 0$).
We also assume that $\nu_G \to \alpha$ as $G\to\infty$. 
This means that in the diffusion time scale the immigration rate is $G\nu_G \to \infty$, which suggests that the number of sites with mutant genes should explode. However, the initial frequency of each diffusion is $1/G \to 0$ as $G\to\infty$, and moreover $0$ is an absorbing state. It can be shown \cite{sawyer1992,amei2010} that only $O(G^{-1})$ of the newborn processes are not almost immediately absorbed. Therefore, there is a balance between the infinite number of newborn mutations and the infinite number of them that goes extinct in the first few generations, in such a way that the net immigration rate is $O(G\nu_G \times G^{-1}) = O(\alpha)$, and hence the limiting stationary measure is nontrivial. Provided that we remove from the model all sites whose frequency hits either the boundary 1 or 0, \cite{sawyer1992} prove that the limiting distribution of the fractions of mutants in the interval $[0,1]$ is a Poisson random field with mean density
\begin{equation} \label{prf0}
\alpha x^{-1} dx.
\end{equation}
This means that at equilibrium the number of sites whose frequencies $X_{i}(t)$ are in any given interval $(a,b]$ is Poisson distributed with rate $\alpha\int_a^b x^{-1} dx$, and these are independent for nonoverlapping intervals. Integrating \eqref{prf0} over $[0,1]$ shows that the number of mutations in the population that has not been fixed or gone extinct is infinite. However, most mutations are present in a very small proportion of the population.

\section{Time-Varying Feature Allocation Model} \label{sec:model}

The derivation of the PRF in the previous section shows that the equilibrium distribution of the PRF is related to the one-parameter beta process \cite{ibp,BP}.  In this section we generalise the PRF so that its equilibrium distribution is related to the two-parameter beta process, and so that it is better adapted to applications in feature allocation modeling.  Specifically, we will identify mutant sites with features, and identify the proportion of the population having the mutant gene with the probability of the feature occurring in a data observation.  The PRF can be then used in a time-varying feature allocation model whereby features arise at some unknown time point, change their probability smoothly according to the $\WF(0,0)$ diffusion process and eventually die when their probability reaches zero.

\subsection{Connection to the Indian buffet and Beta processes} \label{connection}

Recall from the previous section that mutant sites whose frequency hits 1 are removed from the PRF model.  This means that features with high probability of occurrence can be removed from the model instantaneously, which does not make modeling sense.  Instead, one expects a feature probability to change smoothly, and to be only removed from the model once its probability of occurrence is small.  A simple solution to this conundrum is to prevent 1 from being an absorbing state by using  a $\WF(0,\beta)$ diffusion with $\beta>0$ instead.   This is a departure from \cite{sawyer1992}, due to the differing modeling requirements of genetics versus feature allocation modeling.   

We shall denote the modified stochastic process as $\PRF(\alpha,\beta)$.  The following theorem derives the equilibrium mean density of $\PRF(\alpha,\beta)$, with proof given in Appendix A:
\begin{theorem} \label{theo}
The equilibrium mean density of the $\PRF(\alpha, \beta)$ is
\begin{equation} \label{intPRF}
l(x) = \alpha x^{-1} (1-x)^{\beta-1} dx.
\end{equation}
\end{theorem}
In other words, the mean density of the $\PRF(\alpha,\beta)$ is the L\'evy measure of the two-parameter beta process \cite{twopar,BP}, with the immigration rate $\alpha$ identified with the mass parameter, and $\beta$ identified with the concentration parameter of the beta process.  We will assume that the initial distribution of $\PRF(\alpha,\beta)$ is its equilibrium distribution, that is, a Poisson random field with mean density \eqref{intPRF}, so that the marginal distribution of the PRF at any point in time is the same.

We will now make the connection more precise by specifying how a PRF can be used in a time-varying feature allocation model. Denote by $X_k(t)$ the probability of feature $k$ being active at time $t$ and define our PRF as the stochastic process $X:=\{X_k(t)\}$.
Assume that at a finite number of time points $t=t_0,\ldots,t_T$ there are $N_t$ objects whose observable properties depend on a potentially infinite number of latent features. Let $D_{it}$ be the observation associated with object $i=1,\ldots, N_t$ at time $t=t_0,\ldots,t_T$. Consider a set of random feature allocation matrices $Z_t$ such that entry $Z_{ikt}$ is equal to 1 if object $i$ at time $t$ possesses feature $k$, and 0 otherwise. Let $Z:=\{Z_{ikt}\}$.  Finally, let $\rho_k$ be some latent parameters of feature $k$ and $\rho=\{\rho_k\}$ be the set of all feature parameters.  Our complete model is given as follows.
\begin{align}
X &\sim \text{PRF}(\alpha,\beta), \nonumber \\
Z_{ikt} \mid X &\stackrel{ind}{\sim} \text{Bernoulli}(X_k(t)), \nonumber \\
\rho_k &\stackrel{iid}{\sim} H, \nonumber \\
D_{it} \mid \rho, Z_{it} & \stackrel{ind}{\sim} F(\{\rho_k : Z_{ikt}=1\}), \label{eq:model}
\end{align}
where $i=1,\dots,N_t$, $t=t_0,\dots,t_T$ and $k=1,2,\dots$, $H$ is the prior distribution for feature parameters, and where $F(\rho)$ is the observation model for an object with a set of features with parameters $\rho$.  

Since the feature probabilities $X$ have marginal density \eqref{intPRF}, at each time $t$ the feature allocation matrix $Z_t$ has marginal distribution given by the two-parameter Indian buffet process \cite{BP}.  Further, since $X$ varies over time, the complete model is a time-varying Indian buffet process feature allocation model.  The corresponding De Finetti measure would then be a time-varying beta process. More precisely, this is the measure-valued stochastic process $G=\{G(t)\}$ where
\begin{align*}
    G(t) = \sum_{k=1}^\infty X_k(t) \delta_{\rho_k},
\end{align*}
which has marginal distribution given by a beta process with parameters $\alpha$, $\beta$ and base distribution $H$.  
We denote the distribution of $G$ as $\WFBP(\alpha,\beta,H)$.  
We can also express the feature allocations using random measures as well. In particular, let 
$$
B_{it} = \sum_{k=1}^\infty Z_{ikt} \delta_{\rho_k}
$$
be a Bernoulli process $\BeP(G(t))$ with mean measure given by the beta process $G(t)$ at time $t$. An equivalent way to express our model \eqref{eq:model} using the introduced random measures is then
\begin{align}
G &\sim \WFBP(\alpha,\beta,H), \nonumber \\
B_{it} \mid G &\sim \BeP(G(t)), \nonumber \\
D_{it} \mid B_{it}  &\sim F(B_{it}), \nonumber
\end{align}
where $\WFBP$ denotes our time-varying beta process, and we have used $F(B)$ to denote the same observation model as before, but with $B$ being a random measure with an atom for each feature, and whose location is the corresponding feature parameter.  In the following, we will use the notation introduced for \eqref{eq:model} instead of in terms of beta and Bernoulli processes for simplicity.

\section{MCMC inference} \label{sec:mcmcf}
Given a set of observations $D$, a natural inference problem would be to recover the latent feature allocation matrices $Z:=\{Z_t\}_{t=t_0}^{t_T}$ responsible for generating the observed data, the underlying feature probabilities $X$ and their parameters $\rho$.

First observe that, in order to use the model for inference, it is necessary to augment the state space with the features that are not seen in the feature allocation matrices. Simulating the dynamics of the PRF, however, would require generating an infinite number of features, which is clearly unfeasible. One way to deal with this could be to resort to some sort of truncation, considering only features whose probability is greater than a given threshold and are likely to be seen in the data. We will rather choose this truncation level adaptively by introducing a collection of slice variables $\{ S_t \}_{t=1}^{T_t}$ and adopting conditional slice sampling \cite{walkerslice,slice}. This scheme, which will be detailed in Subsection \ref{sec:gibbsandslice}, has the advantage of making inference tractable without introducing approximations.

Partition the set of features into two subsets, one containing the features that have been seen at least once among times $t=t_0,\dots,t_T$, and the other containing the features that have never been seen for all $t=t_0,\dots,t_T$, so that $X = X_{\text{seen}} \cup X_{\text{unseen}}$. Since the unseen features cannot be identified individually based on the matrices $Z$, and as all features are conditionally independent given $Z$, we will consider seen and unseen features separately. As for the seen features, we describe a Particle Gibbs (PG) \cite{Andrieu} algorithm to perform Bayesian inference on the feature trajectories continuously over the interval $[t_0,t_T]$. As for the unseen features, we simulate them via a thinning scheme.

 The general MCMC inference scheme can be then summarized by the following updates, which need to be iterated until convergence.
\begin{itemize}
\item $Z \mid X,S,D$ via Gibbs sampling.
\item $S \mid Z,X$ via slice sampling.
\item $\rho \mid X,Z$ according to the likelihood model.
\item $X_{\text{seen}} \mid Z$ via Particle Gibbs.
\item $X_{\text{unseen}} \mid S$ via thinning.
\end{itemize} 

We will present each of these steps in Subsections \ref{sec:gibbsandslice} and \ref{sec:particlegibbs}, together with a simpler inference scheme in Subsection \ref{finitemodel} in which the number of features is truncated. Before that, we describe how to simulate the feature probabilities and the corresponding feature allocation matrices at a set of time points, laying the basis for the ideas that will be used during inference.

\subsection{Simulating \texorpdfstring{$X$}{}} \label{sec:seq}
\begin{figure*}[ht!]
\centering
\includegraphics[width=0.85\textwidth,trim= 4 17 4 4,clip]{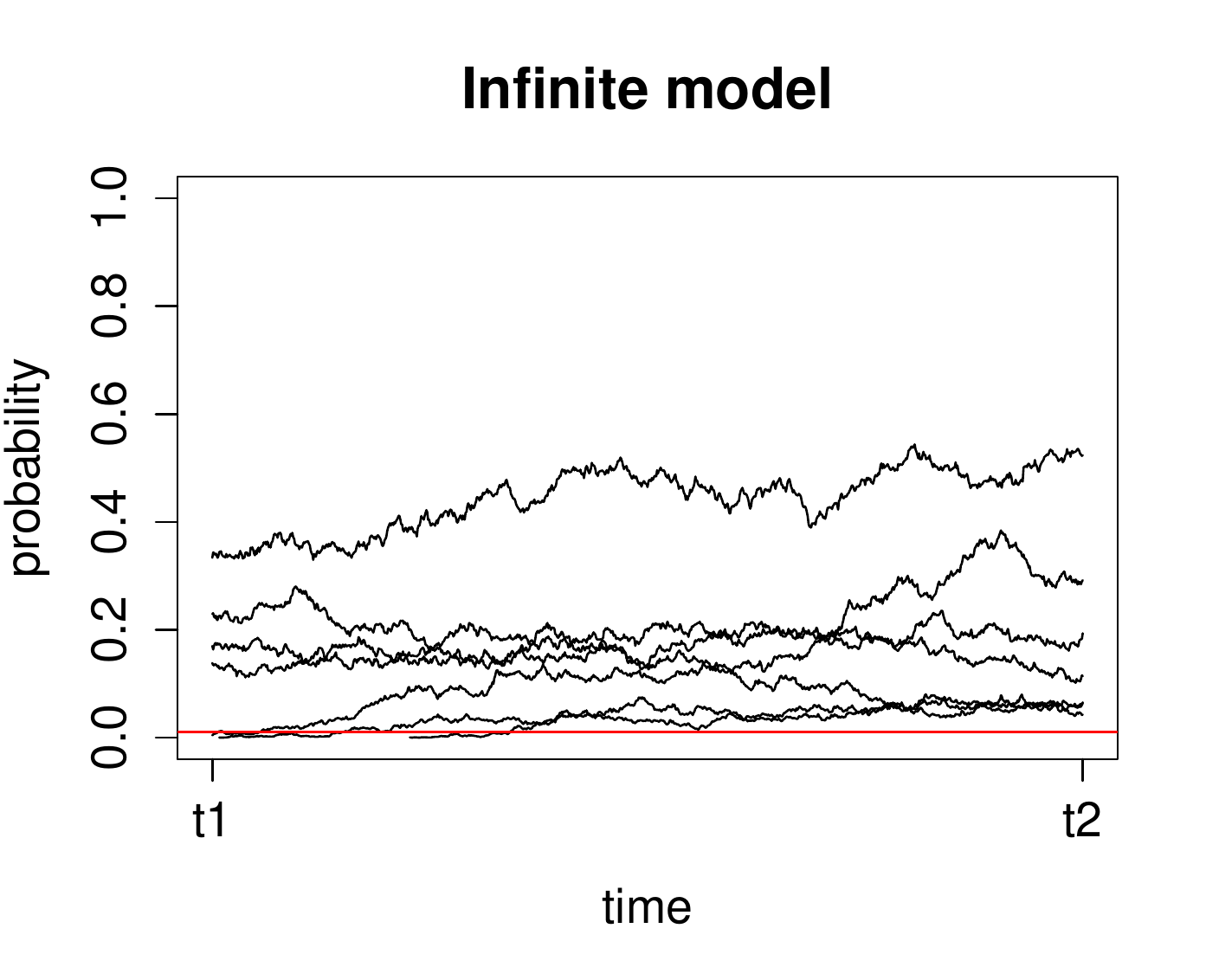} 
\caption {Visualization of the evolution of features via the PRF with parameters $\alpha=\beta=1$ from time $t_1$ to time $t_2$ with $t_2-t_1 = 0.1$. The red horizontal line corresponds to the truncation level $u=0.01$.}
\label{fig:seqcon}
\end{figure*}

Set a truncation level $u>0$ and consider the task of simulating features whose probability is above the threshold $u$ at two times $t_0$ and $t_1$. As we know how to simulate marginally from the beta process, we can first generate the feature probabilities above $u$ at time $t_0$ and let them evolve independently to time $t_1$. This yields features whose probability is greater than $u$ at time $t_0$, meaning that we are still missing those features whose probability is below $u$ at time $t_0$. To simulated these, we proceed as follows: we generate these features by drawing them from the beta process at time $t_1$ and propagate them backwards to time $t_0$. Note that this requires being able to simulate from the W-F diffusion backwards in time, which is not a problem as each $\WF(0,\beta)$ diffusion is time-reversible with respect to the speed density of the PRF \cite{griff}. Finally, in order not to double-count features, all features that in the reverse simulation have probability greater than $u$ at time $t_0$ have to be rejected. Overall, the only features that are not simulated are the ones whose probability falls below the truncation level $u$ both at time $t_0$ and at time $t_1$ (Figure \ref{fig:seqcon}). 

Now translate these ideas into the following sampling scheme. At time $t_0$, sample from a truncated version of the PRF, namely from a Poisson process on $[u,1)$ with rate measure $\alpha x^{-1}(1-x)^{\beta-1}dx$. This can be done, for instance, via an adaptive thinning scheme as described in \cite{ogata}. As the truncation level $u$ eliminates the point zero which has an infinite mass, this sampling procedure yields an almost surely finite number of samples. Denote by $\mathcal{K}$ the resulting set of feature indices and proceed as follows.
\begin{enumerate}
\item For all $k \in \mathcal{K}$, simulate $X_k(t) \mid \{ X_k(t_0) =x_k(t_0)\} \sim \text{WF}(0,\beta)$ for $t \in [t_0,t_1]$\footnote{We will use this notation to denote the following: simulate from a W-F diffusion with initial value $x_k(t_0)$ and set $X_k(t_1) = x_k(t_1)$, the value of the diffusion at time $t_1$.}.
\item At time $t_1$, sample the candidate newborn features $X(t_1)$ from the truncated PRF as above. Let $\mathcal{L}$ denote the resulting set of features.
\item For all $l \in \mathcal{L}$, simulate $X_l(t) \mid \{ X_l(t_1) =x_l(t_1)\} \sim \text{WF}(0,\beta)$ backwards for $t \in [t_1,t_0]$ and remove from $\mathcal{L}$ the indices in the set $\{l: x_l(t_0)\geq u\}$.
\item Generate $Z_{ikt} \mid \{X_k(t) = x_k(t) \}$ $\stackrel{iid}{\sim}$ Bernoulli$(x_k(t))$, for $t=t_0, t_1$, $\forall k\in \mathcal{K}\cup \mathcal{L}$ and $\forall i=1,\dots,N_t$. 
\end{enumerate}
As mentioned previously, the idea behind steps 2 and 3 is to compensate for the features with probability smaller than $u$ that were discarded when generating features at time $t_0$.

This construction generalizes to a set of time points $t=t_0,\dots,t_T$, observing that at a given time $t_{t^*} \in \{t_1,\dots,t_T \}$ step 3 needs to be modified by simulating the W-F diffusions backwards to time $t_0$ and removing from $\mathcal{L}$ the indices such that $\exists t \in \{t_0,\dots,t_{t^*-1}\}$ such that $x_l(t)\geq u$. As described above, for inference we will make use of the slice sampling technique, where for each time $t=t_0,\dots,t_T$ we have a different slice variable that is set to be the truncation level at that time. In this way inference on the feature allocation matrices $Z$ is kept exact, meaning that adopting a truncation level does not lead to any approximation error.

\subsection{Simulating \texorpdfstring{$Z$}{} and the underlying \texorpdfstring{$X$}{}} \label{z1z2}
Consider now the more complex task of simulating both the feature allocation matrices $Z$ and the features $X$ appearing in them. First note that, although the PRF describes the evolution of an infinite number of features, we can sample the feature allocation matrices $Z_{t_0}$ and $Z_{t_1}$ at two times $t_0$ and $t_1$ exactly, as a property of the IBP is that the number of observed features is almost surely finite \cite{ibp}. It is then possible to sample the features that are active in at least one object at times $t_0$ and $t_1$ and the corresponding allocation matrices $Z_{t_0}$ and $Z_{t_1}$ as follows. First, draw $Z_{t_0}$ from the IBP, and use its realisation to draw the posterior probabilities $X(t_0)$ of the features seen in $Z_{t_0}$. Indeed, observing a feature allocation matrix $Z_t$ updates the prior probability of features as in the posterior beta process \cite{BP}. Drawing each seen feature $k$ from the posterior beta process translates into drawing from a Beta($n_{kt}$, $\beta+N_t-n_{kt}$), where $n_{kt} := \sum_{i=1}^{N_t} z_{ikt}$ is the number of objects in which the considered feature is active at time $t$. Then, simulate from the W-F diffusion to propagate these features to time $t_1$ and generate $Z_{t_1}$ using these feature probabilities. We are now only missing the columns of $Z_{t_1}$ corresponding to the features that were seen at time $t_1$ but not at time $t_0$. To add those columns, first draw a candidate $Z^C_{t_1}$ from the IBP and the corresponding feature probabilities; then, simulate these candidate features backwards to time $t_0$ and accept them with probability $(1-X(t_0))^{N_{t_0}}$ to account for the fact that they were not seen at time $t_0$. The columns of $Z^C_{t_1}$ corresponding to the rejected features are deleted. Translating these ideas into an algorithm, consider the following steps.

\begin{enumerate}
\item Draw $Z_{t_0} \sim $ IBP$(\alpha,\beta)$ and index the resulting columns as $1,\dots,K_1$.
\item For $k=1,\dots,K_1$ draw the corresponding feature probabilities 
\begin{align*}
X_k(t_0) \mid \{ Z_{t_0}=z_{t_0} \} \sim \text{Beta} \left( n_{kt_0},\beta+N_{t_0}-n_{kt_0} \right).
\end{align*}
\item For $k=1,\dots,K_1$ simulate $X_k(t) \mid \{ X_k(t_0) =x_k(t_0)\} \sim \text{WF}(0,\beta)$ for $t \in [t_0,t_1]$ and set $X_k(t_1) = x_k(t_1)$.
\item Sample $Z^1_{t_1}$ by drawing each component $Z^1_{ikt_1} \mid \{ X(t_1)=x(t_1) \} \stackrel{iid}{\sim} \text{Bernoulli}(x_k(t_1))$, where $k=1,\dots,K_1$ and $i=1,\dots,N_{t_1}$.
\end{enumerate}
Then, to sample the features that are active only at time $t_1$, add the following steps.

\begin{enumerate}
\setcounter{enumi}{4}
\item Draw a candidate $Z^{\text{C}}_{t_1} \sim $ IBP$(\alpha,\beta)$ and index the resulting columns as $K_1+1,\dots,K_2$.
\item For $k=K_1+1,\dots,K_2$ draw the corresponding candidate feature probabilities 
\begin{align*}
X^{\text{C}}_k(t_1) \mid \{ Z^{\text{C}}_{t_1} = z^{\text{C}}_{t_1} \} \sim \text{Beta} \left(n_{kt_1},\beta+N_{t_1}-n_{kt_1} \right).
\end{align*}
\item For $k = K_1+1, \dots, K_2$, simulate $X_k(t) \mid \{ X_k(t_1) =x_k(t_1)\} \sim \text{WF}(0,\beta)$ backwards for $t \in [t_1,t_0]$ and set $X_k(t_0) = x_k(t_0)$.

\item Accept the candidate columns of $Z^C_{t_1}$ with probability $(1-x^C_k(t_0))^{N_{t_0}}$ and let $Z_{t_1}$ be the matrix obtained by the union of the columns of $Z^1_{t_1}$ with the accepted columns of $Z^C_{t_1}$.
\end{enumerate} 
Note that the rejection in the last step is a way to account for the fact that we are considering features that are active for the first time at time $t_1$. When considering a general set of time points $t_0,\dots,t_T$, it is necessary to account for the features that are active for the first time at each time $t_1,\dots,t_T$. In this more general case, features seen for the first time at time $t_{t^*}$ need to be accepted with probability $\prod_{t=t_0}^{t_{t^*-1}} (1-x^C_k(t))^{N_t}$, as they were not seen in any of the feature allocation matrices at the time points before $t^*$.

\subsection{Gibbs and slice sampling} \label{sec:gibbsandslice}
Turning to the core of the MCMC inference, first augment the parameter space with a set of slice variables. Given the feature allocation matrices $Z_t$ at times $t=t_0,\dots,t_T$, draw a slice variable $S_t \sim \text{Uniform}[0,x^*(t)]$ for each time $t$, where $x^*(t)$ is the minimum among the probabilities of the features seen at time $t$. In this way, when conditioning on the value $s_t$ of the slice variable, we have a truncation level $s_t$ and only need to sample the finite number of features whose probability is above this threshold \cite{slice}. In other words, for all $t=t_0,\dots,t_T$, we only need to update the columns of $Z_t$ whose corresponding feature probability $x_k(t)$ is greater than or equal to the slice variable $s_t$ (note that these include both seen and currently unseen features).

Denote by $Z_{-(ik)t}$ all the components of the matrix $Z_t$ excluding $Z_{ikt}$, and by $Z_{i-kt}$ all the components in row $i$ excluding $k$. Given data $D_{it}$ for every object $i=1,\dots,N_t$ and for every time $t=t_0,\dots,t_T$, we can define a Gibbs sampler for posterior inference over the matrices $Z$. To do this, we first need to derive for all $t$ the distribution of a given component $Z_{ikt}$ conditioning on the state of all other components $Z_{-(ik)t}$, on data $D_{it}$, on the slice variable $S_t$ and on the prior probability $X_k(t)$ of feature $k$. The probability of the entry $Z_{ikt}$ being active is
\begin{equation} \label{eq:Z1}
\begin{multlined} 
 P(Z_{ikt}=1 \mid Z_{-(ik)t},X_k(t),D_{it}) P(S_t  \mid  Z_{ikt}=1,Z_{-(ik)t}) \propto  \\   x_k(t) P(D_{it} \mid Z_{i-kt},Z_{ikt}=1) \frac{1}{x^*(t)}. 
\end{multlined}
\end{equation}

By the same token, the probability of the entry $Z_{ikt}$ being inactive is
\begin{equation} \label{eq:Z2}
\begin{multlined}
 P(Z_{ikt}=0 \mid Z_{-(ik)t},X_k(t),D_{it}) P(S_t  \mid  Z_{ikt}=0,Z_{-(ik)t}) \propto  \\  (1-x_k(t)) P(D_{it} \mid Z_{i-kt},Z_{ikt}=0) \frac{1}{x^*(t)}.
\end{multlined}
\end{equation}
Observe that the term $\frac{1}{x^*(t)}$ is not constant, as updating $Z_{ikt}$ for a currently unseen feature may modify the value of the minimum probability of the active features.
As the matrices $\{Z_t\}_{t=t_0}^{t_T}$ are conditionally independent given the feature probabilities $X$, equations \eqref{eq:Z1} and \eqref{eq:Z2} can be used to sample the matrices $\{Z_t\}_{t=t_0}^{t_T}$ independently given the respective feature probabilities at each time. 

Note that the likelihood $P(D_{it} \mid Z_t)$ needs to be specified according to the problem at hand. A typical choice, detailed in Section \ref{sec:gauss}, is the linear-Gaussian likelihood model, whose parameters can be easily integrated out \cite{ibp}. The update $\rho \mid Z,D$ over the feature parameters is also specific to the likelihood model and, as we will illustrate, can be easily derived in conjugate models such as the linear-Gaussian one.

\subsection{Particle Gibbs and thinning}
\label{sec:particlegibbs}
Assume that the feature allocation matrices $Z_t$ are given at the time points $t=t_0,\dots,t_T$ and we are interested in inferring the probabilities $X$ of the underlying features. This section gives the details of inference for each of the following subpartitions of seen and unseen features: features seen for the first time at a given time $t_j$ (for $j=0,\dots,T$), unseen features alive at time $t_0$ and unseen features born between any two consecutive times $t_{j}$ and $t_{j+1}$ (for $j=0,\dots,T-1$).

\subsubsection{Seen features}
As already mentioned, we apply PG to sample from the posterior trajectories of the seen features. Assume one has an initial reference trajectory $\xi^{r,k}_{t_0:t_T}:=(\xi^{r,k}_{t_0},\dots,\xi^{r,k}_{t_T})$ for each seen feature $k=1,\dots,K$. Draw a given number of particles from the posterior beta process at time $t_0$ and propagate them forward to time $t_1$ according to WF($0,\beta$). At time $t_1$, assign each of these features and the reference feature a weight given by the binomial likelihood of seeing that feature active in $n(t)$ objects out of $N(t)$ in $Z(t)$. Sample the weighted features with replacement and propagate the off-springs forward. This corresponds to using a bootstrap filter with multinomial resampling, but other choices to improve on the performance of the sampler can be made \cite{Andrieu}. Repeat this procedure up to time $t_T$, then weight the particles with the binomial likelihood given by $Z(t_T)$ and sample only one of them. Reject all the others and keep the trajectory that led to the sampled feature as the reference trajectory for the next iteration. Notice that the reference feature is kept intact throughout each iteration of the algorithm. This procedure is illustrated more precisely by Algorithm \ref{algo1}, which needs to be iterated independently for each seen feature to provide posterior samples from their trajectories. We drop the index $k$ to simplify the notation.

More generally, consider features that are seen for the first time at a given time $t_j$. As they cannot be identified individually based on any feature matrix $Z_{t_k}$ for $k<j$, these features need to be drawn from the posterior beta process at time $t_j$ and propagated both forward and backwards. The additional backward propagation requires adjusting Algorithm \ref{algo1} by replacing the steps before the while loop with Algorithm \ref{algo2}, where for simplicity we describe the particular case of features seen for the first time at time $t_1$. This description can be easily generalized to features that are seen for the first time at a generic time point $t \in \{ t_0,\dots,t_T \}$.

\begin{algorithm}
 \caption{PG: features seen at time $t_0$}
 \label{algo1}
\SetAlgoLined
\textbf{Input:} Reference trajectory $\xi^r_{t_0:t_T}; M$.\\
Set $\xi^M_{t_0}$ = $\xi^r_{t_0}$\;
Draw $\xi^i_{t_0} \sim$ Beta($n_{t_0}$, $\beta+N_{t_0}-n_{t_0}$) for $i = 1, \dots, M-1$\;
Simulate $\xi^i_{t} | \xi^i_{t_0} \sim$ WF($0,\beta$) for $t \in [t_0,t_1]$ for $i = 1, \dots, M-1$\;
Set $\xi^M_{t_1}$ = $\xi^r_{t_1}$\;
Compute $w^i_{t_1} = (\xi^i_{t_1})^{n_{t_1}}(1-\xi^i_{t_1})^{N_{t_1}-n_{t_1}}$ for $i = 1, \dots, M$\;
Sample $\bar{\xi}^{i}_{t_1}$ with $P(\bar{\xi}^{i}_{t_1}=\xi^{i}_{t_1}) \propto w^i_{t_1}$ for $i = 1, \dots, M-1$\; 
Set $\bar{\xi}^{M}_{t_1}$ = $\xi^r_{t_1}$\;
Simulate $\xi^i_{t} | \xi^i_{t_1} \sim$ WF($0,\beta$) for $t \in [t_1,t_2]$ for $i = 1, \dots, M-1$\;
Set $j \leftarrow 2$\;
\While{$t_j < t_T$}{
Set $\xi^M_{t_j}$ = $\xi^r_{t_j}$\;
Compute $w^i_{t_j} = (\xi^i_{t_j})^{n_{t_j}}(1-\xi^i_{t_j})^{N_{t_j}-n_{t_j}} w^i_{t_{j-1}}$ for $i = 1, \dots, M$\;
Sample $\bar{\xi}^{i}_{t_j}$ with $P(\bar{\xi}^{i}_{t_j}=\xi^{i}_{t_j}) \propto w^i_{t_j}$ for $i = 1, \dots, M-1$\; 
Set $\bar{\xi}^{M}_{t_j}$ = $\xi^r_{t_j}$\;

Simulate $\xi^i_{t} | \bar{\xi}^{i}_{t_j} \sim$ WF($0,\beta$) for $t \in [t_j,t_{j+1}]$ for $i = 1, \dots, M-1$\;

Set $j \leftarrow j+1$\;  
 }
 Compute $w^i_{t_T} = (\xi^i_{t_T})^{n_{t_T}}(1-\xi^i_{t_T})^{N_{t_T}-n_{t_T}} w^i_{t_{T-1}}$ for $i = 1, \dots, M$\;
 Sample $r_{new}$ with $P(r_{new} = i) \propto w^i_{t_T}$, where $i = 1, \dots, M$\;
 \textbf{Output:} New reference trajectory $\xi^{r_{new}}_{t_0:t_T}$. 
\end{algorithm}

\begin{algorithm}
\caption{PG: features seen for the first time at time $t_1$.}
\label{algo2}
\SetAlgoLined
\textbf{Input:} Reference trajectory $\xi^r_{t_0:t_T}; M$.\\
Set $\xi^M_{t_1}$ = $\xi^r_{t_1}$\;
Draw $\xi^i_{t_1} \sim$ Beta($n_{t_1}$, $\beta+N_{t_1}-n_{t_1}$) for $i = 1, \dots, M-1$\;
Set $\xi^M_{t_0}$ = $\xi^r_{t_0}$\;
Simulate $\xi^i_{t} | \xi^i_{t_1} \sim$ WF($0,\beta$) backwards for $t \in [t_1,t_0]$ for $i = 1, \dots, M-1$\;
Set $\xi^M_{t_2}$ = $\xi^r_{t_2}$\;
Simulate $\xi^i_{t} | \xi^i_{t_1} \sim$ WF($0,\beta$) for $t \in [t_1,t_2]$ for $i = 1, \dots, M-1$\;
Compute $w^i_{t_0} = (1-\xi^i_{t_0})^{N_{t_0}}$ for $i = 1, \dots, M$\;
Compute $w^i_{t_2} = (\xi^i_{t_2})^{n_{t_2}}(1-\xi^i_{t_2})^{N_{t_2}-n_{t_2}} w^i_{t_0}$ for $i = 1, \dots, M$\;
Draw $\bar{\xi}^{i}_{t_2}$ with $P(\bar{\xi}^{i}_{t_2}=\xi^{i}_{t_2}) \propto w^i_{t_2}$ for $i = 1, \dots, M-1$\; 
Set $\bar{\xi}^{M}_{t_2}$ = $\xi^r_{t_2}$\;
Simulate $\xi^i_{t} | \xi^i_{t_2} \sim$ WF($0,\beta$) for $t \in [t_2,t_3]$ for $i = 1, \dots, M-1$\;
Set $j \leftarrow 3$\;  
\end{algorithm}

\subsubsection{Unseen features}
 We now describe a thinning scheme to simulate the unseen features alive at time $t_0$. Denote the slice variable values at each time by $s_{t_0},\dots,s_{t_T}$ and note that sampling the set of unseen features from the truncated posterior beta process at time $t$ means drawing samples from a Poisson process on $[s_t,1)$ with rate measure $x^{-1}(1-x)^{\beta+N_t-1}dx$ \cite{BP}, which yields only a finite number of features whose probability is larger than $s_{t}$. First, draw the unseen features from the truncated posterior beta process at time $t_0$. Then, propagate them forward to time $t_1$ according to the W-F diffusion and accept them with probability $(1-x(t_1))^{N(t_1)}$, namely the binomial likelihood of not seeing them in any object at time $t_1$. Finally, iterate this propagation and rejection steps up to time $t_T$. The details of this thinning scheme are given in Algorithm \ref{algo3}.

\begin{algorithm}
 \caption{Thinning: unseen features alive at time $t_0$}
 \label{algo3}
\SetAlgoLined
Draw from a Poisson process on $[s_{t_0},1)$ with rate measure $\alpha x^{-1}(1-x)^{\beta+N_{t_0}-1}dx$ and denote by $\{ \xi^i_{t_0} \}_{i \in A}$ the resulting candidate particles\;
Set $j \leftarrow 1$\;
\While{$t_j < t_T$}{
Simulate $\xi^i_{t} | \xi^i_{t_{j}} \sim$ WF($0,\beta$) for $t \in [t_j,t_{j+1}]$ for all $i \in A$\;
Accept $\xi^i_{t_{j+1}}$ with probability $(1-\xi^i_{t_{j+1}})^{N_{t_{j+1}}}$ for all $i \in A$\;
Remove from $A$ the indices of the rejected particles\;
Set $j \leftarrow j+1$\;  
} 
\textbf{Output:} Trajectories $\{\xi^i_{t_0:t_T} \}_{i \in A}$ of the unseen features alive at time $t_0$ from the truncated PRF($\alpha,\beta$).\\
\end{algorithm}

\begin{algorithm}
 \caption{Thinning: unseen features born between time $t_0$ and $t_1$}
 \label{algo4}
\SetAlgoLined
Draw from a Poisson process on $[s_{t_1},1)$ with rate measure $\alpha x^{-1}(1-x)^{\beta+N_{t_1}-1}dx$\ and denote by $\{ \xi^i_{t_1} \}_{i \in A}$ the resulting candidate particles\;
Simulate $\xi^i_{t} | \xi^i_{t_{0}} \sim$ WF($0,\beta$) for $t \in [t_0,t_1]$ for all $i \in A$\;
\For{\text{all} $i \in A$}{
\eIf{$\xi^i_{t_0}>s_{t_0}$}{
Reject $\xi^i_{t_0}$\;
Set $A \leftarrow A \setminus \{i\}$\;
} {
Accept $\xi^i_{t_0}$ with probability $(1-\xi^i_{t_0})^{N_{t_0}}$\; \label{op0}
}
} 
Set $j \leftarrow 1$\;
 \While{$t_j < t_T$}{
Simulate $\xi^i_{t} | \xi^i_{t_{j}} \sim$ WF($0,\beta$) for $t \in [t_j,t_{j+1}]$ for all $i \in A$\;
Accept $\xi^i_{t_{j+1}}$ with probability $(1-\xi^i_{t_{j+1}})^{N_{t_{j+1}}}$ for all $i \in A$\;
 Remove from $A$ the indices of the rejected particles\;
 Set $j \leftarrow j+1$\;  
 } 
\textbf{Output:} Trajectories $\{\xi^i_{t_0:t_T} \}_{i \in A}$ of the unseen features born between time $t_0$ and $t_1$ from the truncated PRF($\alpha,\beta$).
\end{algorithm}

Notice that simulating the trajectories of the unseen features born between time $t_0$ and $t_1$ is equivalent to Algorithm \ref{algo3} from time $t_1$ onwards. The only difference is that these features, drawn at time $t_1$, need to be simulated backwards to time $t_0$ as well, hence the additional backward simulation followed by the rejection step in the for loop of Algorithm \ref{algo4}. If a feature that is simulated backwards from time $t_1$ to $t_0$ has probability 0 by time $t_0$, then it is a newborn feature and is accepted with probability 1. On the other hand, if its probability at time $t_0$ is between 0 and $s_{t_0}$, the particle belongs to the category of features that were alive and unseen at time $t_0$. Accepting them with probability $(1-x(t_0))^{N_{t_0}}$ compensates for the features that were below the truncation level $s_{t_0}$ in Algorithm \ref{algo3} and were thus not simulated at time $t_0$. In this way, only the features whose mass is below the slice variables $s_t$ at all times $t \in \{ t_0,\dots,t_T \}$ are not simulated. The exactness of the overall MCMC scheme is preserved by the fact that those features will be inactive in all the feature allocation matrices by the definition of slice variable.

Finally note that, for simplicity's sake, Algorithm \ref{algo4} describes only how to simulate the unseen features that were born between times $t_0$ and $t_1$, but the procedure needs to be generalized to account for the features born between any two consecutive time points $t_j$ and $t_{j+1}$, where $j=0, \dots,T-1$. In order to do this, it is sufficient to draw the candidate particles at every time $t_{j+1}$, with $j= 0,\dots,T-1$, propagate them backwards until time $t_0$ and thin them as follows: if their mass exceeds $s_t$ at any $t \in \{ t_0,\dots,t_T \}$, then they are rejected; otherwise, at each backward propagation to time $t \in \{ t_0,\dots,t_{T-1} \}$ they are accepted with probability $(1-x(t))^{N_t}$.

\subsection{Fixed-\texorpdfstring{$K$}{} truncation} \label{finitemodel}
We now provide an alternative inference scheme in which the number of features is truncated to a finite number $K$ and there is no need of introducing the slice sampler. We show that this simpler scheme marginally converges to the infinite one as $K \to \infty$.

\begin{figure*}[ht!]
\centering
\begin{subfigure}{.45\textwidth}
\centering
\includegraphics[height=8cm,width=7.1cm]{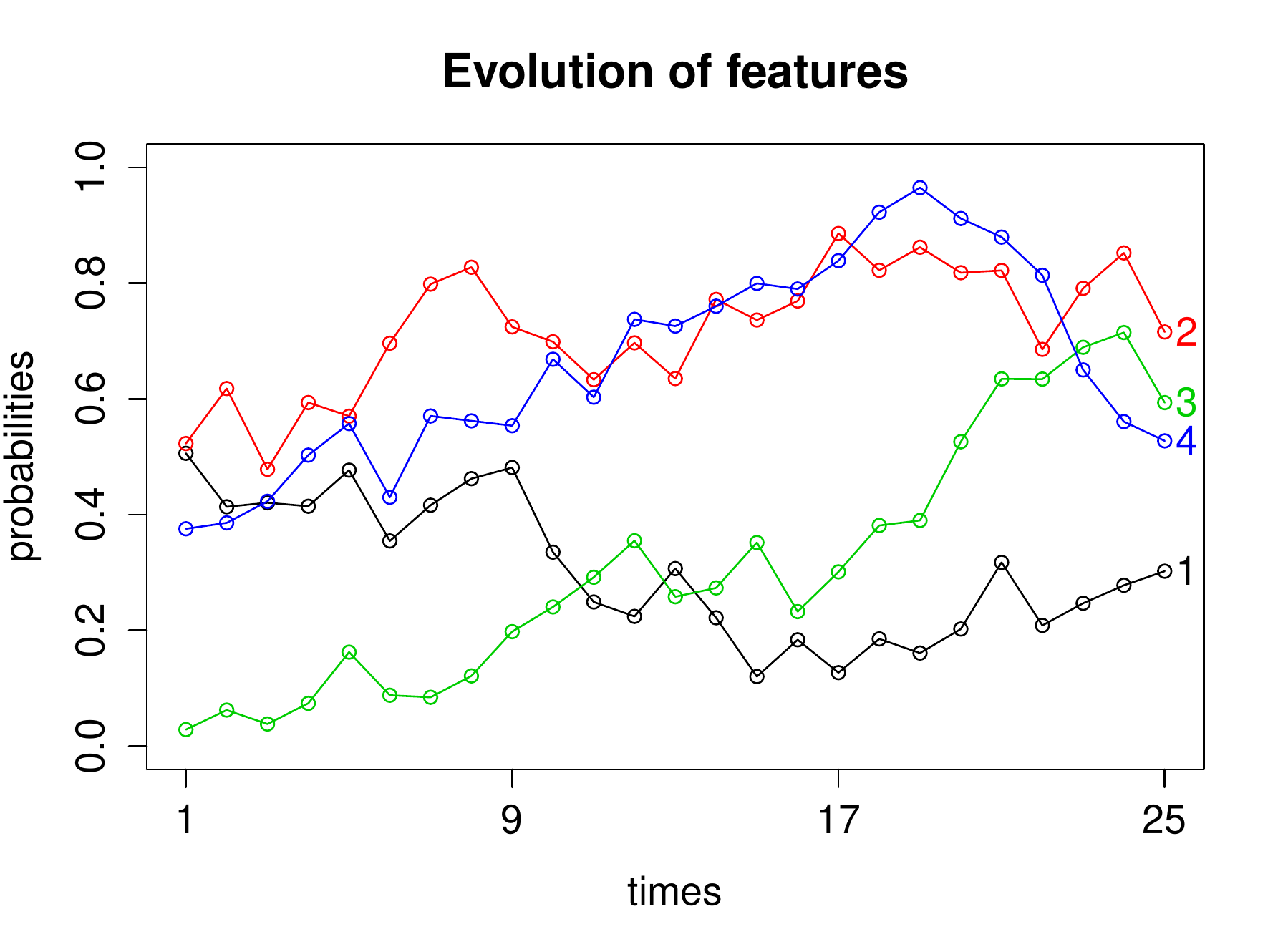}

\label{fig:4sub11}
\end{subfigure}
\begin{subfigure}{.45\textwidth}
\centering
\includegraphics[height=8cm,width=7.1cm]{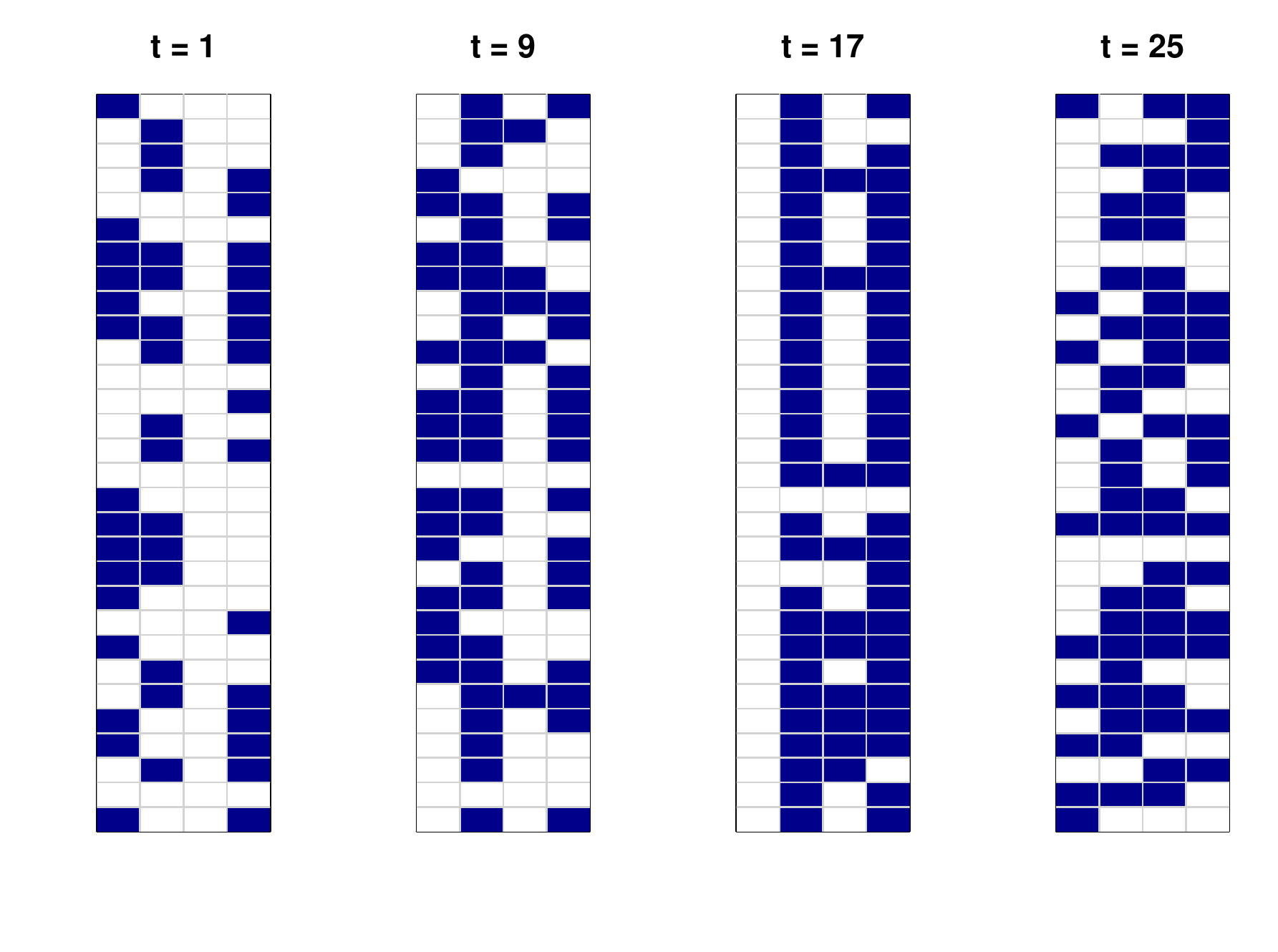}

\label{fig:4sub21}
\end{subfigure}
\caption{Left: underlying feature probabilities. Right: corresponding feature allocation matrices.}
\label{fig:Matrices}
\label{fig:depend}
\end{figure*}

Assume that the random feature allocation matrix $Z_t$ has a fixed number of features, say $K$. First, let $\alpha,\beta>0$ and consider the beta-binomial model
\begin{align*}
Z_{ikt} \mid \{X_k(t) = x_k(t) \} &\stackrel{iid}{\sim} \text{Bernoulli}(x_k(t)),  \\
X_k(t) &\stackrel{iid}{\sim} \text{Beta}\left( \frac{\alpha \beta}{K},\beta \right), 
\end{align*}
$\forall k=1,\dots,K, \forall i=1,\dots,N_t$. This coincides with the pre-limiting model of the two parameter IBP presented in \cite{twopar}. Then, for each feature, think of the Beta$\left(\frac{\alpha \beta}{K},\beta \right)$ distribution as the stationary distribution of a W-F diffusion with parameters $\frac{\alpha \beta}{K}> 0$ and $\beta> 0$. This suggests making the model time-dependent by letting each feature evolve, starting at stationarity, as an independent W-F diffusion with these parameters. Generate for all times $t=t_0,\dots,t_T$ the binary variables $z_{ikt}$ as 
\begin{align*}
Z_{ikt} \mid \{X_k(t) = x_k(t) \} &\stackrel{iid}{\sim} \text{Bernoulli}(x_k(t)),\\
X_k &\sim \text{WF} \left( \frac{\alpha \beta}{K},\beta \right),
\end{align*}
$\forall k=1,\dots,K, \forall i=1,\dots,N_t$. In this way, the closer two time points, the stronger the dependency between the probabilities of a given feature (Figure \ref{fig:depend}). Moreover, as we assume the W-F diffusion to start at stationarity, this construction coincides marginally with the beta-binomial model. Notice that the parameters of the W-F diffusion are positive, which implies that neither fixation nor absorption at the boundaries ever occurs and the number $K$ of features remains constant.

As there is now no need of introducing the slice variables, full inference on the feature allocation matrices $Z$ and on the feature probabilities $X$ can be performed by iterating the following updates.
\begin{itemize}
\item $Z  \mid  X, D$ via Gibbs sampling.
\item $\rho \mid X,Z$ according to the likelihood model.
\item $X  \mid  Z$ via Particle Gibbs.
\end{itemize} 
The conditional probabilities for the Gibbs sampling step coincide with the ones in the Gibbs sampling step outlined in Section \ref{sec:gibbsandslice}, with the difference that it is not necessary to introduce the slice variable as the number of features $K$ is finite. As a consequence, the conditional probabilities of the entries of $Z_t$ being active or inactive are given by equations \eqref{eq:Z1} and \eqref{eq:Z2} without the term $\frac{1}{x^*(t)}$.

As for the last step, in the fixed-$K$ truncation the PG step can be applied to each of the $K$ features. As the prior probability of each feature is a Beta $\left(\frac{\alpha \beta}{K},\beta \right)$ and the column-wise sums of $Z_{t_0}$ are realizations from binomial distributions, by conjugacy we have
\begin{align*}
X_k(t_0) \mid \{ Z_{t_0}=z_{t_0} \} \sim \text{Beta} \left( \frac{\alpha \beta}{K}+n_{kt_0},\beta+N_{t_0}-n_{kt_0} \right),
\end{align*}
where $k=1,\dots,K$. This posterior distribution can be therefore used to draw the whole set of features at time $t_0$ and the trajectories in the interval $[t_0,t_T]$ can be obtained via PG, weighting the features with $x_k(t)^{n_{kt}}(1-x_k(t))^{N_t-n_{kt}}$ at each time $t=t_0,\dots,t_T$.

\begin{figure*}
\centering
\begin{subfigure}{.45\textwidth}
\centering
\includegraphics[height=8cm,width=7.1cm]{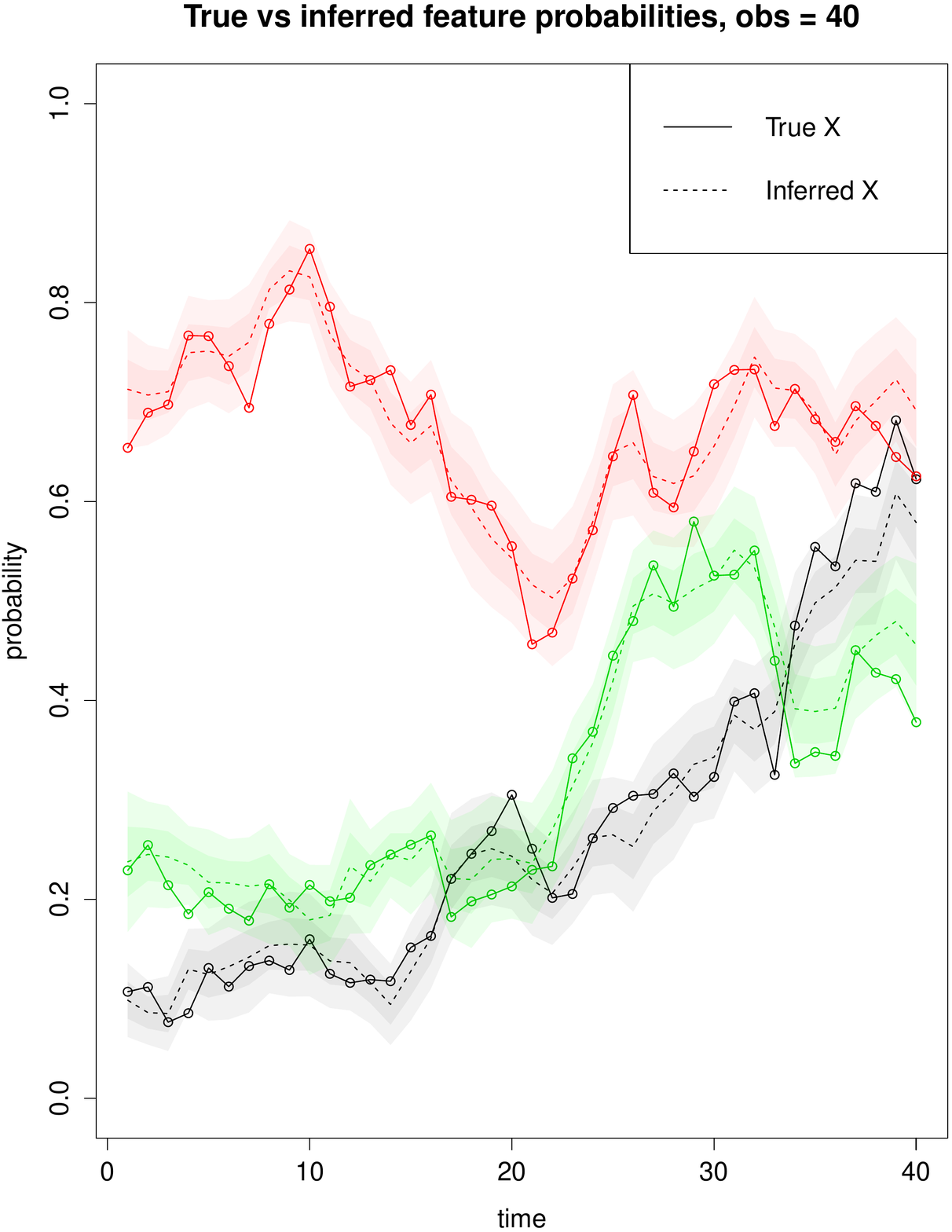}

\label{fig:sub111}
\end{subfigure}
\begin{subfigure}{.45\textwidth}
\centering
\includegraphics[height=8cm,width=7.1cm]{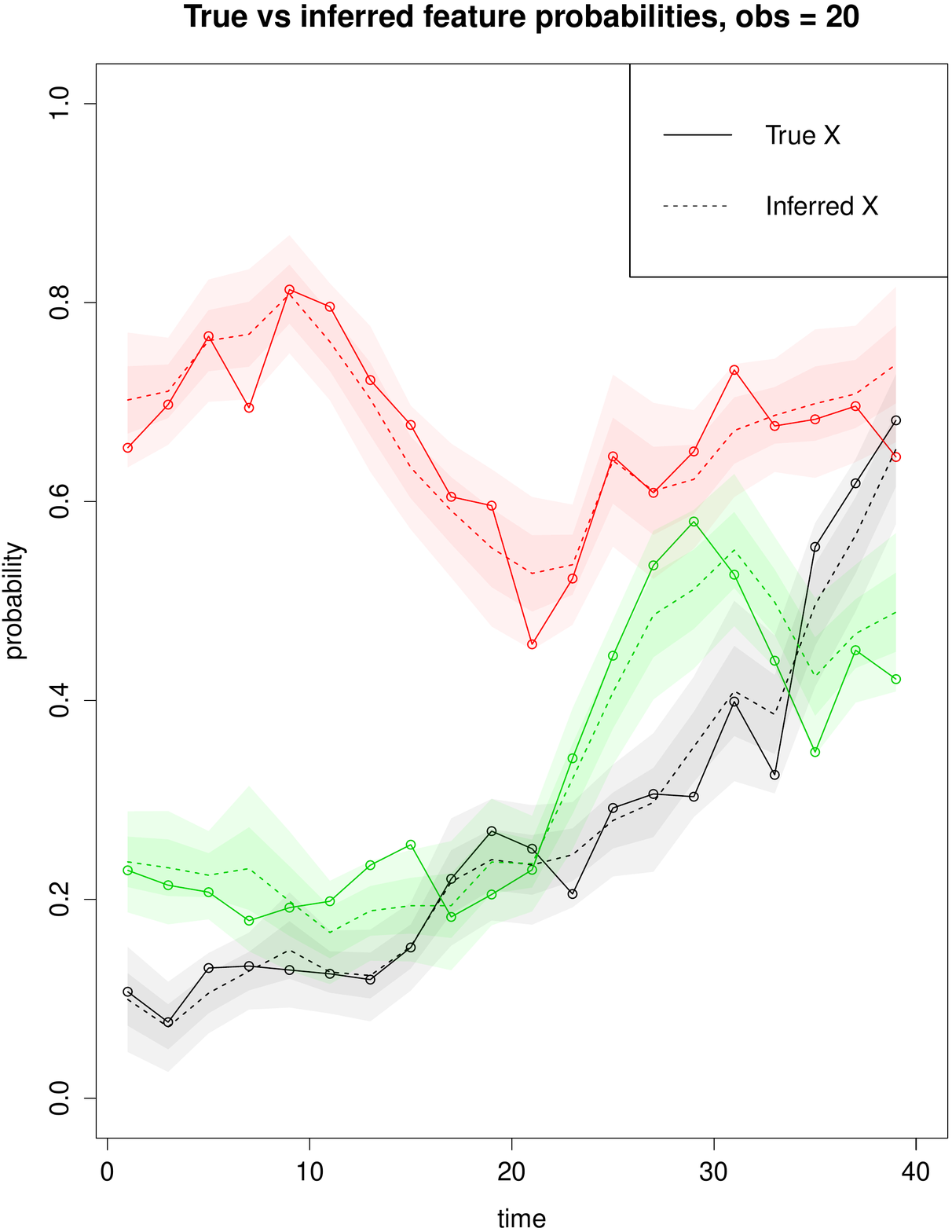}

\label{fig:sub2111}
\end{subfigure}

\centering
\begin{subfigure}{.45\textwidth}
\centering
\includegraphics[height=8cm,width=7.1cm]{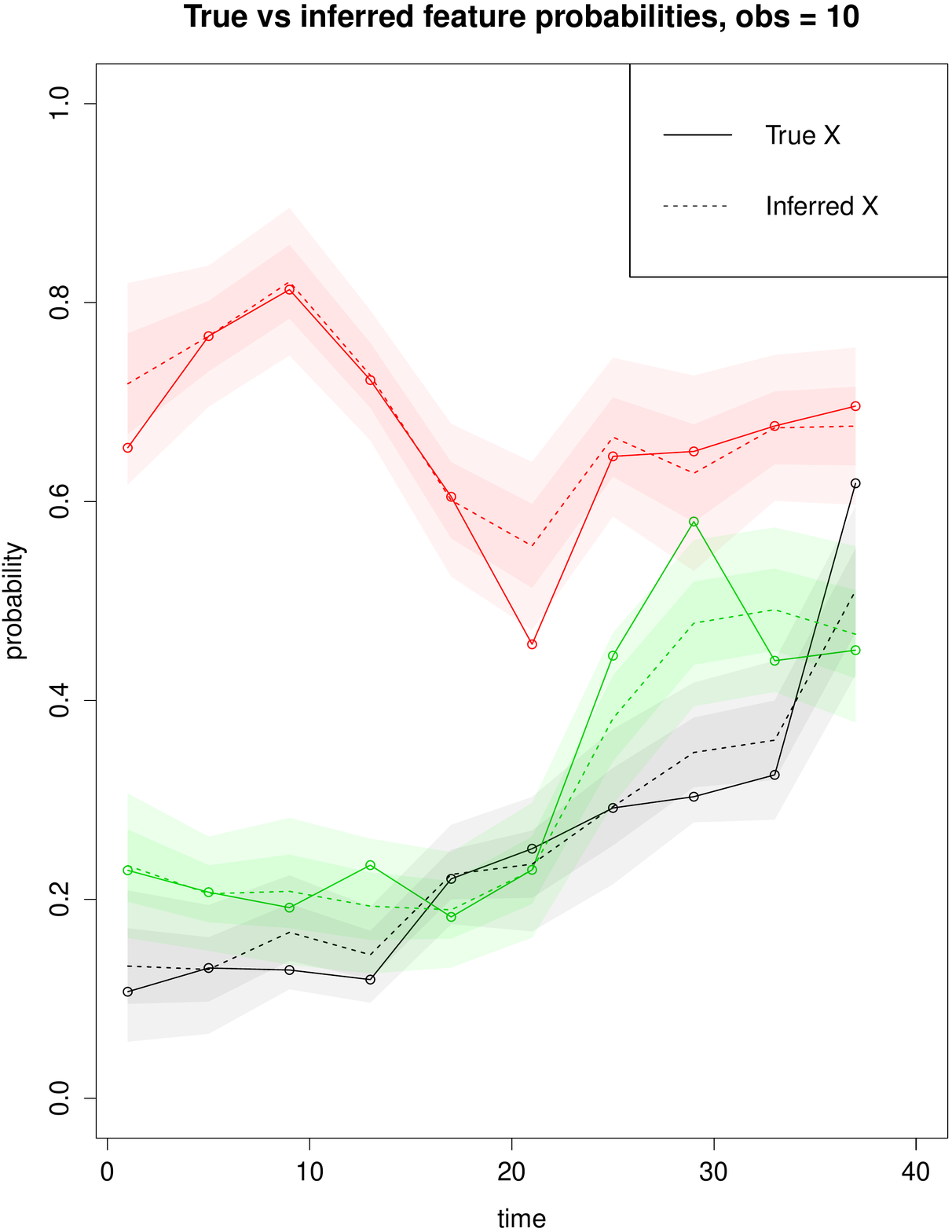}

\label{fig:sub3111}

\end{subfigure}
\centering
\begin{subfigure}{.45\textwidth}
\centering
\includegraphics[height=8cm,width=7.1cm]{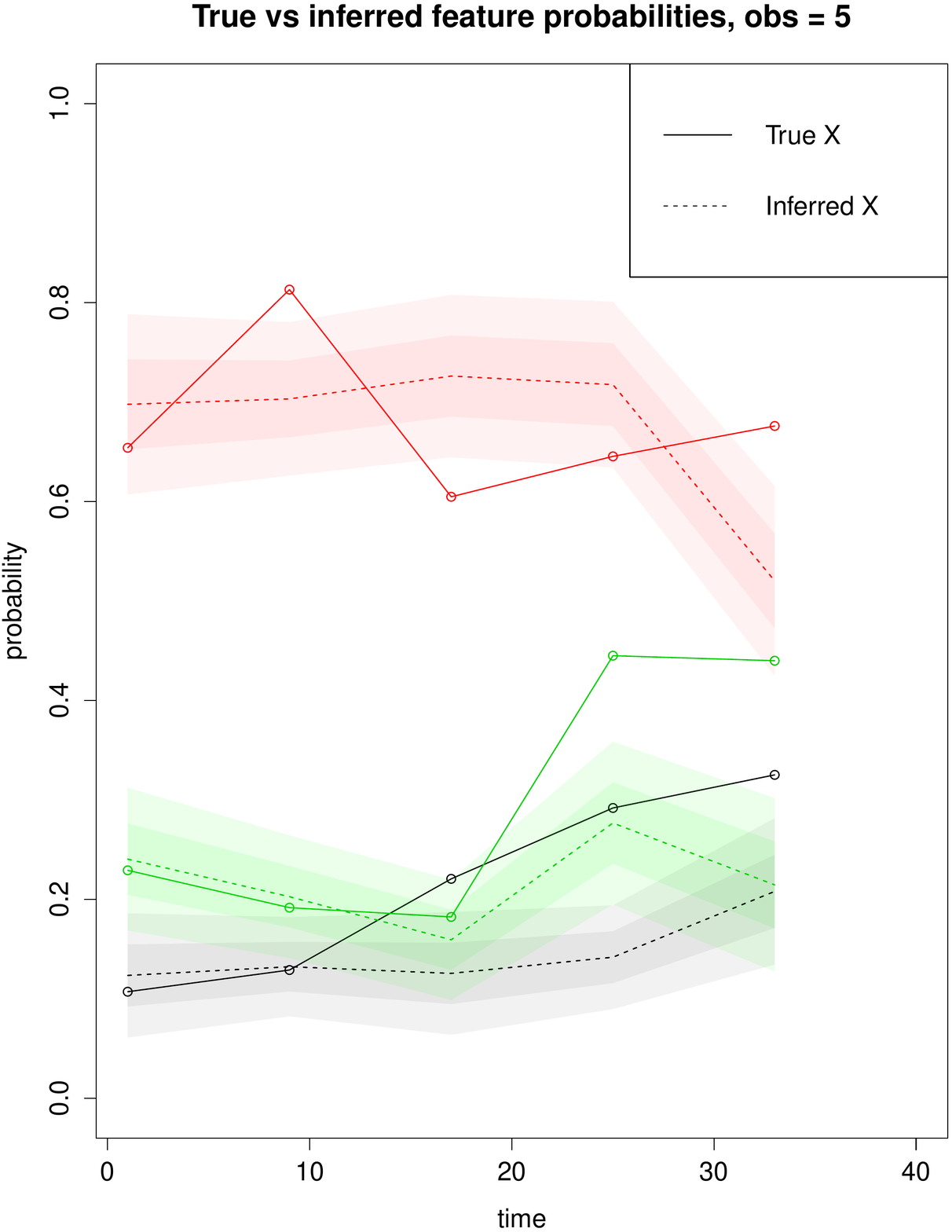}

\label{fig:sub4111}

\end{subfigure}

\caption{Comparison between the true feature probabilities and the posterior means obtained via Particle Gibbs varying the number of observations. The dark and light shaded areas respectively correspond to 1 and 2 standard deviations about the posterior mean.}
\label{fig:PG}
\end{figure*}

Figure \ref{fig:PG} illustrates the performance of Particle Gibbs in tracking the trajectories of feature probabilities over time. Three W-F diffusions representing the evolution of a corresponding number of features were used to generate synthetic observations in the form of feature allocation matrices. The algorithm was tested on a decreasing number of observations. The results show that given a sufficient number of observations, the posterior mean of the target distribution obtained by Particle Gibbs closely corresponds to the true values of the feature probabilities, which always fall within the interval given by two standard deviations about the posterior mean.

\subsubsection{Approximation for large \texorpdfstring{$K$}{}}
As already noted, the marginal distribution with the fixed-$K$ truncation corresponds to the beta-binomial model, which is the pre-limiting model of the two-parameter IBP. As a consequence, at any fixed time $t$ and as $K\to\infty$, the fixed-$K$ truncation converges to the two-parameter generalization of the IBP, which in turns coincides with the marginal distribution of the WF-IBP. 
An aspect of interest is then whether the whole dynamics of the fixed-$K$ truncation can be used as a finite approximation of the infinite model in such a way that, the larger $K$, the better the approximation. Two caveats need to be noted. First, only in the infinite model can features be born. For large $K$, however, the number of particles in the fixed-$K$ truncation whose mass is close to zero becomes so large that, with a sufficient amount of time, some of them gain enough mass to become `visible'. The behavior of these particles resembles the behavior of the newborn features of the infinite model. Second, as the fixed-$K$ truncation has an upwards drift equal to $\alpha \beta/K$, only the infinite model allows for features to be absorbed at 0. This discrepancy is however overcome by the fact that, when $K$ goes to infinity, 0 behaves like an absorbing boundary, in that features get trapped at probabilities close to 0. For these reasons, a comparison of the two models requires relabeling the particles in the finite model in such a way that, whenever a particle goes below a certain threshold $\epsilon \approx 0$, it is considered as a dead feature, while if its probability is below $\epsilon$ and later exceeds $\epsilon$ the particle is labeled as newborn. We choose this threshold to be $\epsilon = 1/K$, as then $\lim_{K \to \infty} \epsilon=0$. 

\begin{figure*}[ht!]
\centering
\includegraphics[height=7cm,width=10cm]{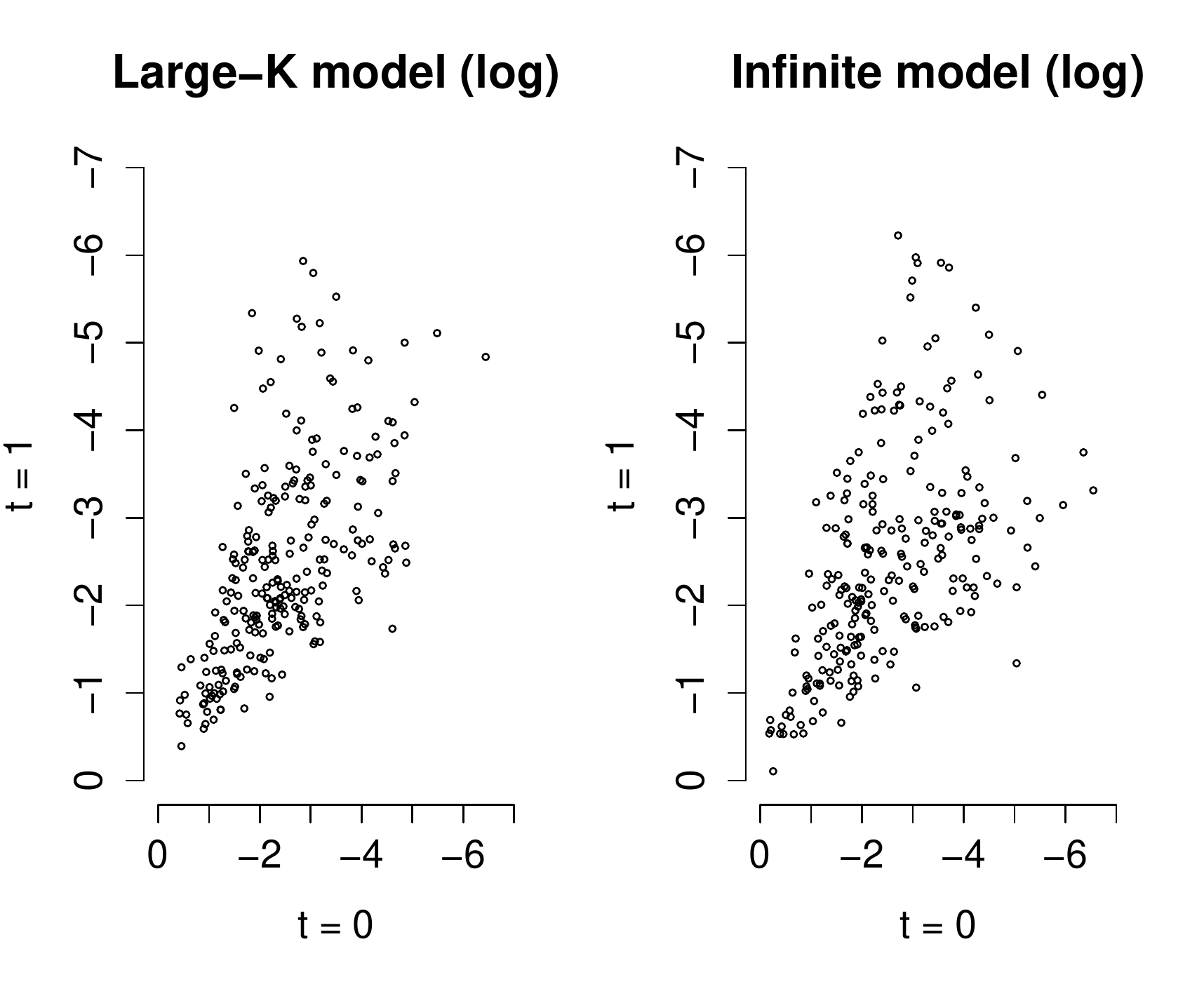}
\caption{Scatterplot of the log feature probabilities greater than $1/K$ at time $t_0=0$ and at time $t_1=1$ to compare the fixed-$K$ ($K=1000$) and infinite model.}
\label{fig:scatter}
\end{figure*}

Taking these caveats into account, we performed an empirical comparison of the fixed-$K$ truncation with the infinite model. Consider the joint distribution at two given time points $t_0=0$ and $t_1=1$ of the feature probabilities, first in the fixed-$K$ and then in the infinite model. Separately for each model, we took 1000 samples of the feature probabilities at time $0$ and at time $1$, excluding the ones below $1/K$. Figure $\ref{fig:scatter}$ shows the logarithm of these values for the two models, suggesting a remarkable similarity between the two underlying joint distributions. The validity of this comparison is supported by the maximum mean discrepancy (mmd) test \cite{kernel}, which does not reject the null hypothesis of the two joint distributions being the same. Although this suggests a strong similarity between the dynamics of the fixed-$K$ truncation and the infinite model, we leave a proof of the convergence of these joint distributions as $K\to\infty$ for future work.

\section{Application: linear-Gaussian likelihood model} \label{sec:gauss}
Turning to an application of the WF-IBP, the linear-Gaussian likelihood model is a typical choice to model the way latent feature allocation matrices are related to the observed data \cite{velez,ddep,ibp}. Assume that the collection of observations $O_t$ at time $t=t_0,\dots,t_T$ is in the form of an $N \times D$ matrix generated by the matrix product $O_t = Z_t \times A + \epsilon_t$. $Z_t$ is the $N \times K$ binary matrix of feature assignments at time $t$ and $A$ is a $K \times D$ factor matrix whose rows represent the feature parameters $\rho$. The matrix product is the way $Z_t$ determines which features are active in each observation, and $\epsilon_t$ is a $N \times D$ Gaussian noise matrix, whose entries are assumed to be distributed as independent $\mathcal{N}(0,\sigma^2_X)$. 

A typical inference problem is to infer both the feature allocation matrices $\{Z_t\}_{t=t_0}^{t_T}$ and the factor matrix $A$. In order to achieve this, we place on each element of A an independent prior $\mathcal{N}(0,\sigma^2_A)$ and on the hyper-parameter $\sigma^2_A$ an inverse-gamma prior $\Gamma ^{-1}(1,1)$. This choice of priors is convenient as it is easy to obtain the posterior distributions of $\sigma^2_A$ and $A$ (for the case $T=1$, see \cite{velez}).

For simplicity of notation, consider a fixed number of features $K$. Denote by $\bar{Z}$ the $TN \times K$ matrix obtained by concatenating the feature matrices $\{Z_t\}_{t=t_0}^{t_T}$ vertically, and by $\bar{O}$ the $TN \times D$ matrix obtained by combining the observations $\{O_t\}_{t=t_0}^{t_T}$ in the same way. The posterior of $A$ is matrix Gaussian with the following mean $\mu^A$ (a $K \times D$ matrix) and, for each column of $A$, the following covariance matrix $\Sigma^A $ (a $K \times K$ matrix).
\begin{align*}
\mu^A &= \left( \bar{Z}^T\bar{Z} + \frac{\sigma_X^2}{\sigma_A^2}I \right)^{-1} \bar{Z}^T \bar{O} \\
\Sigma^A &= \sigma_X^2 \left( \bar{Z}^T\bar{Z} + \frac{\sigma_X^2}{\sigma_A^2}I \right) ^{-1}.
\end{align*}
By conjugacy, the posterior distribution for $\sigma^2_A$ is still inverse gamma with updated parameters, namely
\begin{align*}
\sigma^2_A \sim \Gamma ^{-1} \left(1+ \frac{1}{2}KD,1+ \frac{1}{2}\sum_k \sum_d A_{kd}^2 \right).
\end{align*}

\begin{comment}
Starting with an arbitrary initialization $A=\bar{A}$ and $X=\bar{X}$, the updates that need to be iterated until convergence can be summarized as follows.

\begin{align*}
&\bar{A} \rightarrow \Sigma^A \\ & \dasheddownarrow  \text{ } \quad \quad  \downarrow \\
\bar{X} \dashrightarrow &Z \rightarrow \text{ } A \rightarrow \Sigma^A \\ & \Downarrow \quad \text{ }  \dasheddownarrow \quad \quad \text{   }  \downarrow  \\
& X \dashrightarrow Z \rightarrow A \rightarrow ...\\   & \quad \quad  \text{    } \Downarrow  \quad  \text{  } \downarrow   \quad  \quad \text{  } \downarrow \\ & \quad \quad \quad  \text{  }  ... \rightarrow  ...  \rightarrow  ...
\end{align*}

The dashed arrows denote the slice sampling update and the double arrows denote the Particle Gibbs update.
\end{comment}

\subsection{Simulation and results}
We tested the WF-IBP combined with a linear-Gaussian likelihood on a synthetic data set. Starting from the fixed-$K$ truncation, we generated $N=50$ observations at each of $40$ equally-spaced time points as in the linear-Gaussian model. The true factor matrix $A$ contained $K=3$ latent features in the form of binary vectors of length $D=30$. Their probability of being active was determined continuously over time by three independent $\WF(1,1)$ diffusions, simulated for $0.01$ diffusion time-units between every two consecutive time points. The resulting observations were corrupted by a large amount of noise ($\sigma_X=0.5$). 2000 iterations of the overall algorithm were performed, choosing a burn-in period of 200 iterations and setting the time-units and drift parameters of the W-F diffusion equal to the true ones in the PG update. As ground truth was available, we were able to test the ability of the algorithm to recover the true feature allocation matrices, the values of the latent features and their probabilities over time.

\begin{figure*}
\centering
\begin{subfigure}{.45\textwidth}
\centering
\includegraphics[height=8cm,width=6.6cm]{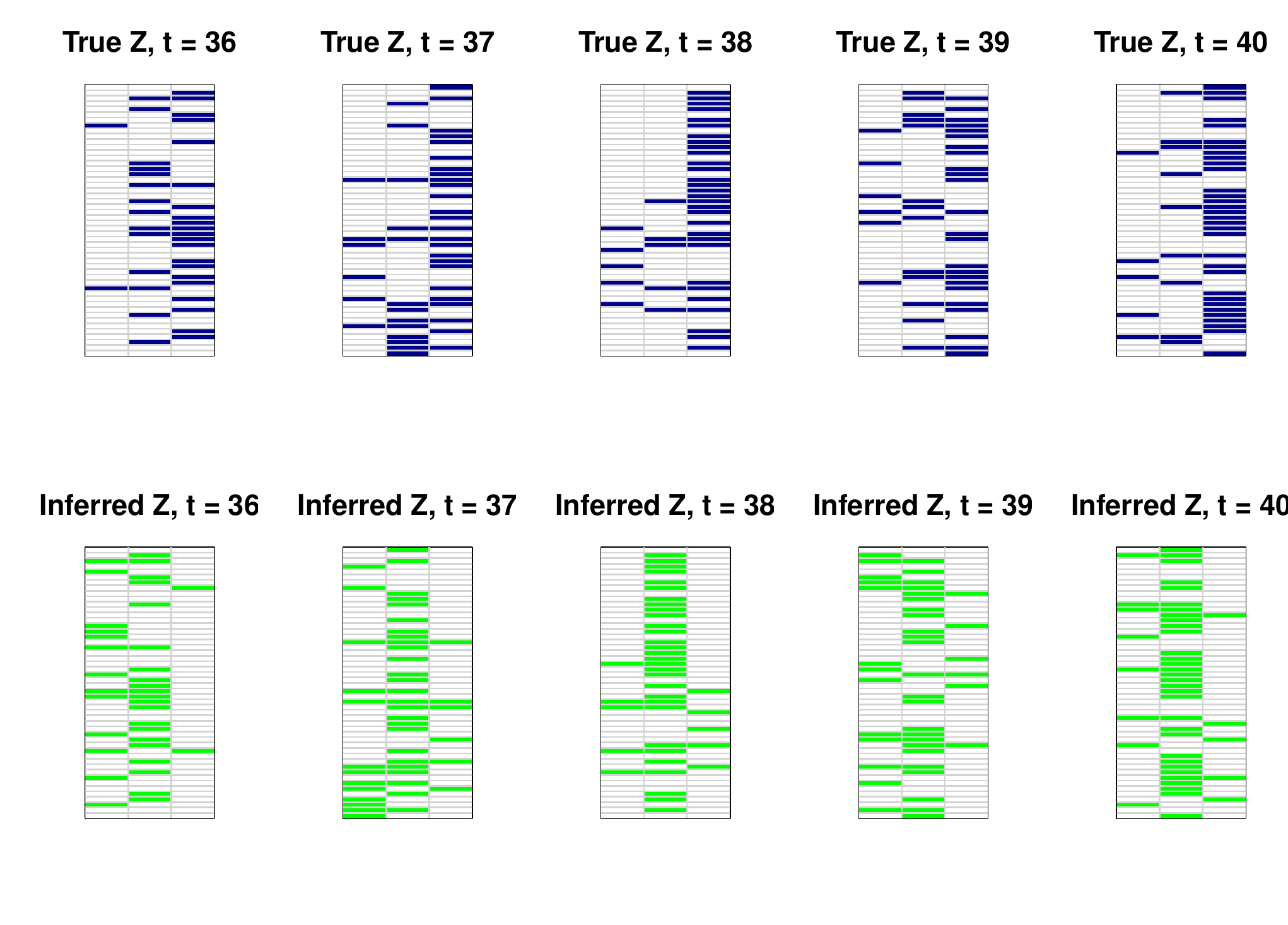}

\label{fig:sub11}
\end{subfigure}
\begin{subfigure}{.45\textwidth}
\centering
\includegraphics[height=8cm,width=6.6cm]{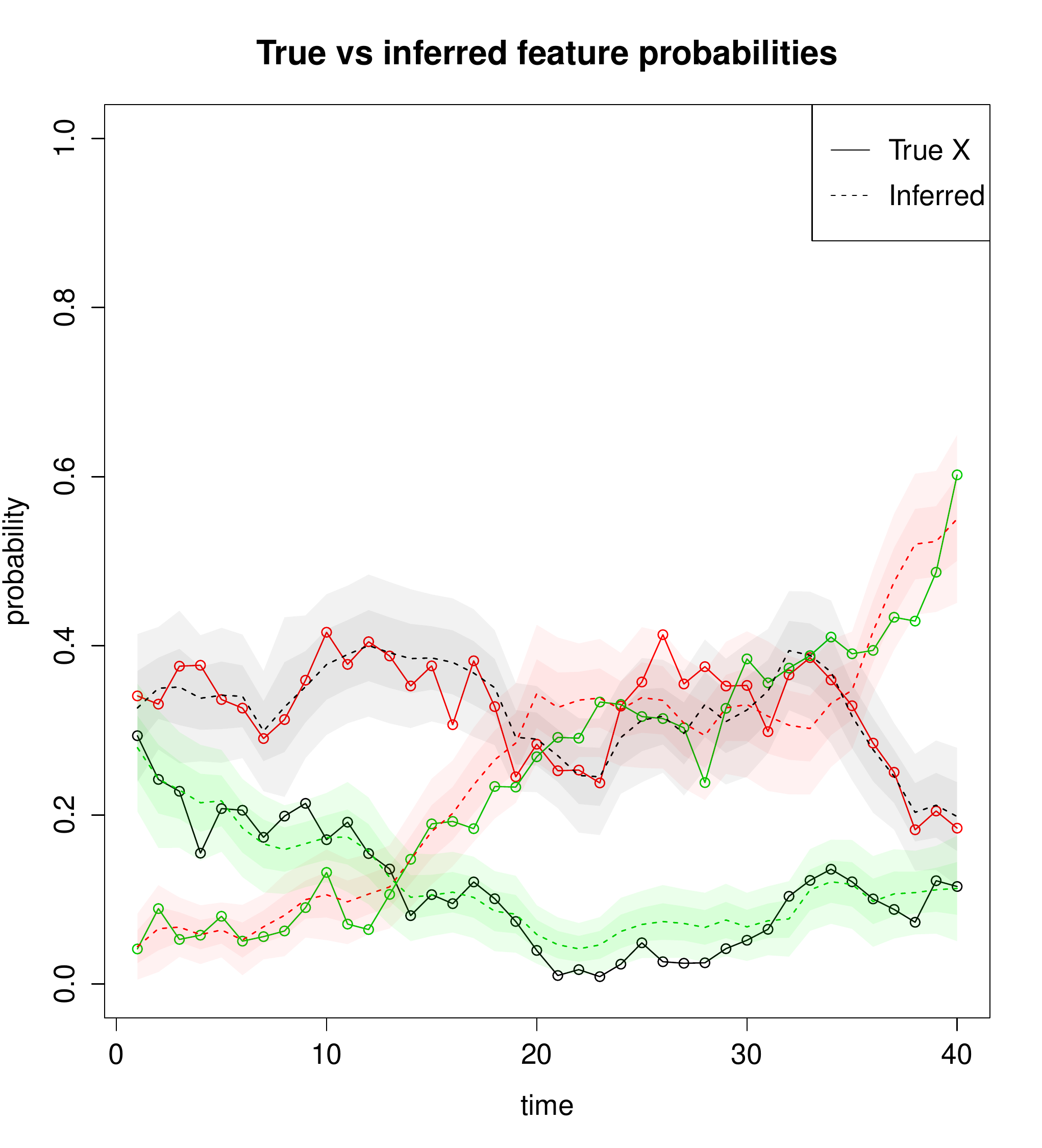}

\label{fig:sub21}
\end{subfigure}
\caption{Fixed-$K$ truncation. Left: Comparison of a subset of the true feature allocation matrices with the corresponding inferred ones. Rows represent objects and columns features, which can be active (blue/green) or inactive (white). Right: Comparison of true and inferred feature probabilities over time. The dark and the light shaded areas respectively indicate one and two standard deviations about the posterior mean.}
\label{fig:matrices}
\end{figure*}

Figure $\ref{fig:matrices}$-left compares the true underlying feature matrices at some example time points $t=36,\dots,40$ with the most frequently active features in the posterior mean matrices, where a feature is set to be active if that is the case in more than half of the samples of the Markov chain. Up to a fixed-across-times permutation of the columns, the resulting mean matrices closely match the true underlying feature matrices. Notice that the choice of the likelihood is such that the order of the columns of the feature matrices is not relevant as long as it is consistent with the order of the rows of the factor matrix $A$. This property of the model is reasonable in all settings in which there is no labeling of the features. 

Figure $\ref{fig:matrices}$-right compares the trajectories over time of the true feature probabilities with the inferred ones. The latter are never more than two standard deviations away from the former and thus the true feature trajectories are closely tracked. 

\begin{figure*}
\centering
\begin{subfigure}{.45\textwidth}
\centering
\includegraphics[height=7cm,width=6.5cm]{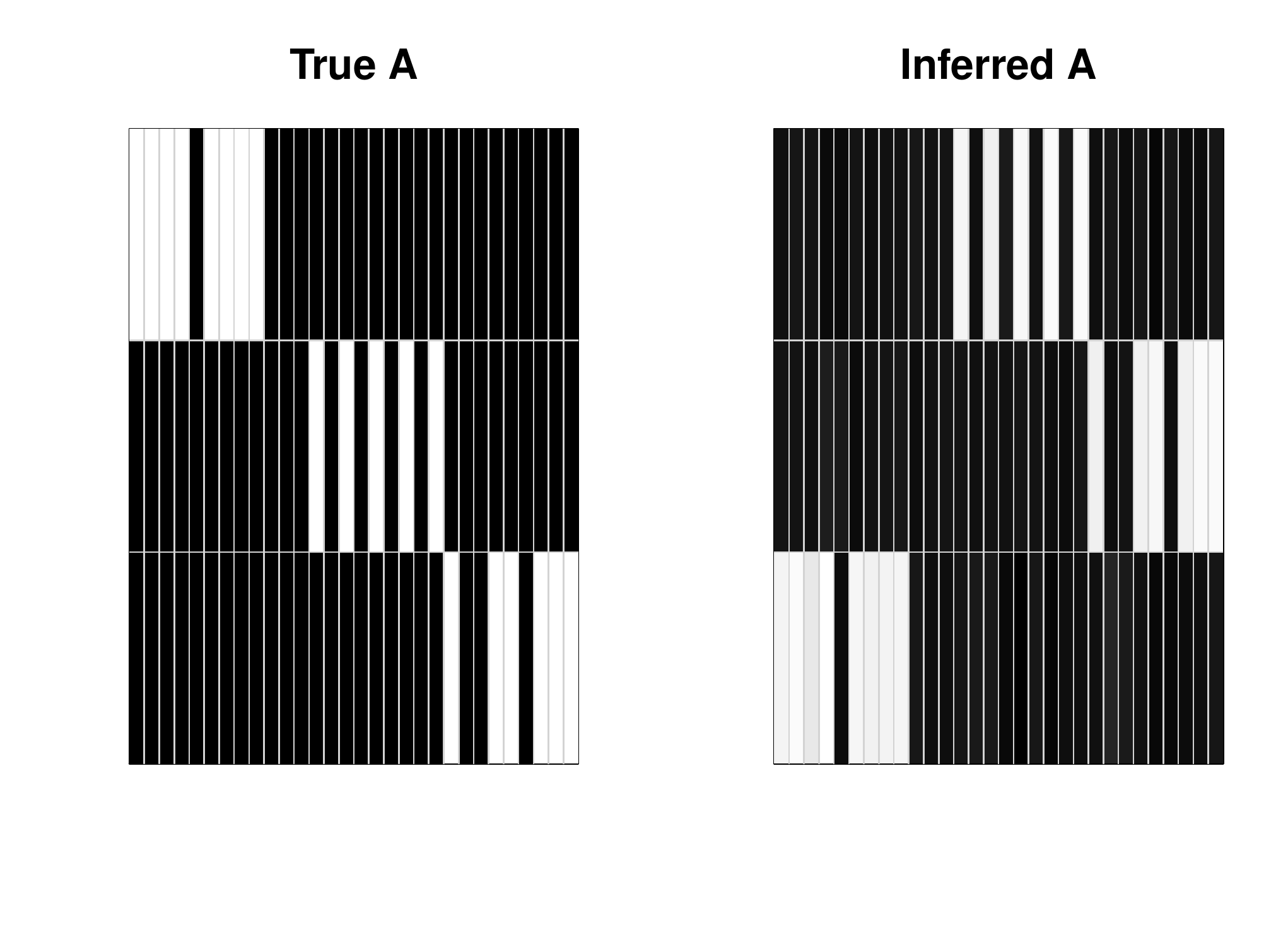}

\label{fig:1sub11}
\end{subfigure}
\begin{subfigure}{.45\textwidth}
\centering
\includegraphics[height=7cm,width=6.5cm]{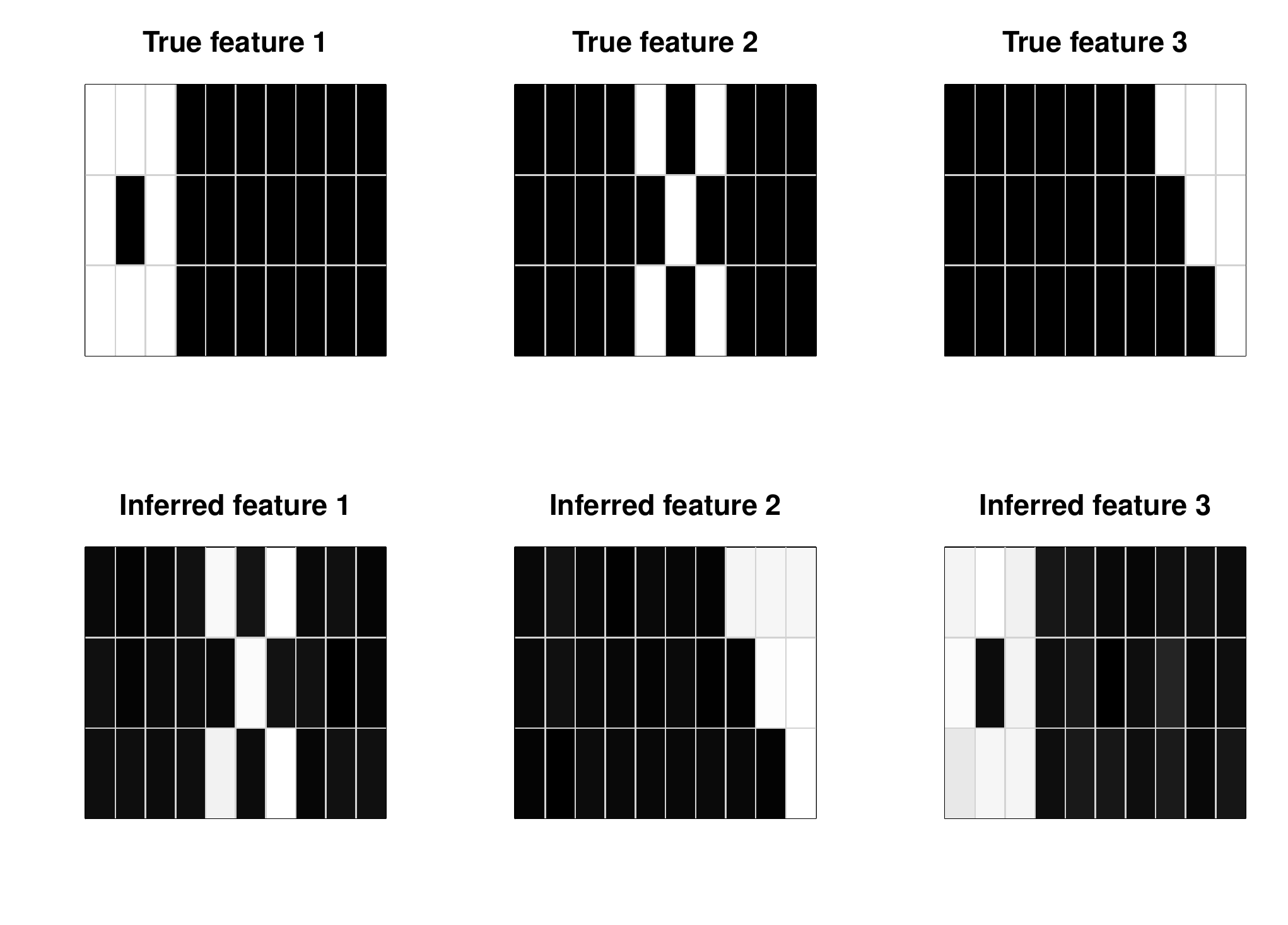}

\label{fig:sub213}
\end{subfigure}
\caption{Fixed-$K$ truncation. Black and white entries respectively represent the values 0 and 1, while the shades of grey represent the values between 0 and 1 stemming from the uncertainty in the inference. Left: Comparison between the true factor matrix $A$ and the inferred one $A$. Right: Visualization of true and inferred features. (Note that the order of the inferred features is consistent with the order of the inferred feature allocation matrices in Figure \ref{fig:matrices}-left.)}
\label{fig:factor}
\end{figure*}

Figure $\ref{fig:factor}$-left compares the true factor matrix A and the posterior mean matrix $\hat{A}$. The similarity between the true and inferred features can be assessed more easily in Figure $\ref{fig:factor}$-right, where a matrix is built out of each row. The results illustrate that the algorithm was able to recover accurately the hidden features underlying the noisy observations.

\begin{figure*}
\centering
\begin{subfigure}{.4\textwidth}
\centering
\includegraphics[height=9cm,width=6cm]{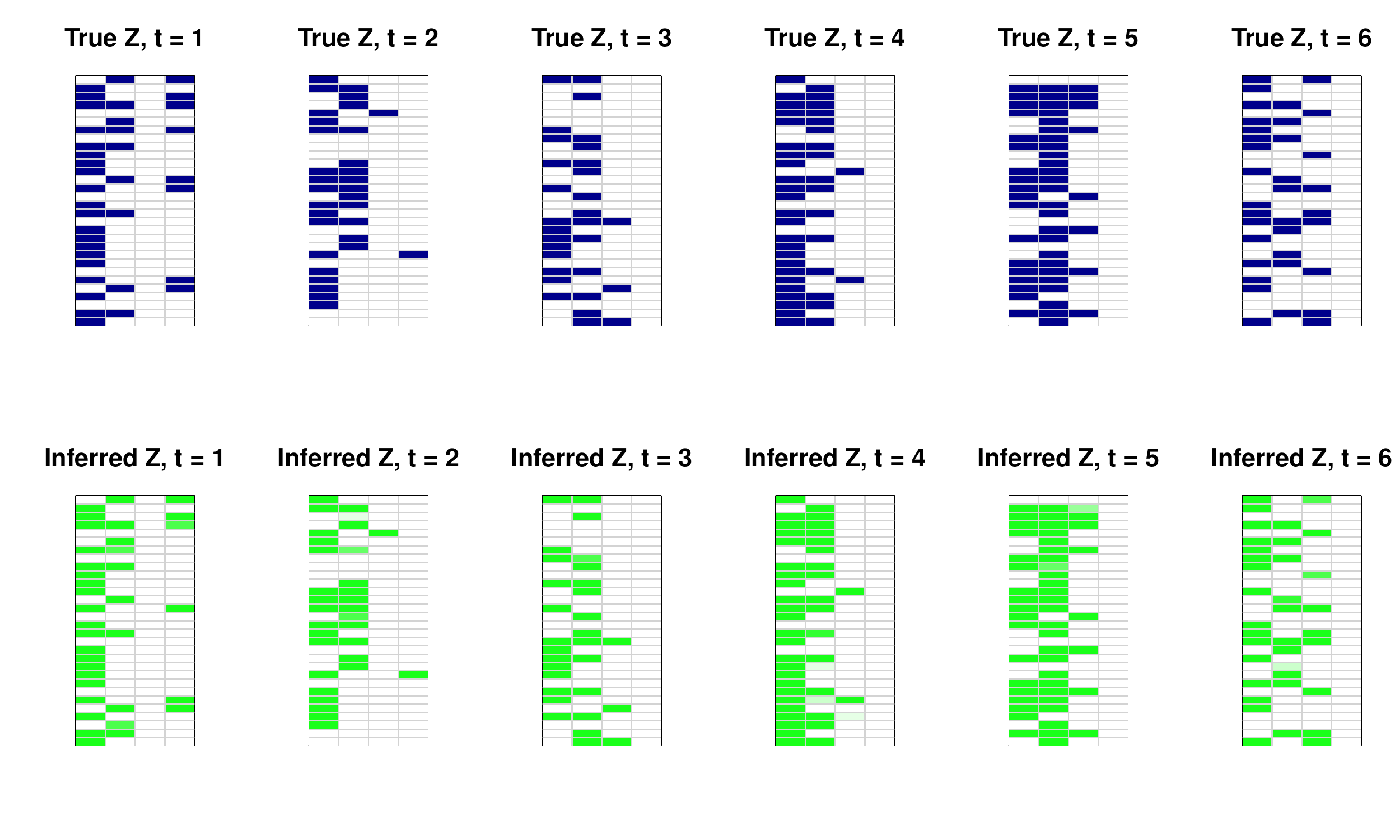}

\label{fig:2sub11}
\end{subfigure}
\begin{subfigure}{.5\textwidth}
\centering
\includegraphics[height=6cm,width=8cm]{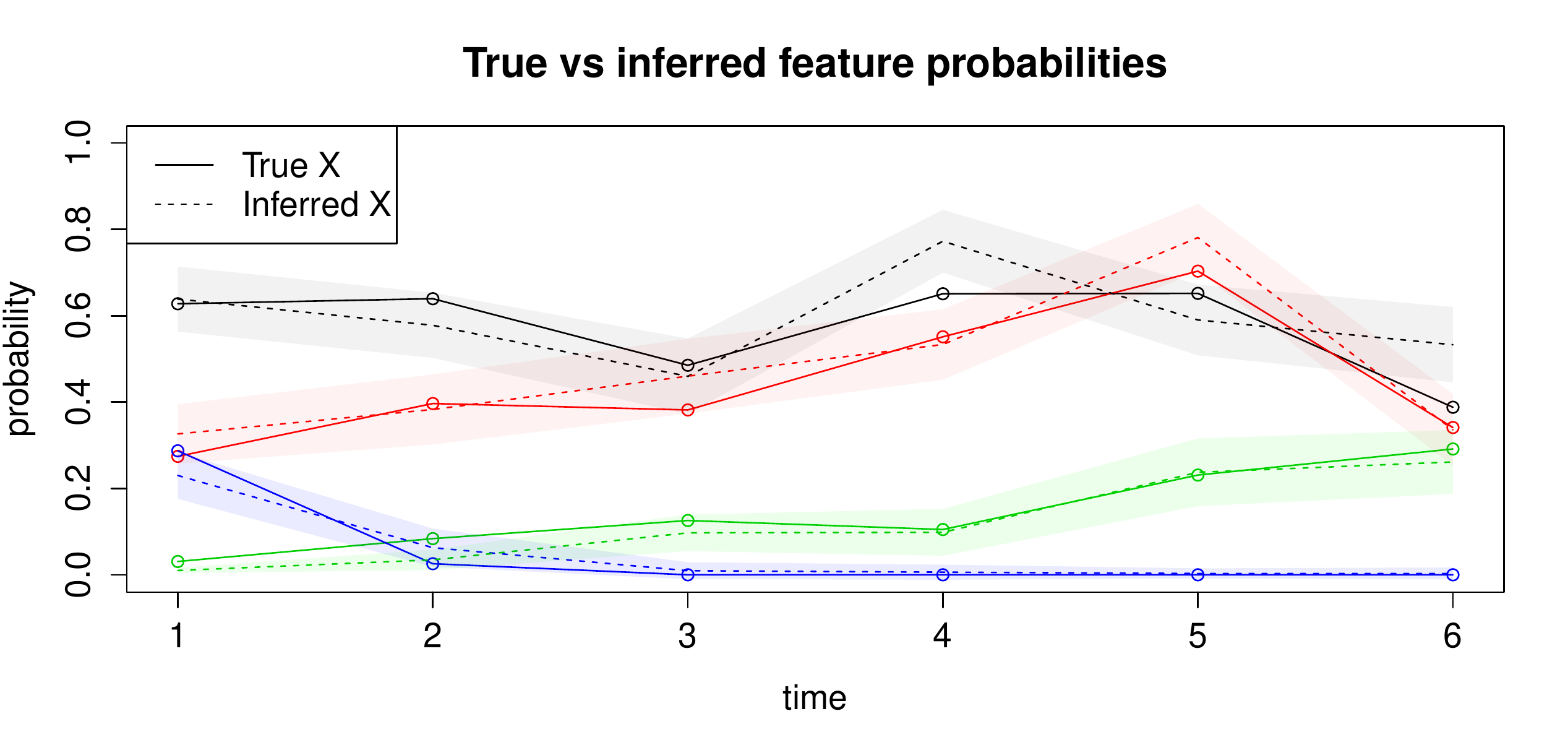}

\label{fig:prova1}
\end{subfigure}
\caption{WF-IBP. Left: Comparison between true and inferred feature allocation matrices. Right: Comparison between true and posterior mean feature trajectories over time (the shaded areas represent one standard deviation).}
\label{fig:inf_factor}
\end{figure*}

\begin{figure*}
\centering
\begin{subfigure}{.45\textwidth}
\centering
\includegraphics[height=9cm,width=7.1cm]{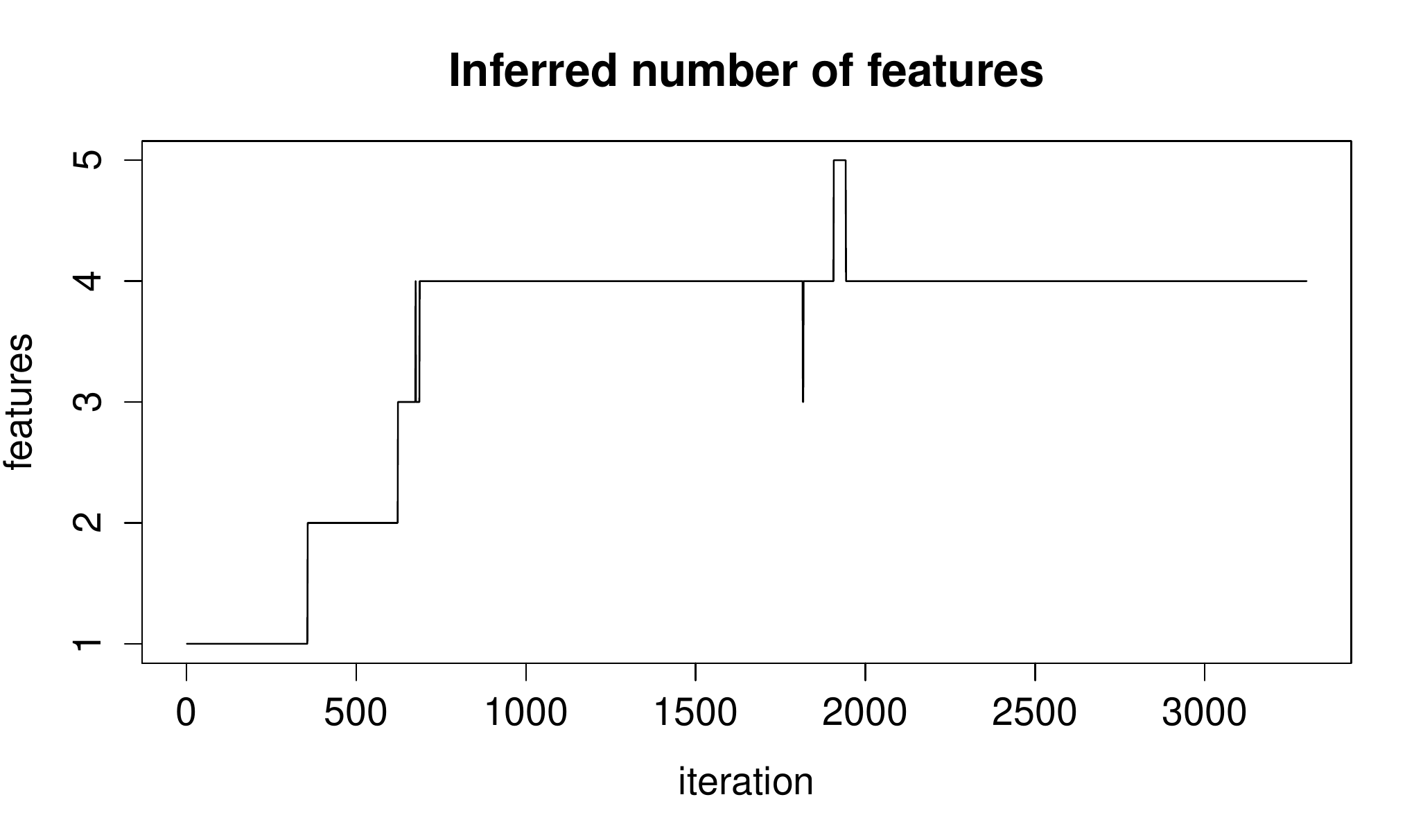}

\label{fig:3sub11}
\end{subfigure}
\begin{subfigure}{.45\textwidth}
\centering
\includegraphics[height=9cm,width=8.1cm]{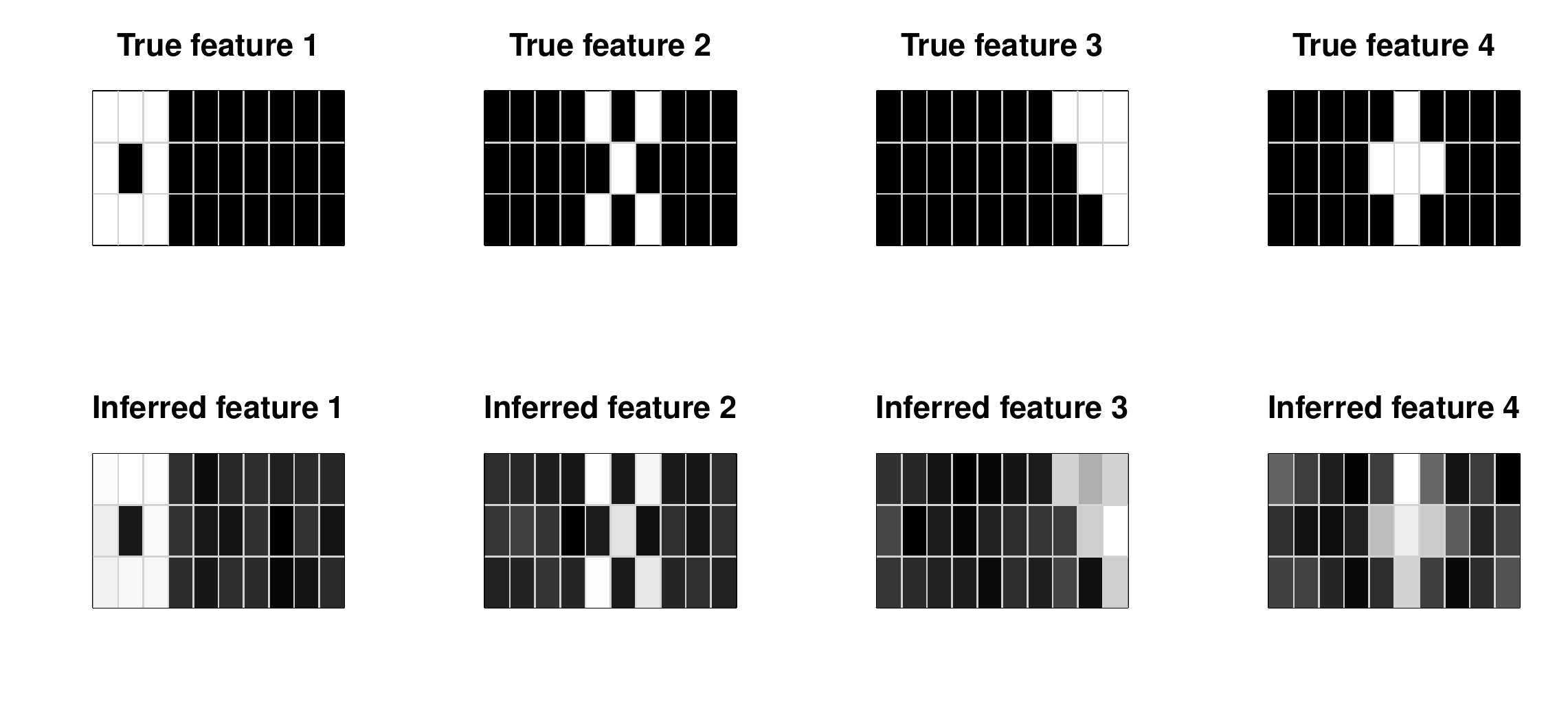}

\label{fig:prova2}
\end{subfigure}
\caption{WF-IBP. Left: Convergence to the true number of features. Right: Comparison between true and inferred features.}
\label{fig:inf_number}
\end{figure*}

Then, we tested the ability of the slice sampler-based algorithm to recover the correct number of latent features when given a similar set of synthetic data, this time consisting of $4$ latent features evolving over $6$ time points. The algorithm was initialized with one feature and run for 3300 iterations with a burn-in period of 2000 iterations. As in the finite case, the true underlying feature allocation matrices and feature probabilities have been closely recovered as illustrated by Figure $\ref{fig:inf_factor}$, as well as the features as shown by Figure $\ref{fig:inf_number}$-right. As shown by Figure $\ref{fig:inf_number}$-left, the correct number of latent features was detected in about 700 iterations.

\section{Topic modeling application} \label{sec:topic}
In this section we present an application of the WF-IBP to the modeling of corpora of time-stamped text documents. This is a natural application in that documents can be seen as arising from an unknown number of latent topics whose popularity is evolving over time. 

A class of models called Dirichlet processes aim at modeling the evolution of topics in a time-dependent and nonparametric way. Some of these models, however, assume the evolution of topic probabilities to be unimodal \cite{xing,gamma}, while others do not benefit from the conjugacy properties of the HDP and complicate inference \cite{varying,ddp}. The model presented in \cite{topictime} addresses these issues but, as most of HDP-based models, implicitly assumes a positive correlation between the probability of a topic being active and the proportion of that topic within each document. Such an assumption is undesirable as rare topics may account for a large proportion of words in the few documents in which they appear.

Our nonparametric and time-dependent topic model decouples the probability of a topic and its proportion within documents and offers a flexible way to model the probabilities of topics over time. We achieve this by generalizing to a time-dependent setting the focused topic model presented in \cite{ibpc}, which makes use of the IBP to select the finite number of topics that each document treats.

\subsection{WF-IBP topic model}
First, consider the case in which the number of topics $K$ underlying the corpus of seen documents is known. Define topics as probability distributions over a dictionary of $D$ words and model them as $(\rho_k)_{k=1}^K \stackrel{iid}{\sim} \text{Dirichlet}{(\bar{\eta})},$ given a vector $\bar{\eta}$ of length $D$. Let $\rho$ be the resulting vector and assume the components of $\bar{\eta}$ to be all equal to a constant $\eta>0$. Consider the usual setting in which the time-dependent popularity of topic (feature) $k$ is denoted by $X_k$  and the binary variables $Z_{ikt}$ indicate whether document $i$ contains topic $k$ at time $t$. Then, for all $t=t_0,\dots,t_T$ and $k=1,\dots,K$, sample
\begin{align*}
 \theta_{it} \mid \{ Z_{it}=z_{it},\phi_t = \phi'_t \} &\sim \text{Dirichlet}(z_{it} \circ \phi'_t), \quad \forall i=1,\dots,N_t,\\
  \phi_{kt} &\sim \text{Gamma}(\gamma,1), \\
(Z_{ikt})_{i=1}^{N_t} \mid \{ X_k=x_k(t)\} & \stackrel{iid}{\sim} \text{Bernoulli}(x_k(t)),\\
X_k & \sim \text{WF}\left(\frac{\alpha\beta}{K},\beta \right), 
\end{align*}
where $\phi_{kt}$ is the $k$th component of $\phi_t$, a $K$-long vector of topic proportions, and  $\theta_{it}$ the $i$th row of $\theta_t$, a $N_t \times K$ matrix with the distributions over topics for each document at time $t$. Note that the vectors $(\theta_{it})_{i=1}^{N_t}$ are independent. The operation $z_{it} \circ \phi_t$ stands for the Hadamard product between $z_{it}$ and $\phi_t$ and the Dirichlet is defined over the positive components of the resulting vector. While the topic allocation matrix $Z_t$ encodes which subset of the $K$ topics appears in each document at time $t$, the variables $\phi_{kt}$ are related to the proportion of words that topic $k$ explains within each document. Unlike HDP-based models, these two quantities are here modeled independently.

For every document $i=1,\dots,N_t$, draw the total number of words from a negative-binomial $W_{it} \sim$ NB($\sum_k z_{ikt} \phi_{kt}, 1/2)$ and, for each word $w_{ilt},$ $l=1,\dots,W_{it}$, sample first the topic assignment 
\begin{align*}
a_{ilt} \mid \{ \theta_{it} = \theta'_{it} \} \sim \text{Categorical}(\theta'_{it})
\end{align*}
and then the word
\begin{align*}
w_{ilt} \mid \{ a_{ilt} = a'_{ilt},\rho=\rho' \} \sim \text{Categorical}(\rho'_{a'_{ilt}}).
\end{align*}

Assume now that the number of potential topics $K$ needs to be learned from the data. The Bayesian nonparametric extension of this model is easily obtained by replacing the process generating the topic allocation matrices with the WF-IBP, so that topics arise as in the PRF and evolve as independent WF$(0,\beta)$. The feature allocation matrices can be drawn as described in Section \ref{z1z2}. In this way, we obtain a time-dependent extension of the IBP compound Dirichlet process presented in \cite{ibpc}.
\subsection{Posterior inference}
In order to infer the latent variables of the model, it is convenient to integrate out the parameters $\rho$ and $\theta$. This can be done easily thanks to the conjugacy between the Dirichlet and the Categorical distribution. In this way, we can run a Gibbs sampler for posterior inference only over the remaining latent variables and are able to follow the derivation of conditionals given by \cite{williamson}. Note that, in our case, we have introduced the slice variable and do not integrate out the topic allocation matrix. Denote by $W$ the complete set of words and by $A$ the complete set of topic assignments $a_{ilt}$ for all times $t=t_0,\dots,t_T$, documents $i=1,\dots, N_t$ and words $l=1,\dots,W_{it}$. Denote by $S_t$ the slice variable and by $W_t$ the complete set of words at time $t$. The conditional distributions that we need to sample from for all times $t=t_0,\dots,t_T$ are 
\begin{align*}
&p(A_t  \mid Z_t, w, \phi_t),\\
&p( \phi_t,  \gamma  \mid  A_t, W_t, Z_t), \\
&p( S_t  \mid  Z_t,X(t)), \\
&p(Z_t  \mid  A_t, X(t), \phi_t,S_t).
\end{align*}
Conditioning on all the other topic assignments $a_{-il}$, each topic assignment $a_{ilt}$ can be sampled from
 \begin{align*}
 p(a_{ilt}=k  \mid  a_{-il},Z_t, W,  \phi_t)  \propto
  (n^{w_{il}}_k + \eta) \frac{n^i_{kt} + \phi_{kt} z_{ikt}}{n_k + \eta D - 1},
\end{align*}
where $n^{w_{il}}_k$ denotes the number of times that word $w_{il}$ has been assigned to topic $k$ excluding assignment $a_{il}$, $n^i_{kt}$ the number of words assigned to topic $k$ in document $i$ excluding assignment $a_{il}$ and $n_k$ the total number of words assigned to topic $k$. 

After placing a hyper-prior on $p(\gamma)$, we can sample $\phi$ and $\gamma$ via a Metropolis-Hastings step. Indeed, we know that
\begin{align*}
p( \phi_{kt},  \gamma  \mid  A, Z_t) \propto
\frac{\phi_{kt}^{\gamma-1}e^{-\phi_{kt}}}{\Gamma(\gamma)} p(\gamma) 
 \prod_{i=1}^{N_t} \frac{\Gamma(\phi_{kt}z_{ikt} + n^i_{kt})}{\Gamma(\phi_{kt}z_{ikt})n^i_{kt}! 2^{\phi_{kt}z_{ikt}+n^i_{kt}}}.
\end{align*}
Conditioning on $Z_t$, the slice variable is sampled according to its definition:
\begin{align*}
p( S_t  \mid  Z_t,X(t))  = \frac{1}{x^*(t)} 1 _{s_t}([0,x^*(t)]),
\end{align*}
where  $x^*(t)$ is the minimum among the probabilities of the active topics at time $t$.
As for the feature allocation matrices $Z$, we sample only the finite number of its components whose topic probability $x_k(t)$ is greater than the slice variable $S_t$. Assume we are sampling each entry $Z_{ikt}$ sequentially and denote respectively by $x^*_1(t)$ and $x^*_0(t)$ the minimum active topic probability in the cases $Z_{ikt}=1$ and $Z_{ikt}=0$. Let $n_{ikt}$ denote the total number of words assigned to topic $k$ in document $i$ at time $t$. Then we have that
 \begin{equation*}
 p(Z_{ikt}=1  \mid  A, x_k(t), \phi_{kt}) =
    \break
    \newline
    \begin{cases}
      1, & \text{if}\ n_{ikt} > 0 \\
      \frac{x_k(t)x^*_0(t)}{x_k(t)x^*_0(t) + 2^{\phi_{kt}}(1-x_k(t))x^*_1(t)}, &  \text{if}\ n_{ikt} = 0.
    \end{cases}
  \end{equation*}
  
 \begin{equation*}
 p(Z_{ikt}=0  \mid  A, x_k(t), \phi_{kt}) =
    \break
    \newline
    \begin{cases}
      0, & \text{if}\ n_{ikt} > 0 \\
      \frac{2^{\phi_{kt}}(1-x_k(t))x^*_1(t)}{x_k(t)x^*_0(t) + 2^{\phi_{kt}}(1-x_k(t))x^*_1(t)}, &  \text{if}\ n_{ikt} = 0.
    \end{cases}
  \end{equation*}
The full conditional distributions presented so far are derived in Appendix B. As in the linear-Gaussian case, when considering the probability of setting $Z_{ikt}=1$ for a feature that is currently inactive across all observations at time $t$, it is necessary to jointly propose a new value $\phi_{kt}$ by drawing it from its prior distribution $\text{Gamma}(\gamma,1)$. Finally, inference on the trajectories of the topic probabilities is a direct application of the Particle Gibbs and thinning scheme outlined for the general case.

\subsection{Simulation and results}
\begin{figure*}[!htb]
\centering
\begin{subfigure}{.45\textwidth}
\centering
\includegraphics[height=8cm,width=7cm]{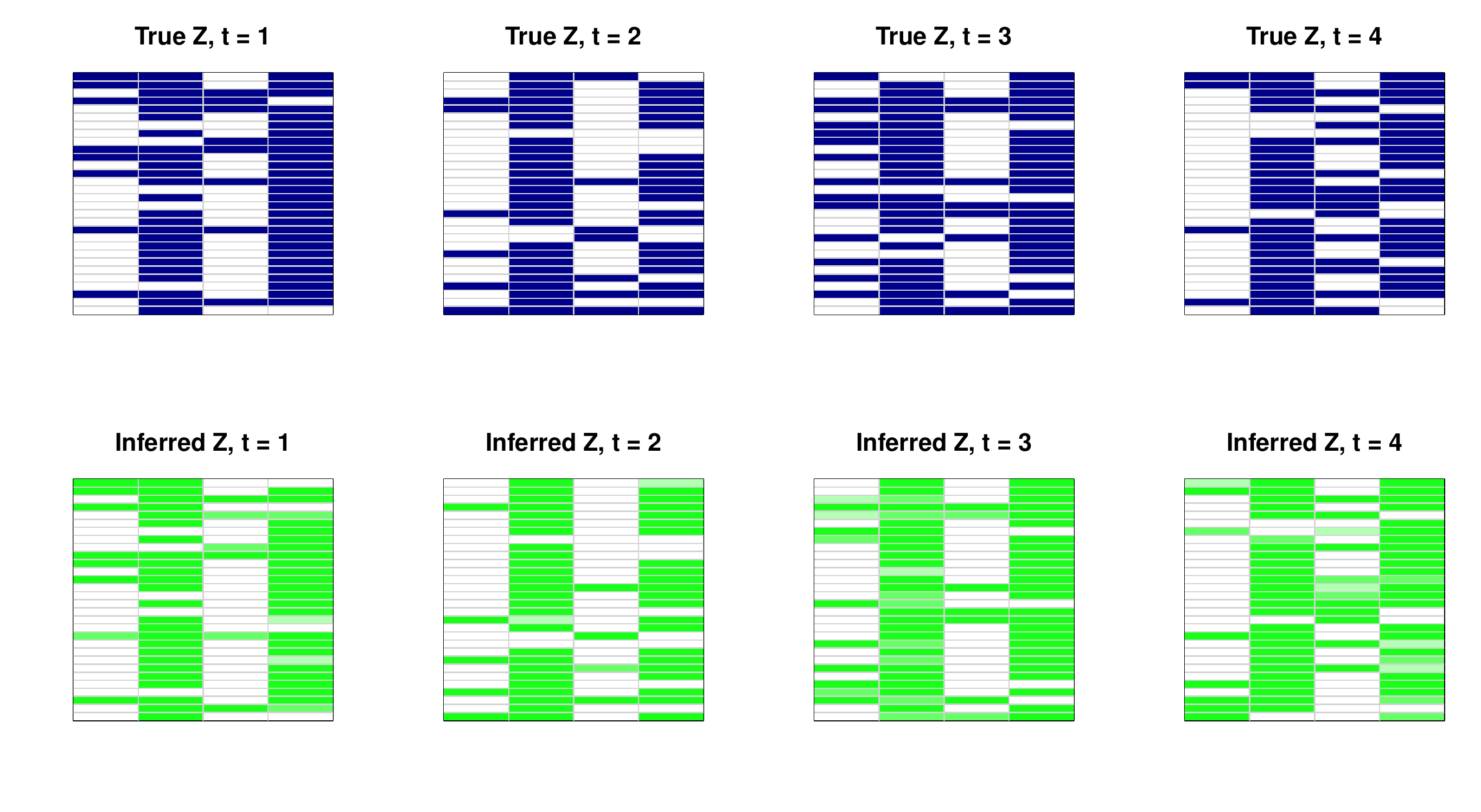}

\label{fig:41sub11}
\end{subfigure}
\begin{subfigure}{.45\textwidth}
\centering
\includegraphics[height=8cm,width=7cm]{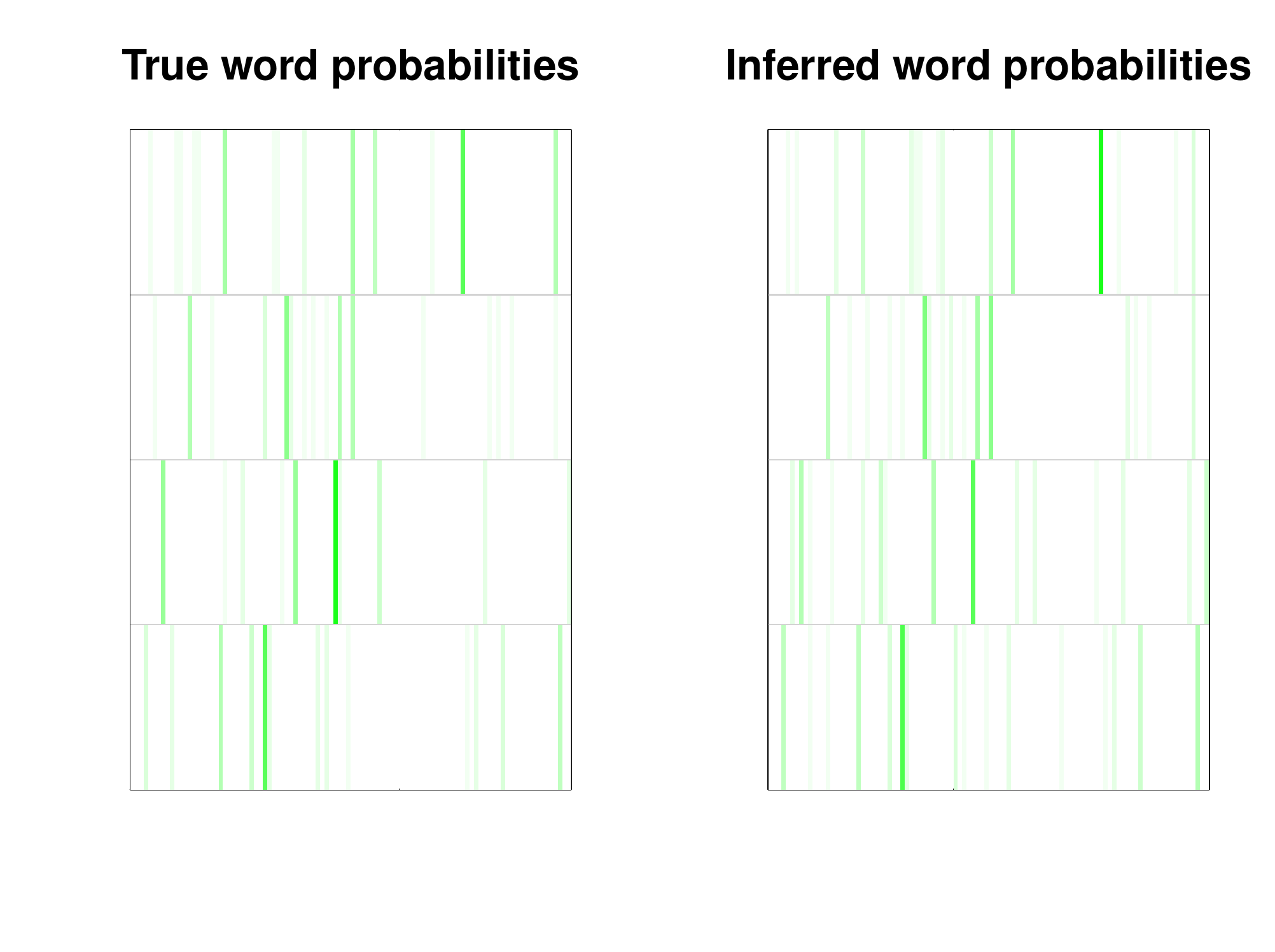}

\label{fig:3sub21}
\end{subfigure}
\caption{Left: Comparison between the true and the posterior mean topic allocation matrices at each time. Right: True vs inferred distributions over words for each topic. Each row is a topic ($K = 4$) and each column is a word from the dictionary ($D = 100$) (the darker the green, the larger the probability of the corresponding word).}
\label{fig:topic2}
\end{figure*}

At each of $4$ time points, a small corpus of $N=30$ documents was simulated by selecting up to $K=4$ latent topics for each document and picking words from a dictionary of $D=100$ words. The hyper-parameter of the Dirichlet prior over words was chosen to be a vector with components equal to $\eta=0.1$, a Gamma(5,1) hyper-prior was placed on $\gamma$ and we let topic probabilities evolve as independent $\WF(1,1)$ diffusions with $0.1$ diffusion time-units between each observation. Assume $K$ is known and focus on inference over the remaining parameters. Fix the time-units and drift parameters of the W-F diffusion to their true values in the PG update. We ran the Gibbs sampler for 5000 iterations with a burn-in period of 300 iterations. The algorithm was able to infer closely the latent topic allocation matrices (Figure $\ref{fig:topic2}$-left) and the percentage of words assigned to the correct topics were $81 \%$ at $t_1$, $82 \%$ at $t_2$, $83 \%$ at $t_3$ and $85 \%$ at $t_4$.

A Monte Carlo estimate for the probability of word $w=1,\dots,D$ under topic $k$ is
\begin{align*}
\hat{\rho}_{kw} = \frac{n_k^w + \eta}{n_k + D \eta},
\end{align*}
where $n_k^w$ is the number of times word $w$ has been assigned to topic $k$,  $n_k$ is the total number of words assigned to topic $k$ and $D$ is the number of words in the dictionary.  
A Monte Carlo estimate for the probability of topic $k$ in document $i$ is
\begin{align} \label{topic_prop}
\hat{\theta}_{ikt} = \frac{n_{ikt} + z_{ikt}\phi_{kt}}{\sum_k (n_{ikt} + z_{ikt}\phi_{kt})}.
\end{align}
These two quantities are given by the posterior mean of the Dirichlet distribution under a categorical likelihood. $\hat{\rho}_{kw}$ has been used to plot the posterior distribution over words in Figure $\ref{fig:topic2}$-right. These results confirm the ability of the algorithm to recover ground truth and provide useful information both at word and topic level.

\begin{figure*}[ht!]
\centering
\includegraphics[height=10cm,width=15.cm]{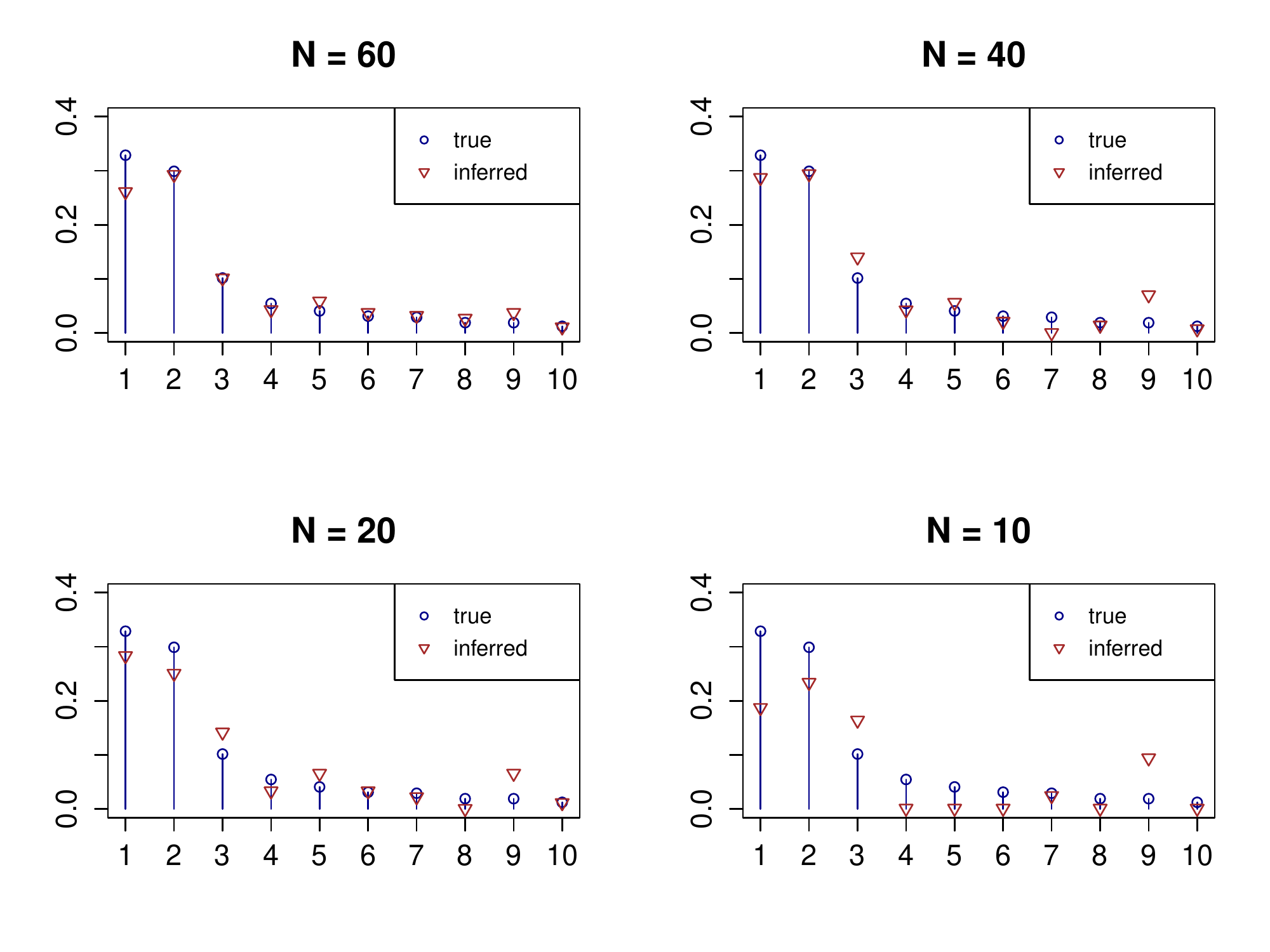}
\caption{Comparison, given different numbers $N$ of observed documents, between the true probabilities of the 10 most likely words within a given topic and the inferred probabilities of those words for that topic. The more documents are observed, the better the reconstruction of the topic is.}
\label{fig:topic_comp}
\end{figure*}

Finally, we tested the ability of the algorithm to reconstruct topics when presented with a decreasing number of observed documents. 100 documents were simulated at each of two time points as described above. Figure \ref{fig:topic_comp} shows how well the probability of the 10 most likely words of the first topic was reconstructed for $N=60$, $40$, $20$ and $10$ observed documents per time point. As expected, the more documents are observed the more accurate the reconstruction of topics is, with a drop in performance when only 10 documents per time point are observed.

\FloatBarrier
\subsubsection{Comparison with a static version}
We compare the dynamic topic model presented so far with a static counterpart in which time is not modeled and thus information about the timestamp of documents is not exploited. In particular, we investigate whether incorporating time into the model improves upon test-set perplexity, a measure widely used in topic modeling settings that assesses the ability of topic models to generalize to unseen data. Given the model parameters $\Phi$, perplexity on documents $D_{test}:=\{d_i\}_{i=1}^M$ is defined as 
\begin{align*}
\text{perplexity} (D_{test}  \mid  \Phi ) = \text{exp } \left( - \frac{ \sum _{m=1}^{M}  \text{log } p(d_i  \mid  \Phi)}{ \sum _{m=1}^{M} W_i}  \right),
\end{align*}
where $W_i$ denotes the number of words in document $d_i$. As we assume that words within each document are drawn independently given the model parameters $\Phi$, the probability of document $d_i$ can be computed as 
\begin{align*}
p(d_i  \mid  \Phi) = \prod_{l=1}^ {W_i} p(w_{il}  \mid  \Phi),
\end{align*} 
where
\begin{align*}
p(w_{il}  \mid  \Phi) = \sum_{k=1}^K \theta_{ik} \rho_{kl},
\end{align*}
recalling that $\theta_{ik}$ is the probability of a generic word belonging to topic $k$ in document $i$ and $\rho_{kl}$ is the probability of word $l$ under topic $k$. These two quantities can be approximated at each iteration of the MCMC algorithm by their current values $\hat{\theta}_{ik}^{(s)}$ and $\hat{\rho}_{kl}^{(s)}$, so that we can approximate the probability of each word by averaging over $S$ samples of the Markov Chain.
\begin{align*}
\hat{p}(w_{il}  \mid  \Phi) = \frac{1}{S} \sum_{s=1}^{S} \sum_{k=1}^K \hat{\theta}_{ik}^{(s)} \hat{\rho}_{kl}^{(s)}.
\end{align*}
Note that the perplexity is inversely proportional to the likelihood of the data and thus lower values indicate better performance. Chance performance, namely assuming each word to be picked uniformly at random from the dictionary, yields a perplexity equal to the size $D$ of the dictionary.

\begin{figure*}[htp!]
\centering

\includegraphics[height=7cm,width=16cm]{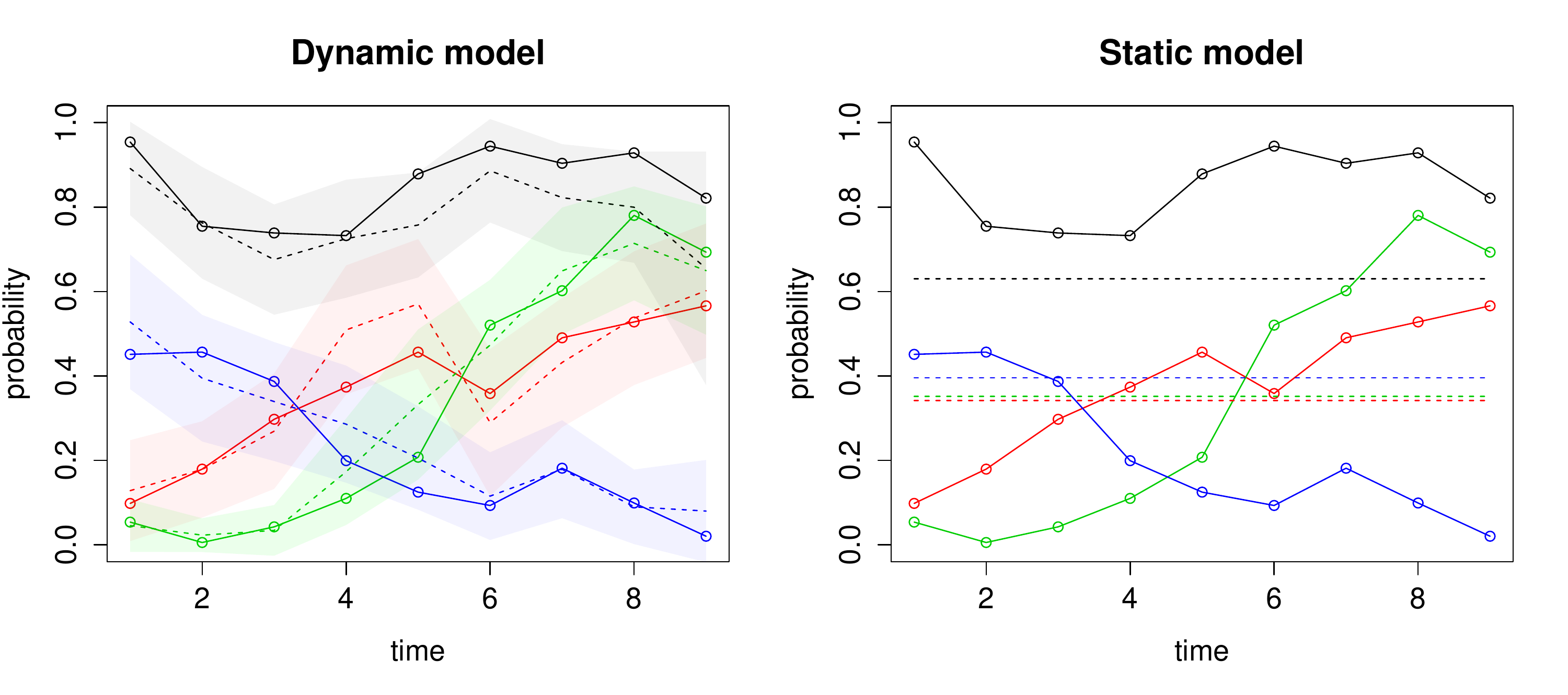}

\caption{Comparison between true and inferred feature probabilities (respectively continuous and dotted lines) in the fixed-$K$ topic model (left) and in a static version that does not incorporate time (right).}
\label{fig:dynstat}
\end{figure*}

\begin{figure*}[!htbp]
\centering

\includegraphics[height=9cm,width=14cm]{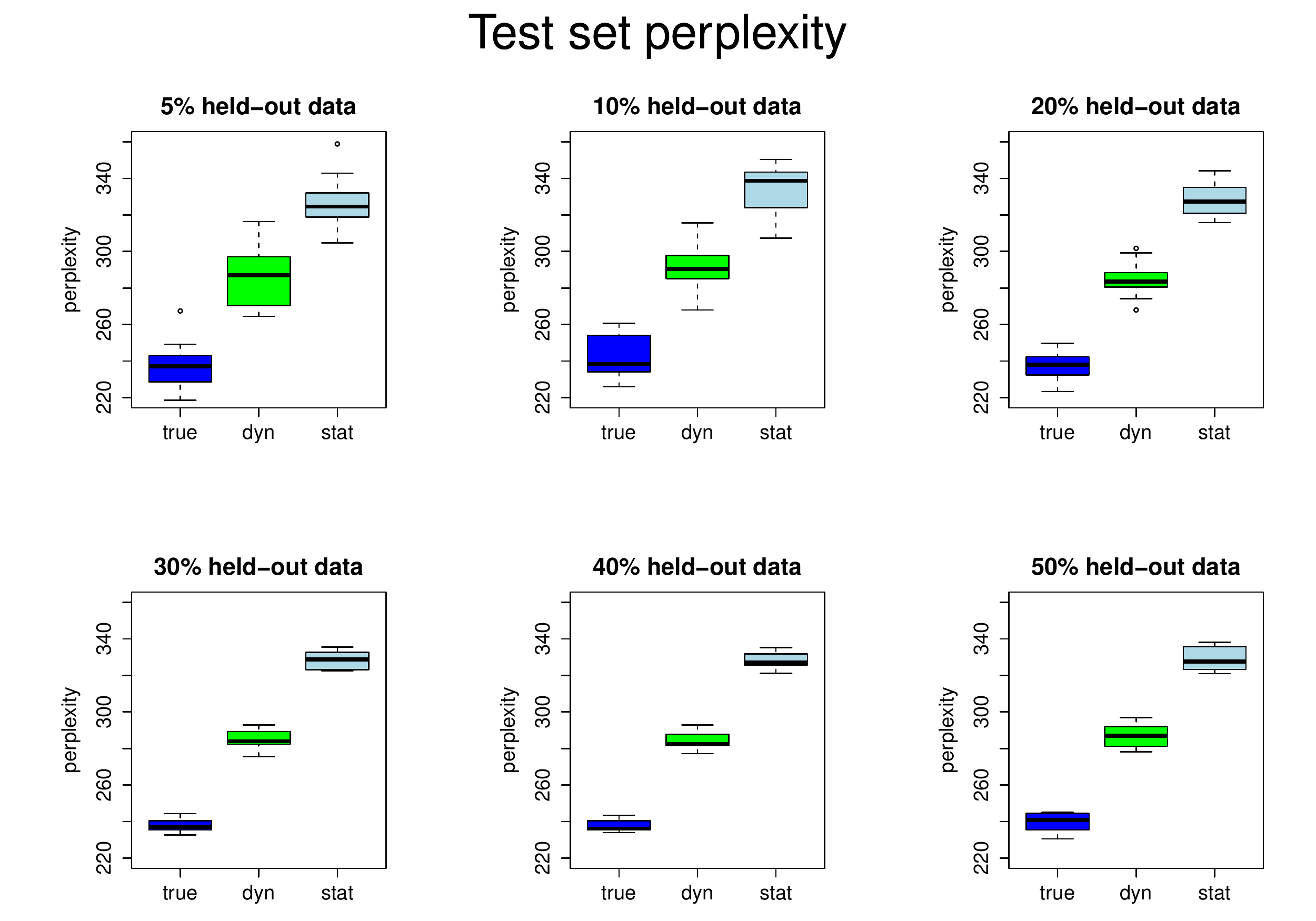}

\caption{Boxplots of test-set perplexity in the true model, in the fixed-$K$ dynamic model and in the static model at different percentages of held-out data. Each boxplot was obtained by computing the perplexity after holding-out several different random subsets of words in the data. Lower values indicate better performance.}
\label{fig:perplexity}
\end{figure*}

Different percentages of words were held-out and the model was trained on the remaining data. Testing the model on held-out words is a way to avoid comparing different hyper-parameters, as different treatments of the hyper-parameters could strongly affect the results \cite{perplex}. A dictionary of $D=1000$ words was used to generate 30 documents at each of 9 time points. The number of features was fixed to 4 in both the dynamic and the static algorithm and they were run for 3000 iterations with a burn-in period of 300 iterations. Even though both algorithms approximately recover the true topic allocations matrices, incorporating time leads to a closer match. This can be measured, for instance, by the Frobenius norm of the difference between each true and inferred $Z_t$. Table $\ref{table:frobenius}$ shows that at each time the dynamic model leads to a lower discrepancy with the true topic allocation matrix. Figure $\ref{fig:dynstat}$ shows the posterior trajectories of topic probabilities inferred by the dynamic model and compares them with the constant values inferred by the static model. Finally, Figure $\ref{fig:perplexity}$ compares the test-set perplexity of the two models (recall that in this case chance performance results in a perplexity of $D=1000$). The results show that incorporating time leads to a more flexible model that improves upon the ability to recover ground truth as well as to generalize to unseen data.

\begin{table}[ht!]
\centering
\caption{Frobenius norm of the difference with the true $Z_t$.}
\label{table:frobenius}
\begin{tabular}{ | l | l | l | l | l | l | l | l | l | l | }
\hline
  & $t_1$ & $t_2$ & $t_3$ & $t_4$ & $t_5$ & $t_6$ & $t_7$ & $t_8$ & $t_9$\\ \hline
$Dynamic$ &1.78 & 1.37 & 0.72 & 2.99 & 3.24 & 2.01 & 2.59 & 2.45 & 2.88 \\ \hline
$Static$ &3.63 & 4.23 & 2.84 & 3.01 & 3.55 & 5.27 & 4.89 & 3.31 & 5.90  \\ \hline

\end{tabular}
\end{table}

\subsubsection{Real-world data experiments}
We used the WF-IBP topic model to explore the data set consisting of the full text of 5811 NIPS conference papers published between 1987 to 2015. We pre-processed the data and removed words appearing more than 5000 times or fewer than 250 times. Our goal was to discover what topics appear in the corpus and to track the evolution of their popularity over these 29 years.

Observe that the time points at which observations are seen influence the number of time units for which the W-F diffusions should be simulated. This adds a time parameter to the model, which can be tuned to account for how strong the time-dependency between the observed data is or the fact that the time gaps between successive observations may vary. We set the hyperparameters $\alpha, \beta$ of the W-F diffusion equal to 1 and the time step to $0.12$ diffusion time-units per year so as to reflect realistic evolutions of topic popularity. The Markov chain was run for 2000 iterations with a burn-in period of 200 iterations, setting $\eta = 0.001$ and placing a Gamma(5,1) hyper-prior on $\gamma$.

\begin{figure*}
\centering
\includegraphics[height=8cm,width=15.cm,trim= 4 14 4 20,clip]{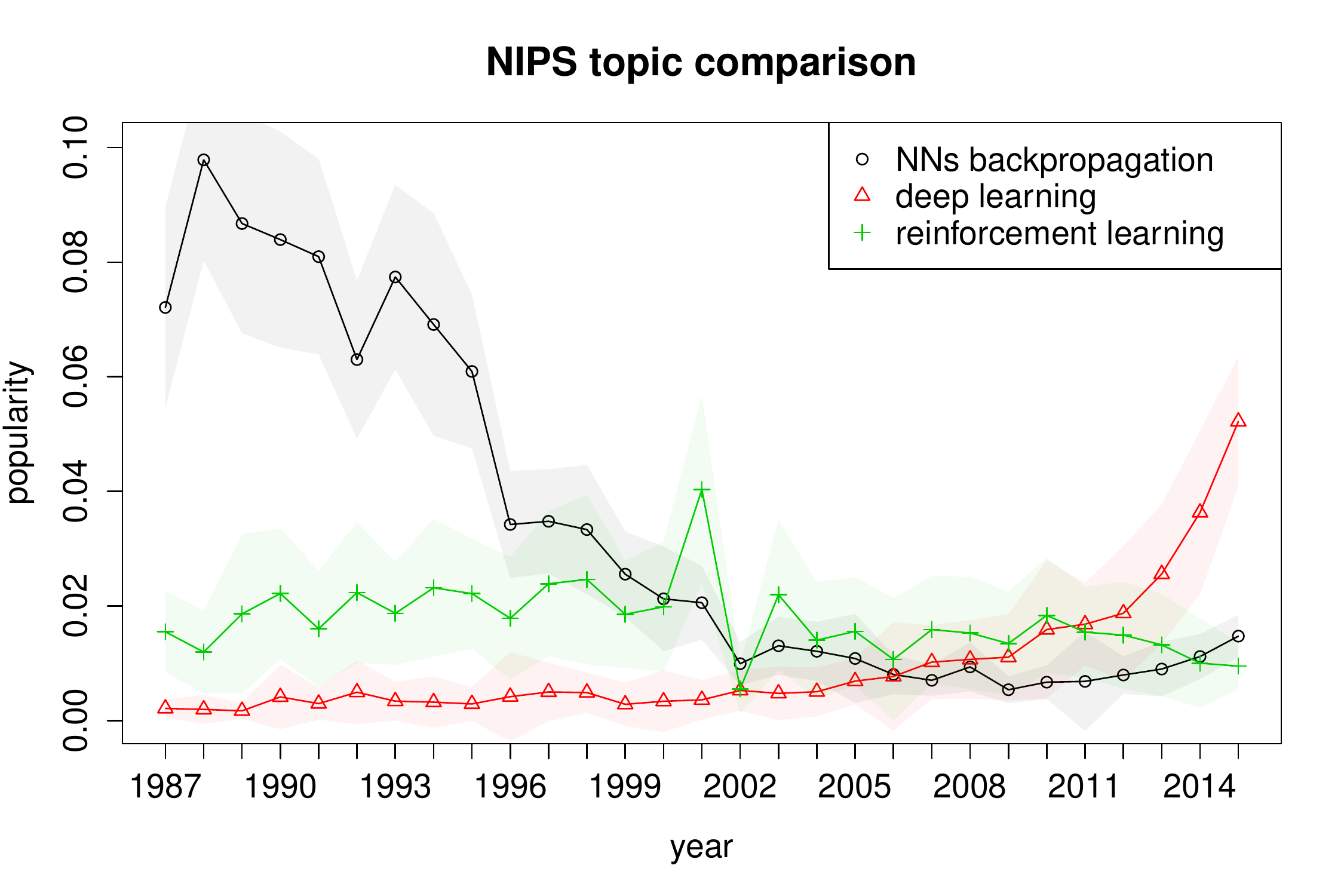}
\caption{Comparison of a set of posterior topic proportions over the years 1987-2015 (NIPS data set).}
\label{fig:topicTHETA}
\end{figure*}

Recall that topics are defined by their distribution over words, so it is possible to label them by looking at their most likely words. The 12 most likely words of 32 topics found in the corpus together with the evolution of their topic proportions are given in Figure \ref{fig:individual_topics}, where the shaded areas represent one standard deviation around the posterior means. Observe that, with very few exceptions, the topics detected in the corpus are meaningful and easily interpretable. 
One of the qualitative advantages of modeling time dependency explicitly is that interesting insights into the evolution of topics underlying large collections of documents can be obtained automatically, and uncertainty in the predictions naturally incorporated. For instance, Figure \ref{fig:topicTHETA} compares how the popularity of three different approaches to machine learning evolved over time. The results indicate that standard neural networks ('NNs backpropagation') were extremely popular until the early 90s. After this, they went through a steady decline, only to increase in popularity later on. This confirms the well known fact that NNs were largely forsaken in the machine learning community in the late 90s \cite{deeplearning}. On the other hand, it can be observed that the popularity of deep architectures and convolutional neural networks ('deep learning') steadily increased over these 29 years, to the point that deep learning became the most popular among all topics in NIPS 2015.

\begin{figure*}[ht!]
\centering
\includegraphics[height=8cm,width=13cm]{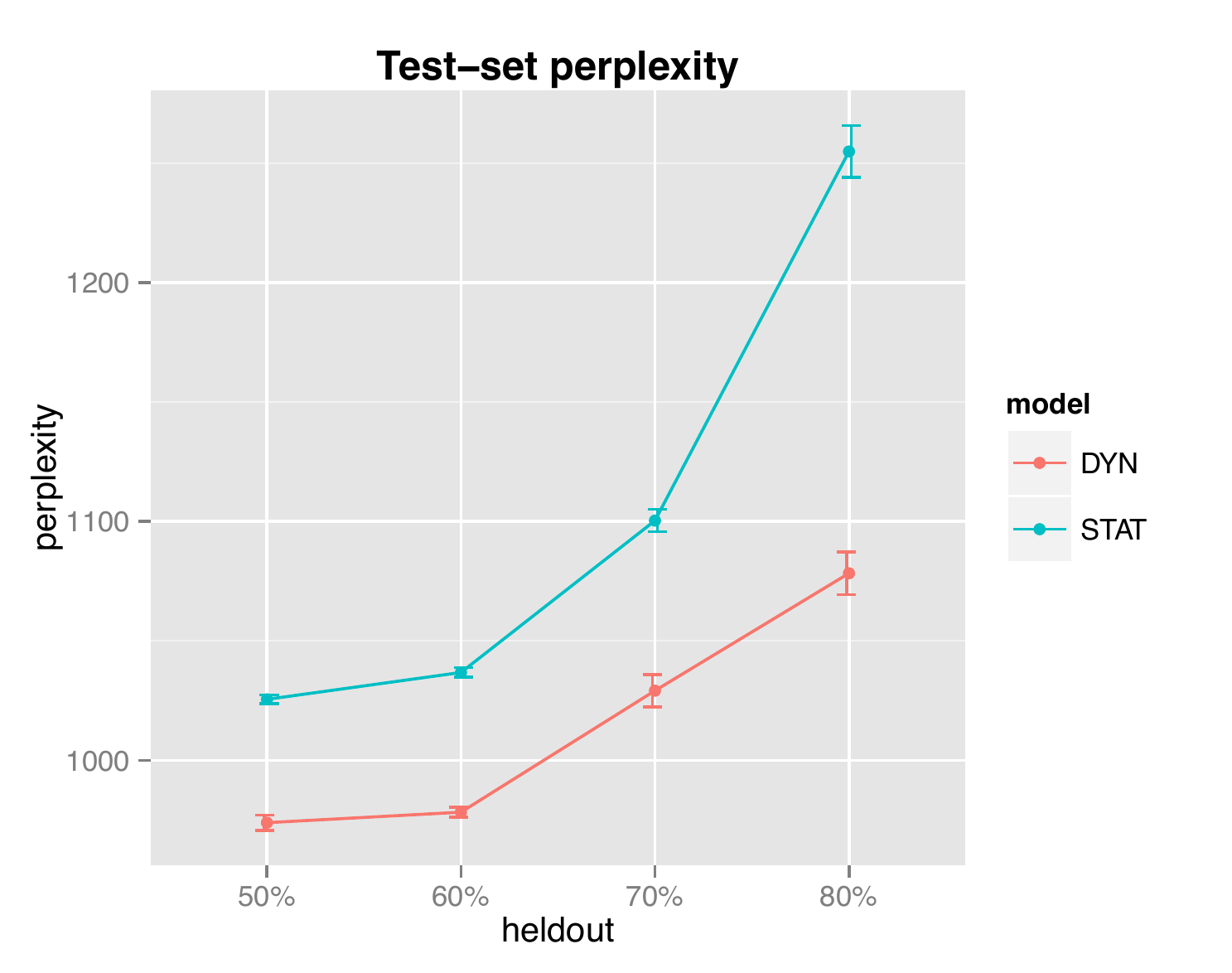}
\caption{Comparison between the test-set perplexity of the dynamic model and its static counterpart after holding out different percentages of words (NIPS data set). The former consistently performs better than the latter, with a substantial difference when the percentage of held-out words is large.}
\label{fig:perplexityNIPS}
\end{figure*}

More quantitatively, we compared the predictive performance of our fixed-$K$ truncation with its static counterpart in which time is not modeled. The results in Figure \ref{fig:perplexityNIPS} were obtained by holding out different percentages of words (50$\%$, 60$\%$, 70$\%$ and 80$\%$) from all the papers published in 1999 and by training the model over the papers published from 1987 to 1998. The goal was to investigate whether incorporating time dependency improves the predictions on future documents at time $t+1$ when given the documents up to time $t$. The held-out words were then used to compute the test-set perplexity after 5 repeated runs with random initializations (the error bars represent one standard deviation). The dynamic model led to consistently better results, especially as the number of held-out words was increased. This follows by the fact that, the less training data is available in the year in which the models are tested, the more crucial capturing time-dependence.

\begin{figure*}[ht!]

\centering

\begin{subfigure}{.3243\textwidth}
\centering
\includegraphics[height=4.4cm,width=1.18\textwidth]{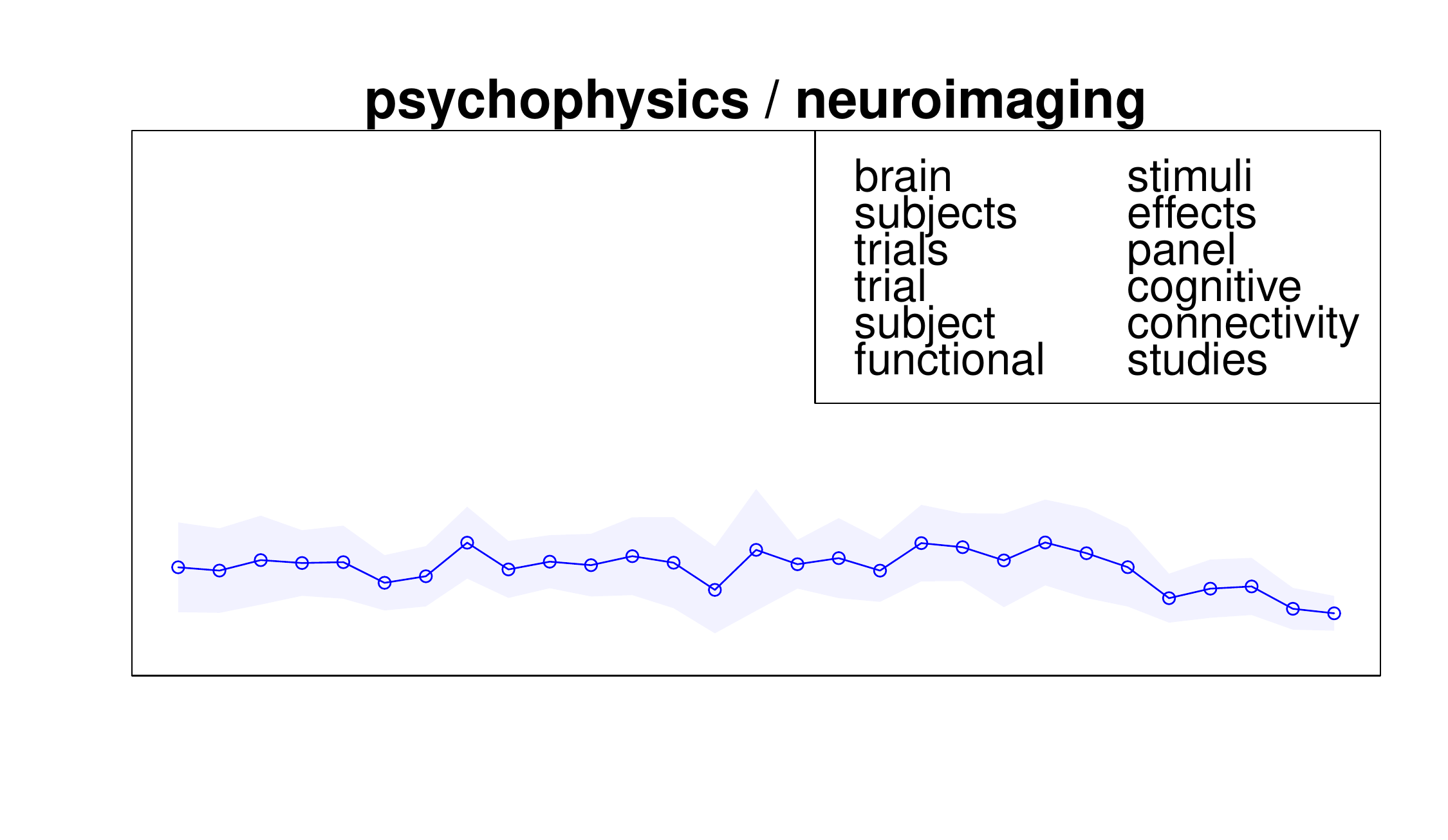}
\end{subfigure}
\begin{subfigure}{.3243\textwidth}
\centering
\includegraphics[height=4.4cm,width=1.18\textwidth]{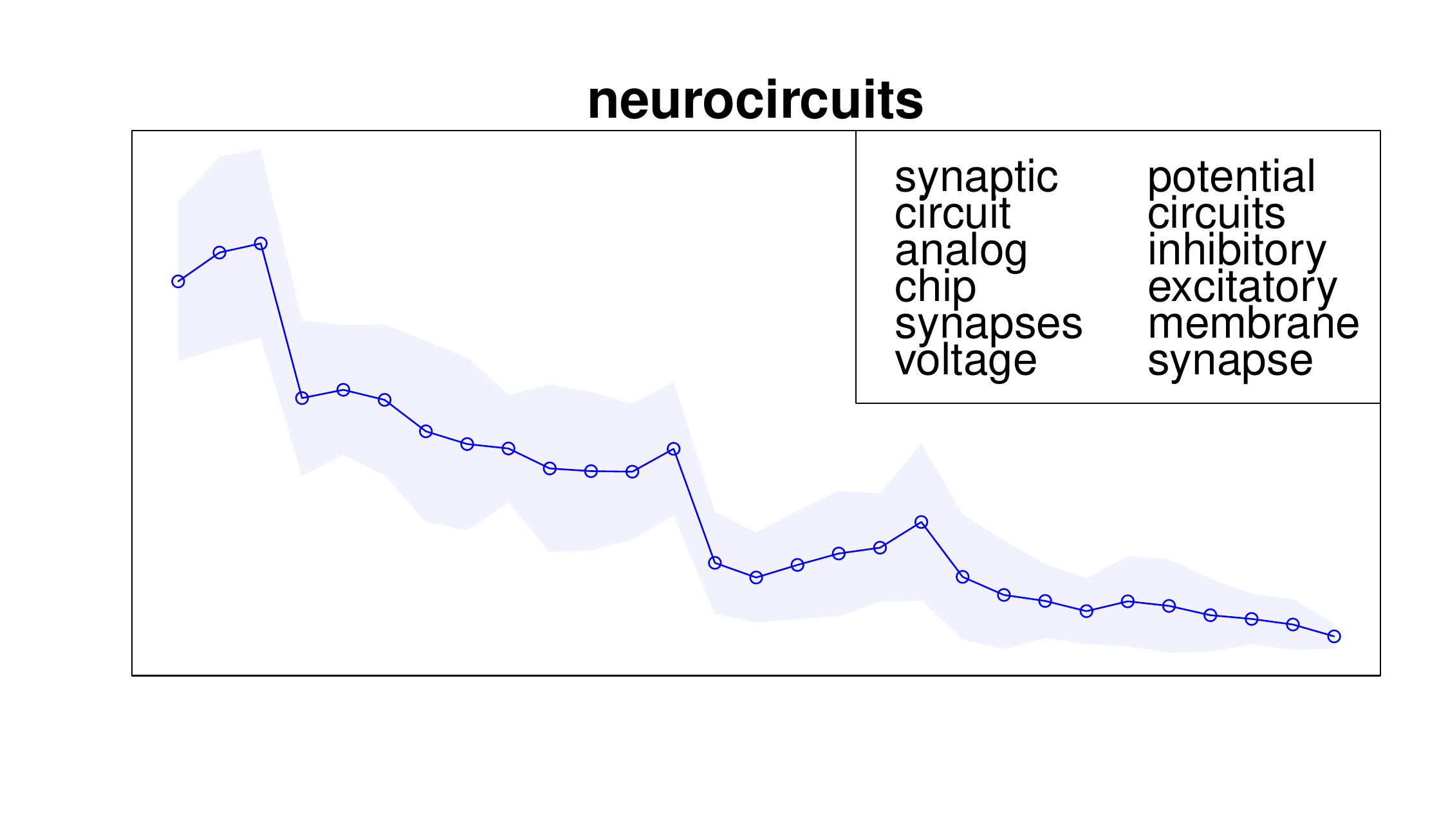}
\end{subfigure}
\begin{subfigure}{.3243\textwidth}
\centering
\includegraphics[height=4.4cm,width=1.18\textwidth]{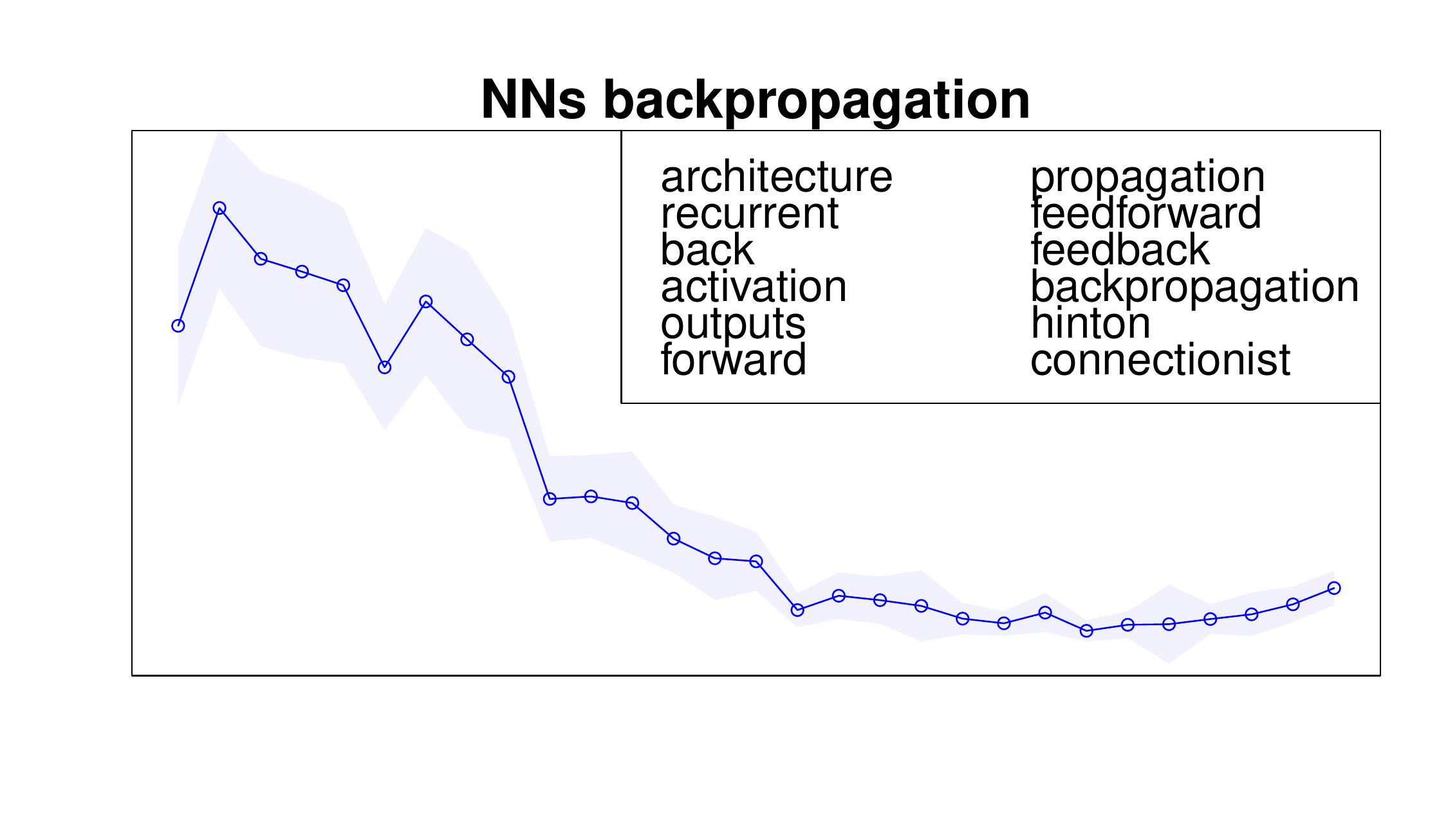}
\end{subfigure}

\vspace{-0.4in}

\begin{subfigure}{.3243\textwidth}
\centering
\includegraphics[height=4.4cm,width=1.18\textwidth]{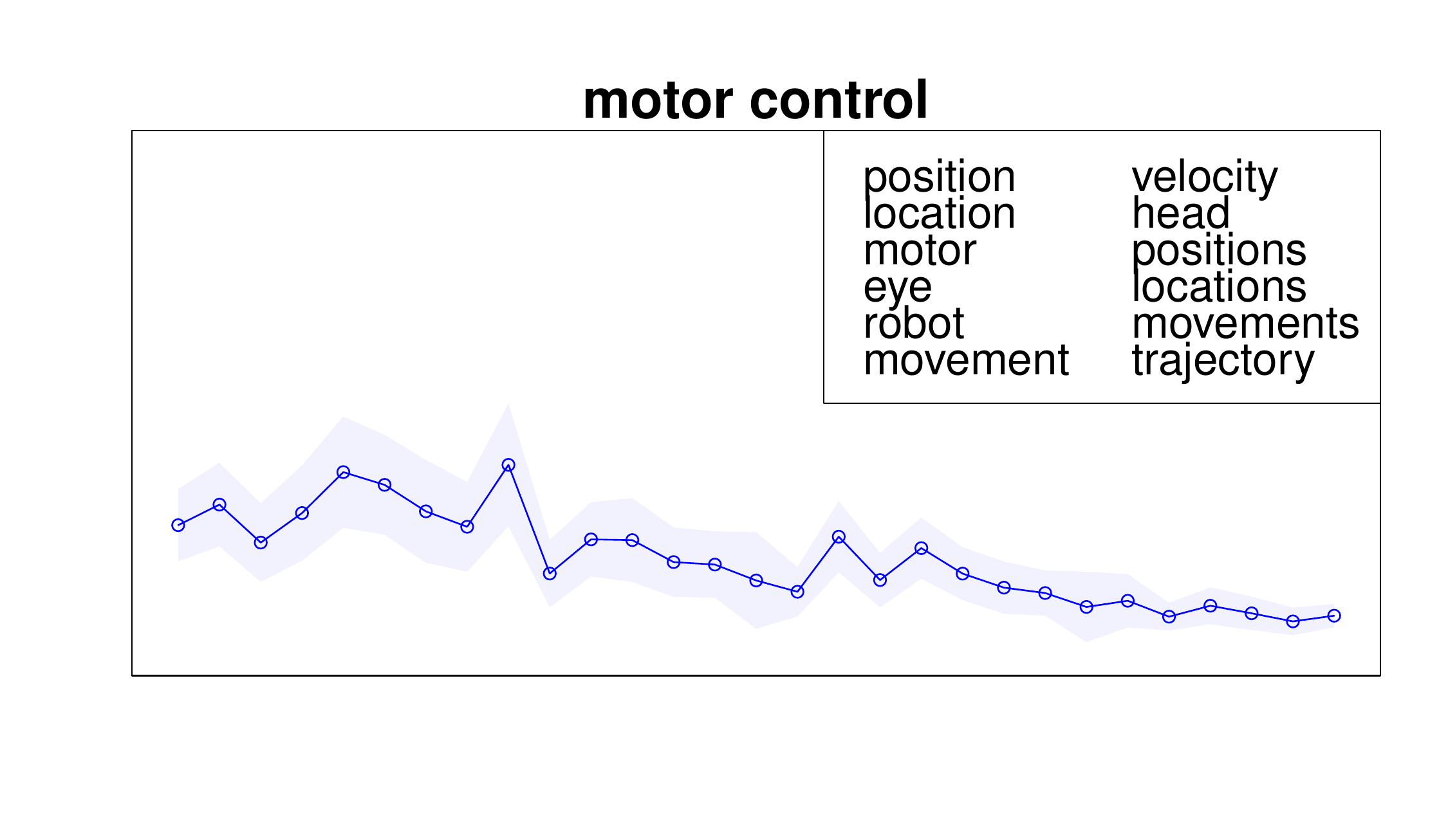}
\end{subfigure}
\begin{subfigure}{.3243\textwidth}
\centering
\includegraphics[height=4.4cm,width=1.18\textwidth]{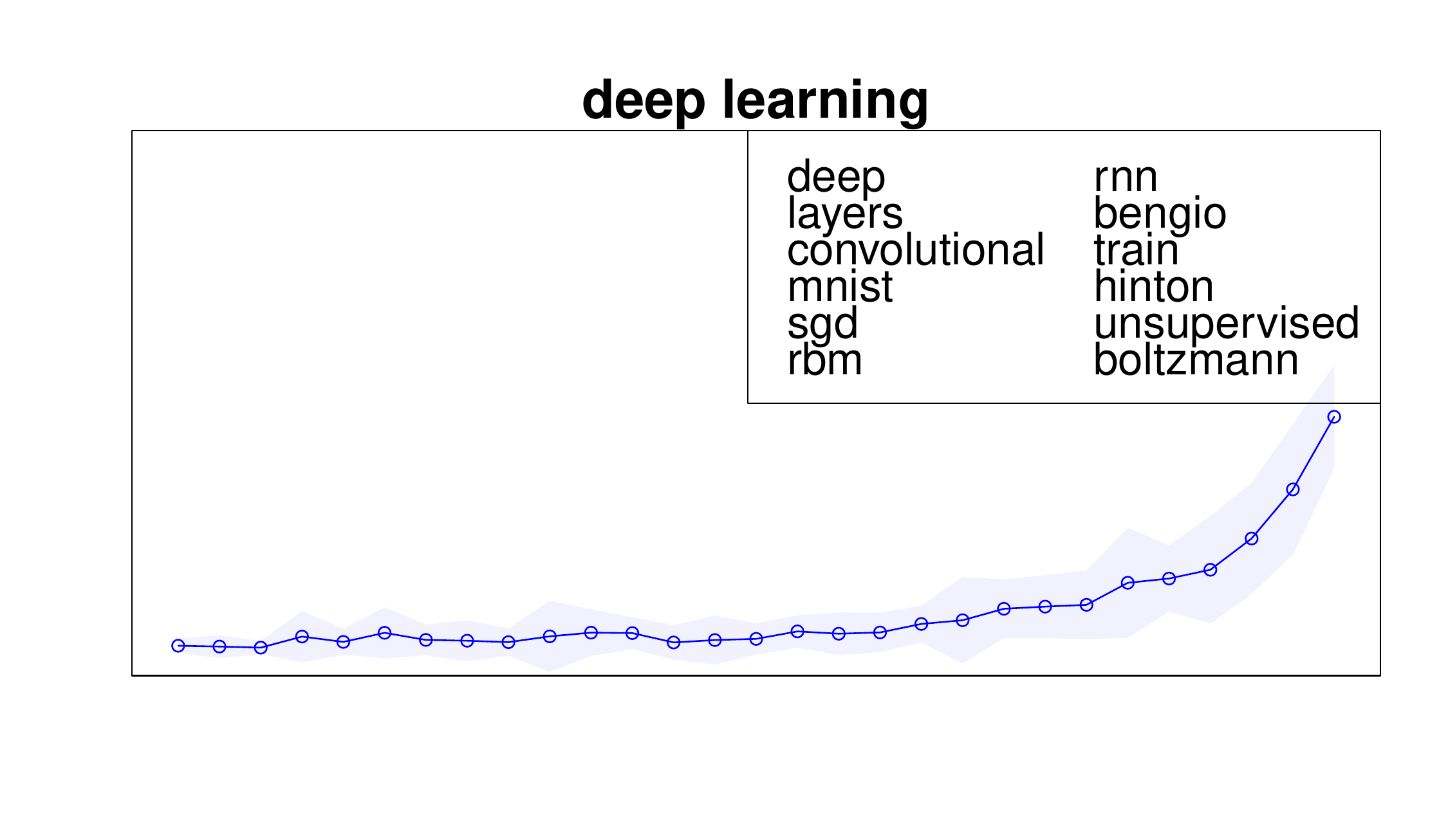}
\end{subfigure}
\begin{subfigure}{.3243\textwidth}
\centering
\includegraphics[height=4.4cm,width=1.18\textwidth]{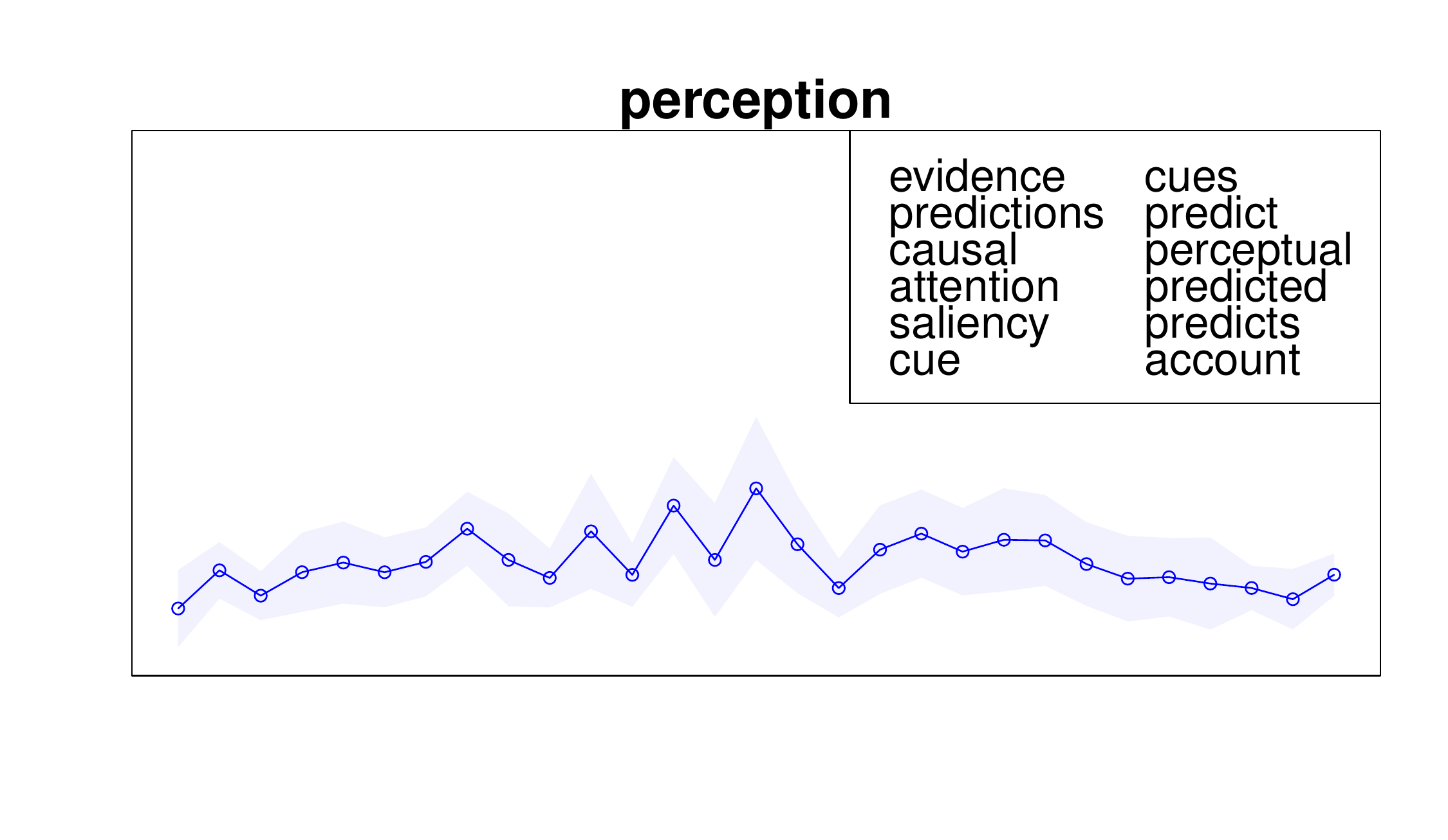}
\end{subfigure}

\vspace{-0.4in}

\begin{subfigure}{.3243\textwidth}
\centering
\includegraphics[height=4.4cm,width=1.18\textwidth]{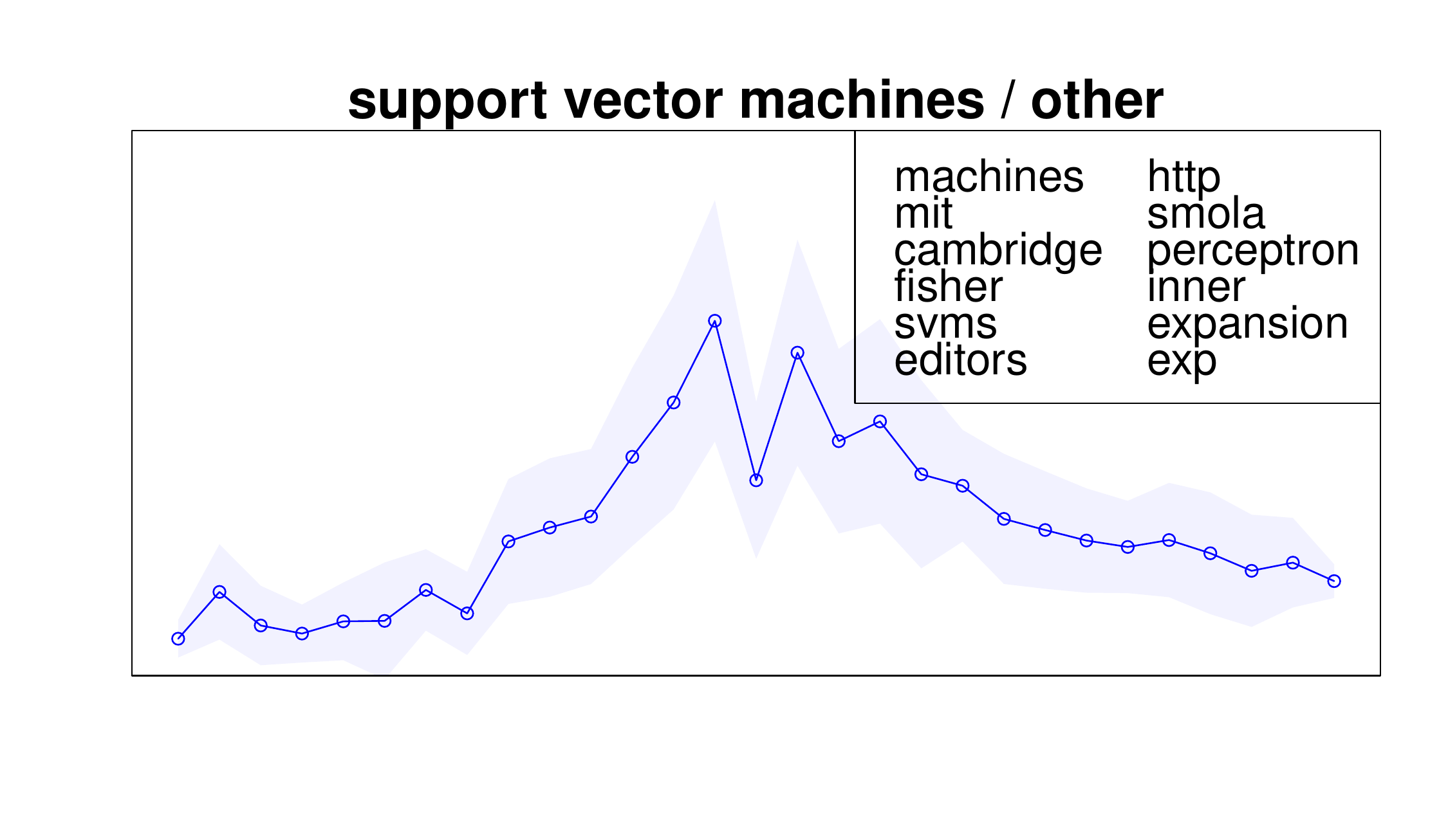}
\end{subfigure}
\begin{subfigure}{.3243\textwidth}
\centering
\includegraphics[height=4.4cm,width=1.18\textwidth]{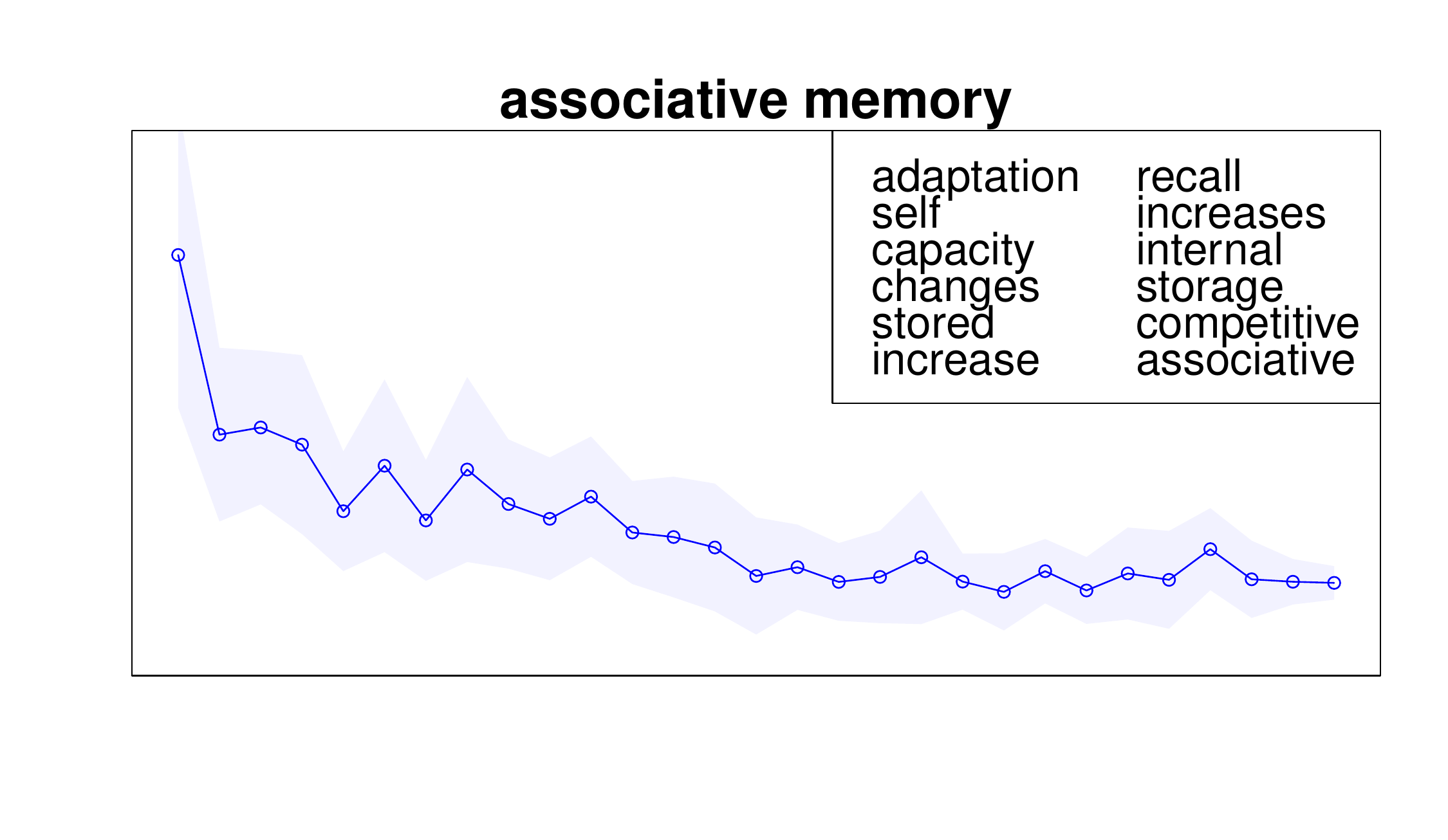}
\end{subfigure}
\begin{subfigure}{.3243\textwidth}
\centering
\includegraphics[height=4.4cm,width=1.18\textwidth]{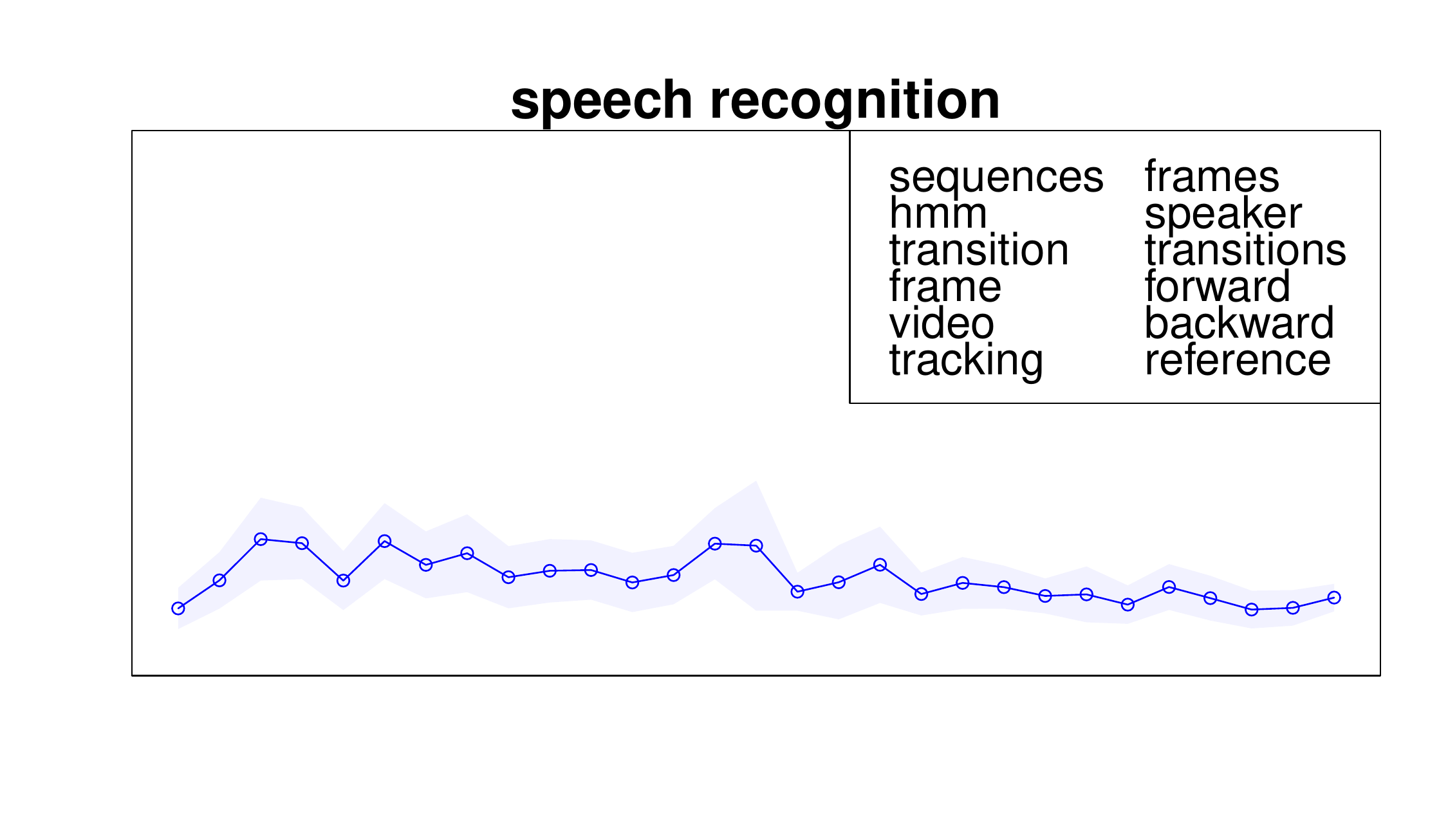}
\end{subfigure}

\vspace{-0.4in}

\begin{subfigure}{.3243\textwidth}
\centering
\includegraphics[height=4.4cm,width=1.18\textwidth]{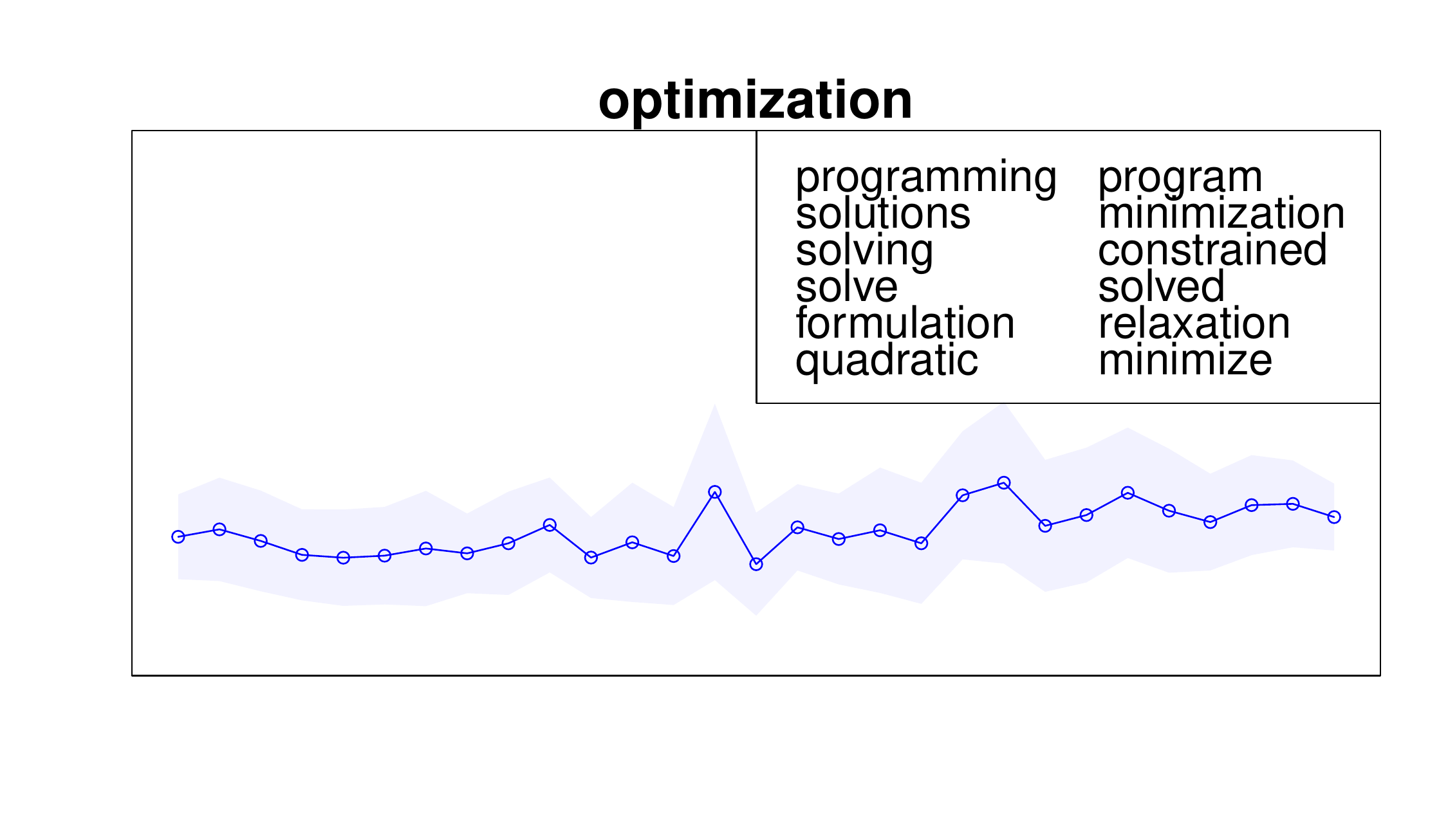}
\end{subfigure}
\begin{subfigure}{.3243\textwidth}
\centering
\includegraphics[height=4.4cm,width=1.18\textwidth]{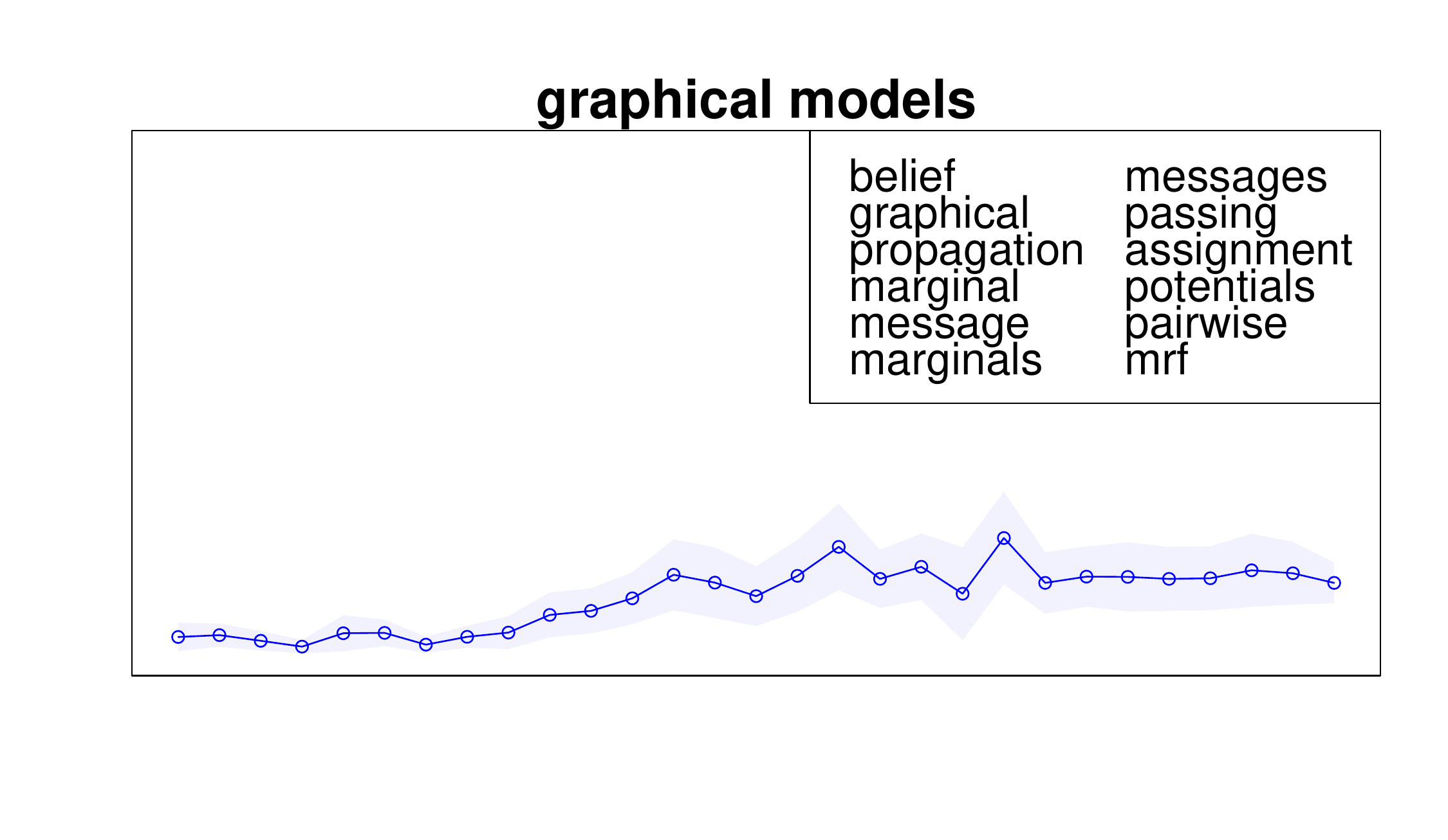}
\end{subfigure}
\begin{subfigure}{.3243\textwidth}
\centering
\includegraphics[height=4.4cm,width=1.18\textwidth]{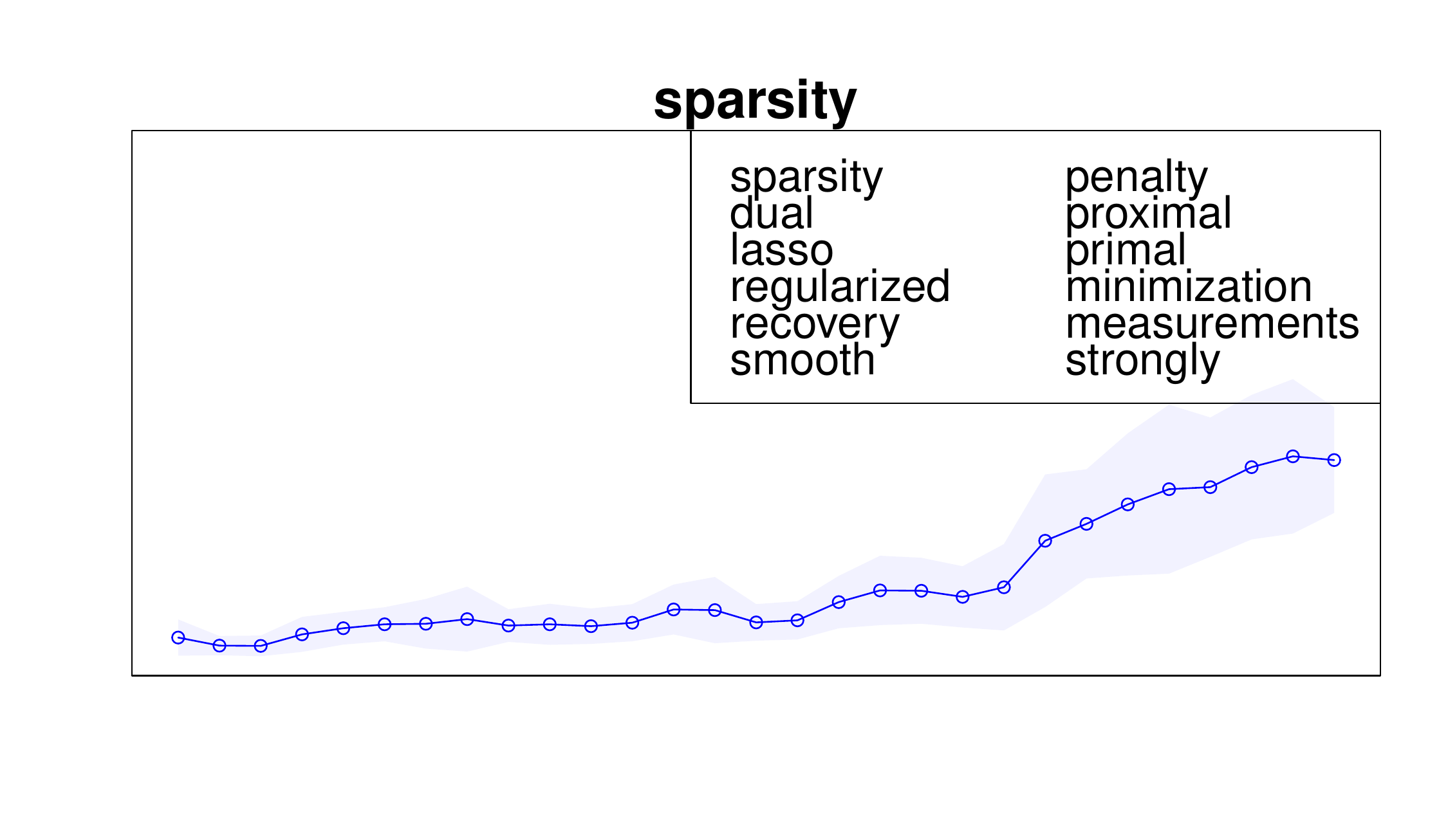}
\end{subfigure}

\vspace{-0.4in}

\begin{subfigure}{.3243\textwidth}
\centering
\includegraphics[height=4.4cm,width=1.18\textwidth]{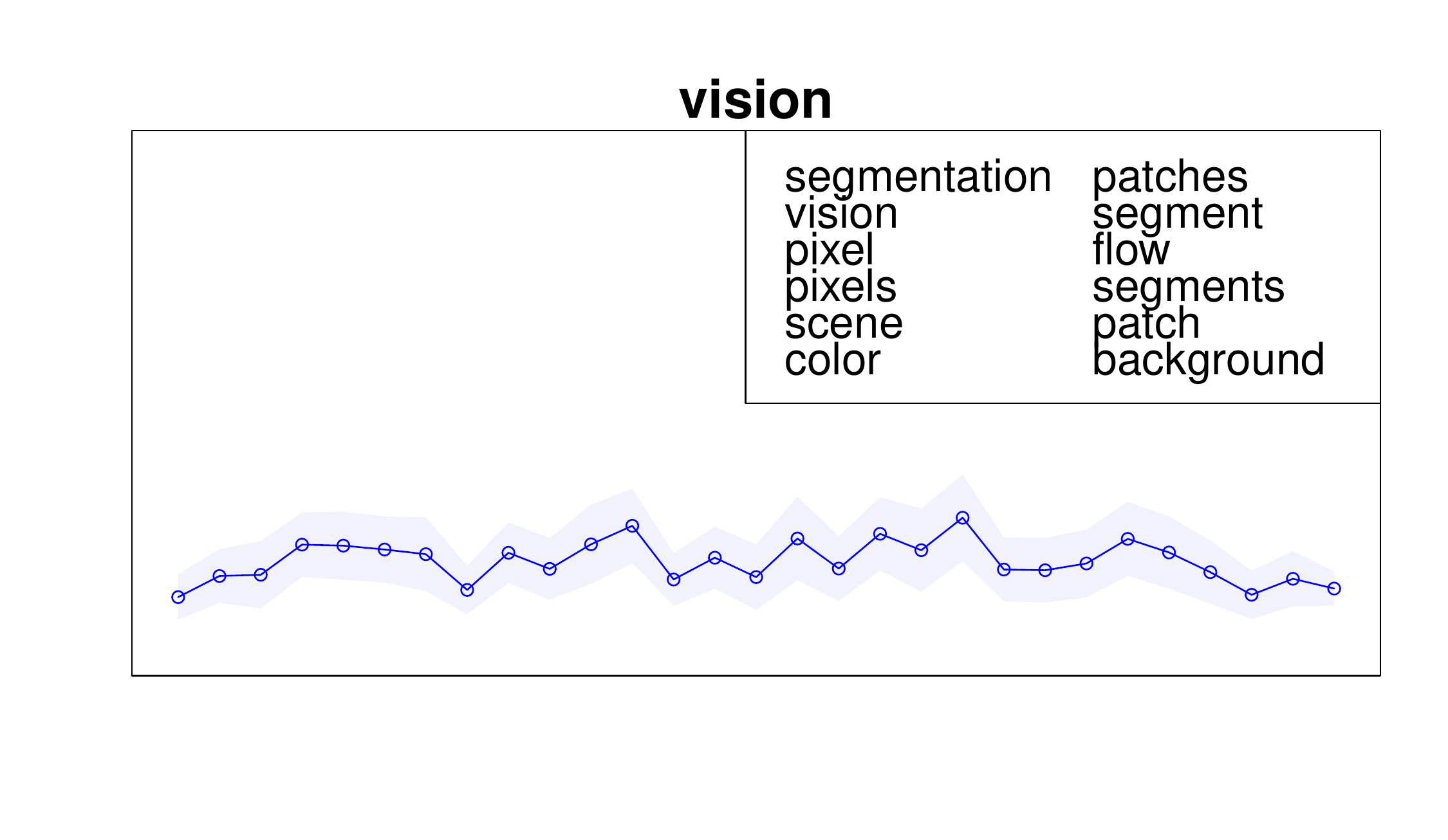}
\end{subfigure}
\begin{subfigure}{.3243\textwidth}
\centering
\includegraphics[height=4.4cm,width=1.18\textwidth]{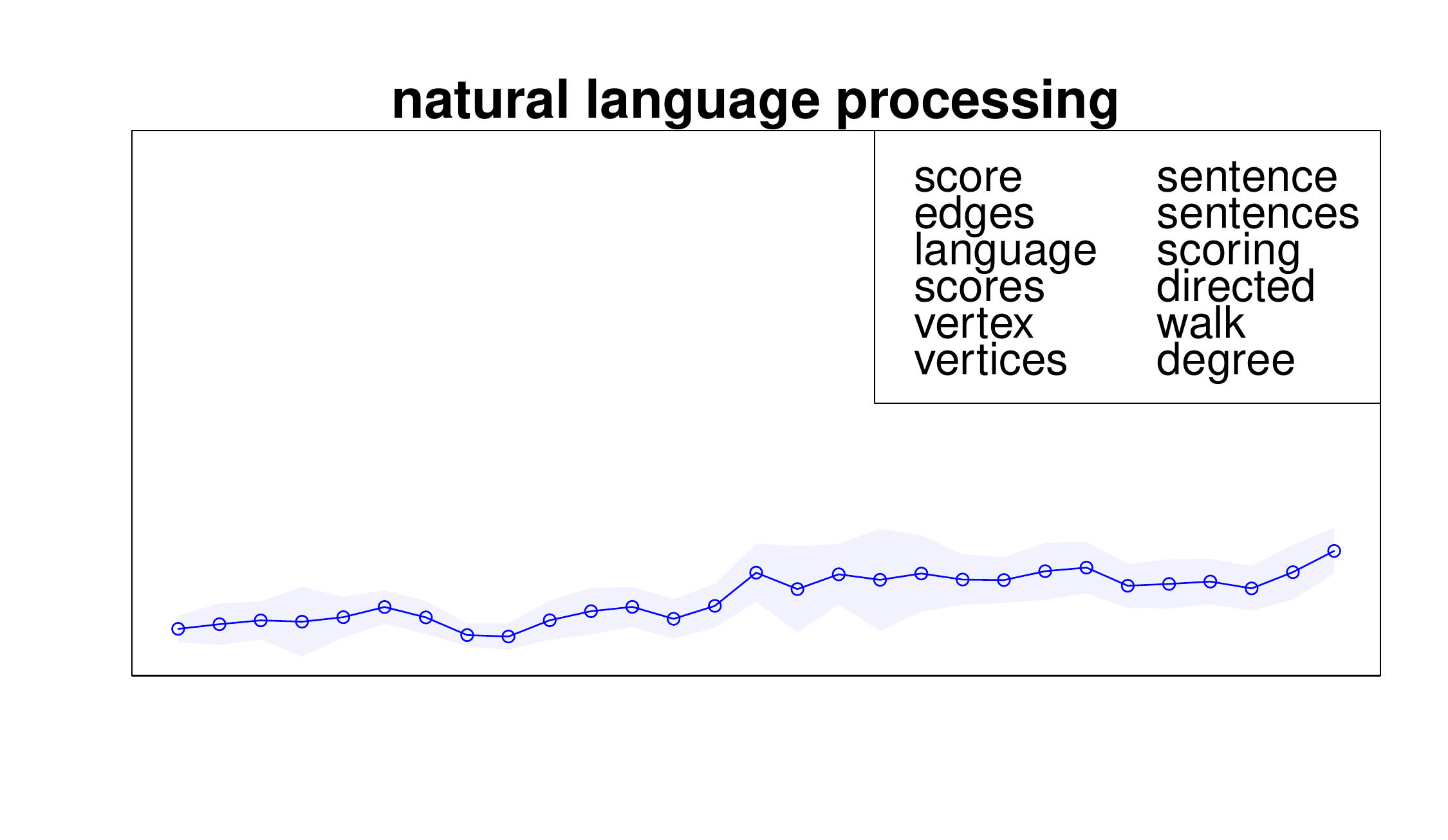}
\end{subfigure}
\begin{subfigure}{.3243\textwidth}
\centering
\includegraphics[height=4.4cm,width=1.18\textwidth]{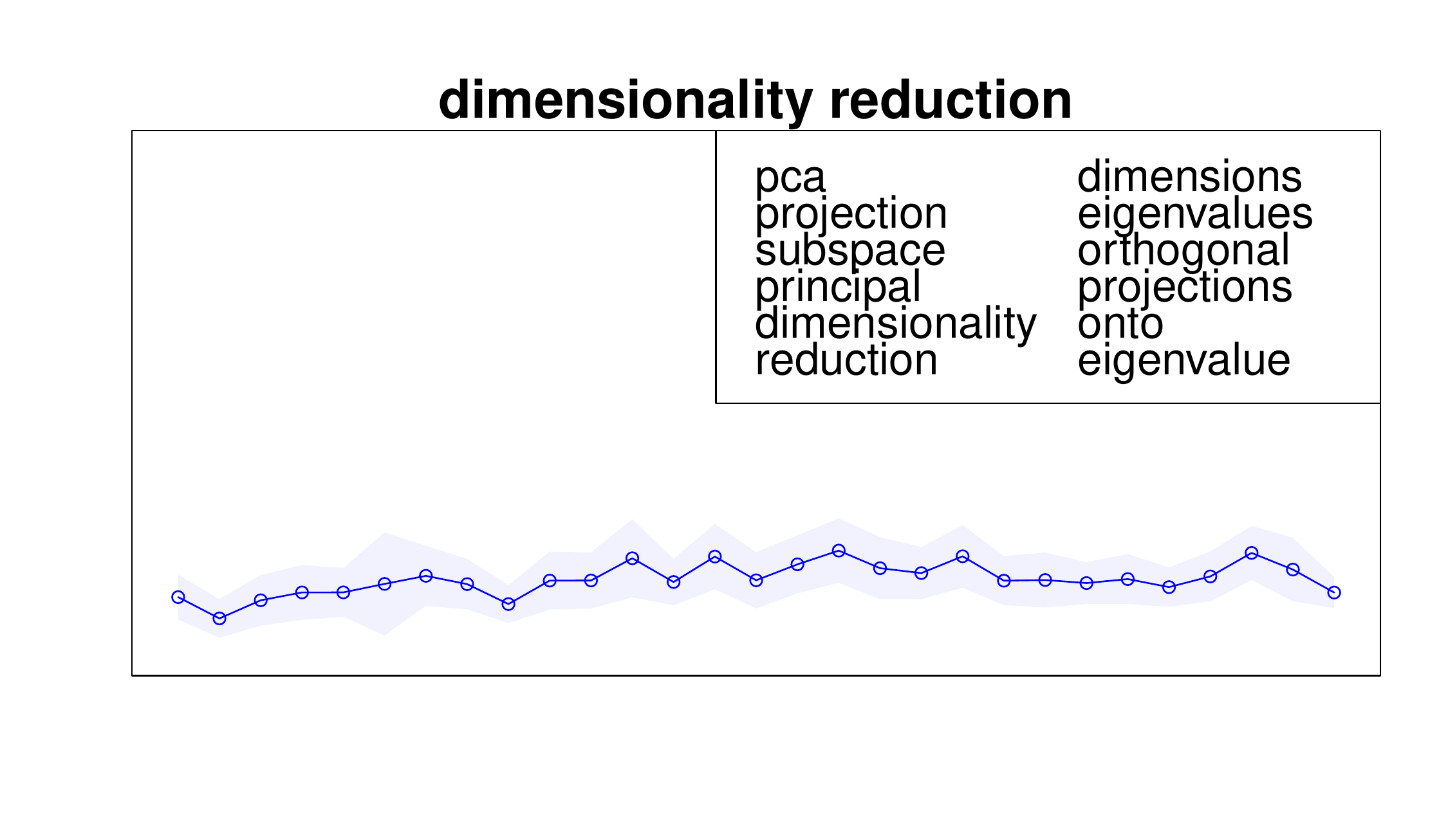}
\end{subfigure}

\vspace{-0.4in}

\centering
\begin{subfigure}{.3243\textwidth}
\centering
\includegraphics[height=4.4cm,width=1.18\textwidth]{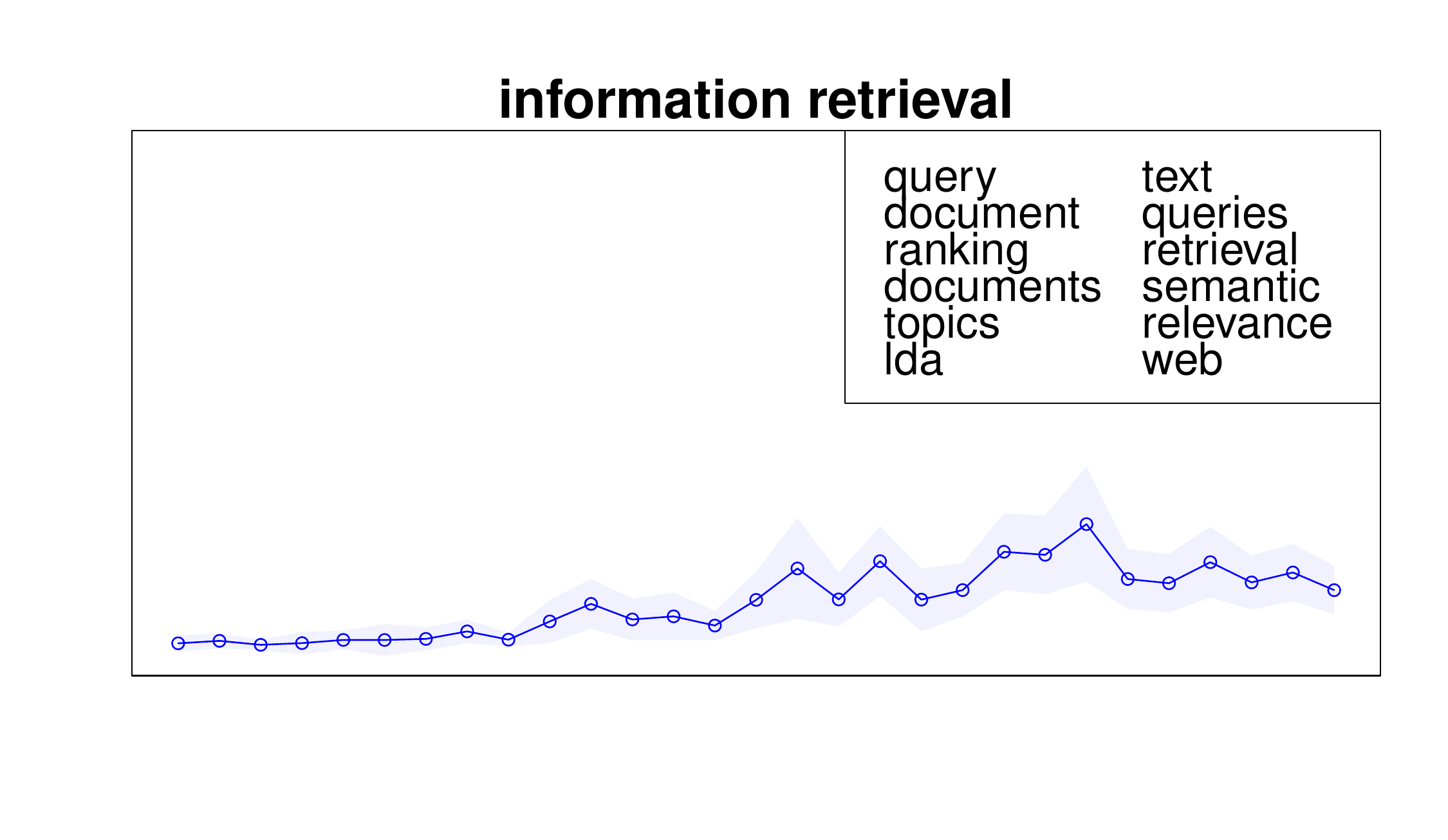}
\end{subfigure}
\begin{subfigure}{.3243\textwidth}
\centering
\includegraphics[height=4.4cm,width=1.18\textwidth]{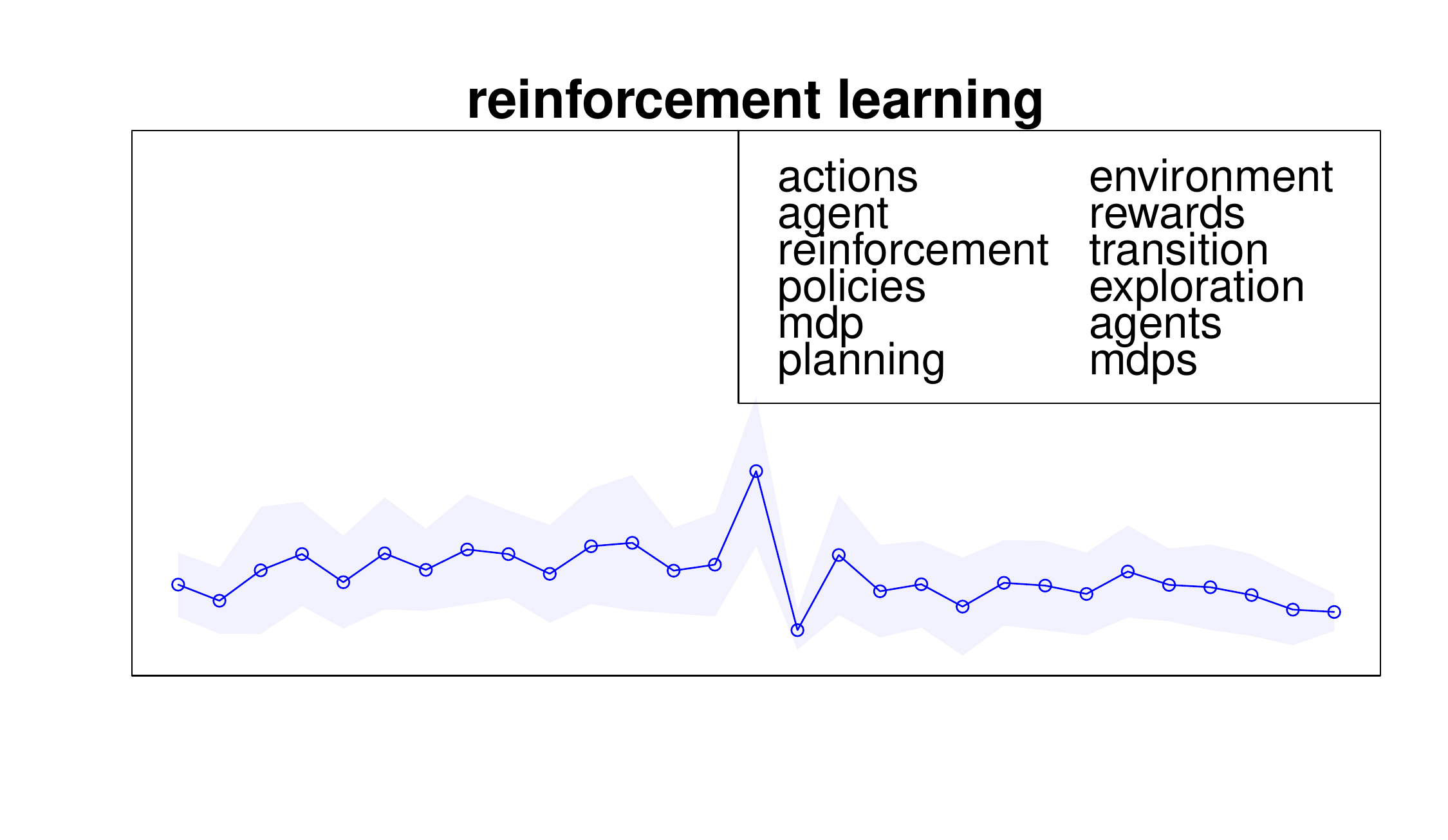}
\end{subfigure}
\begin{subfigure}{.3243\textwidth}
\centering
\includegraphics[height=4.4cm,width=1.18\textwidth]{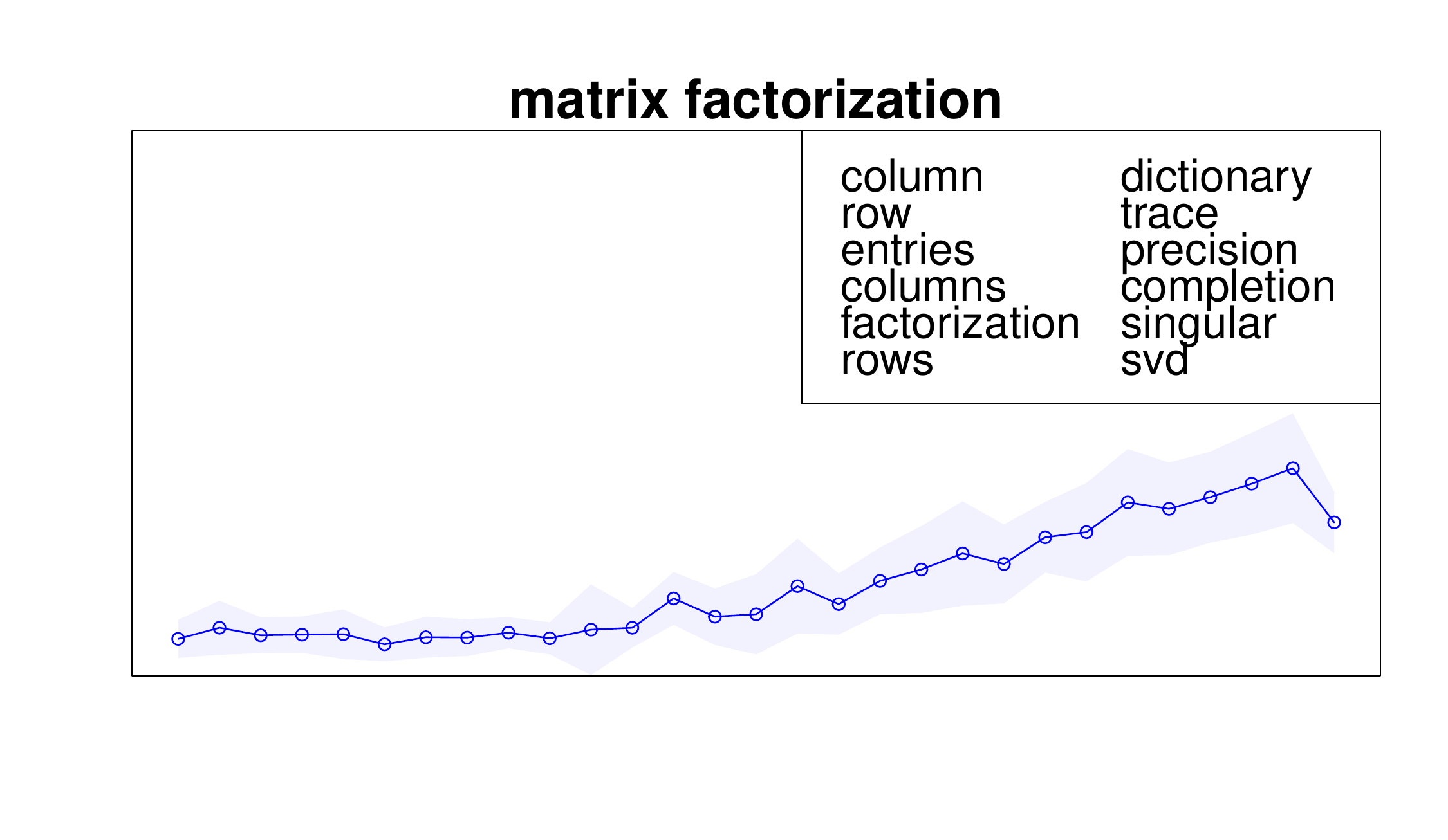}
\end{subfigure}

\end{figure*}

\vspace{-0.4in}

\begin{figure*} \ContinuedFloat
\vspace{-0.7in}
\begin{subfigure}{.3243\textwidth}
\centering
\includegraphics[height=4.4cm,width=1.18\textwidth]{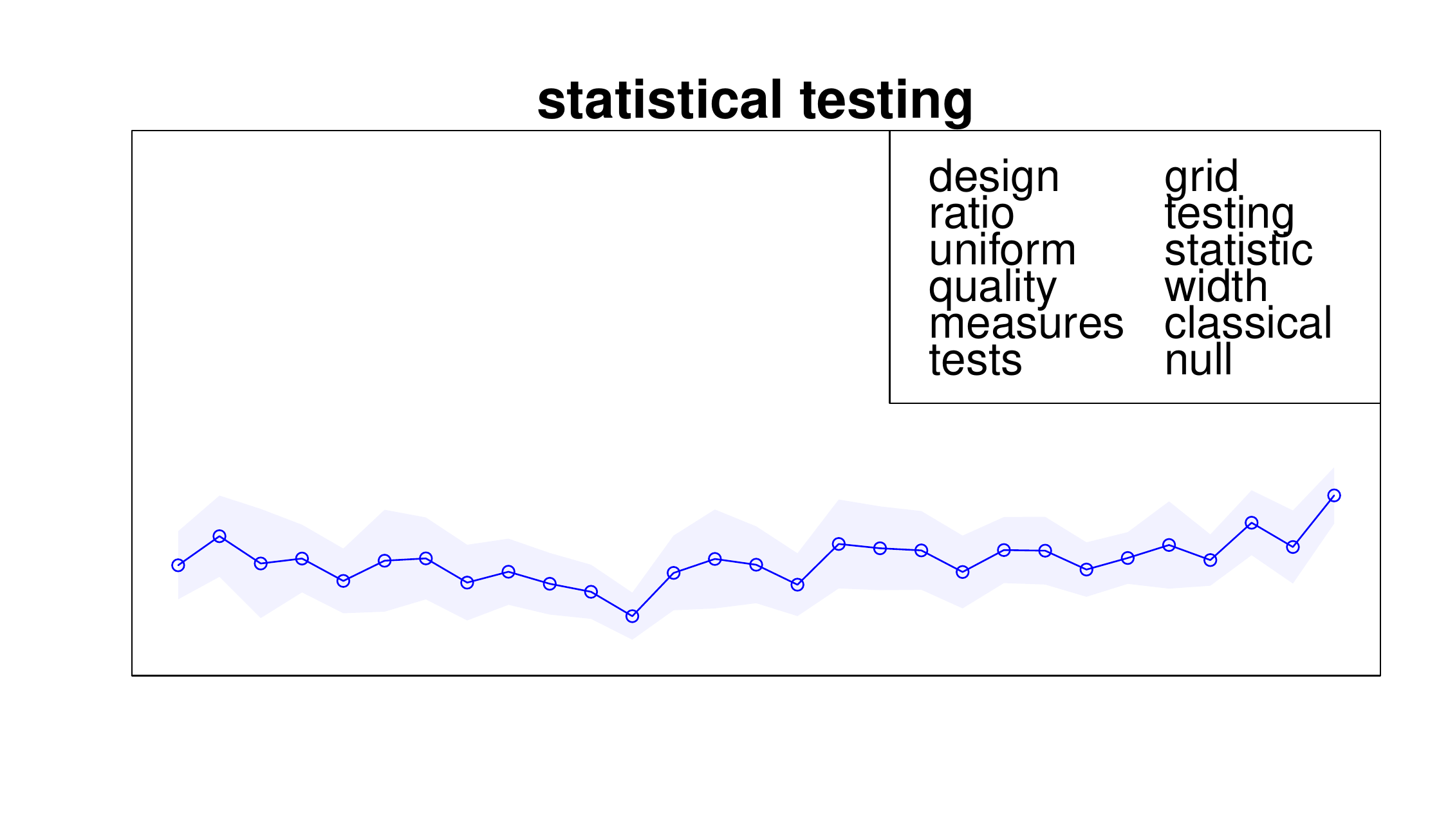}
\end{subfigure}
\begin{subfigure}{.3243\textwidth}
\centering
\includegraphics[height=4.4cm,width=1.18\textwidth]{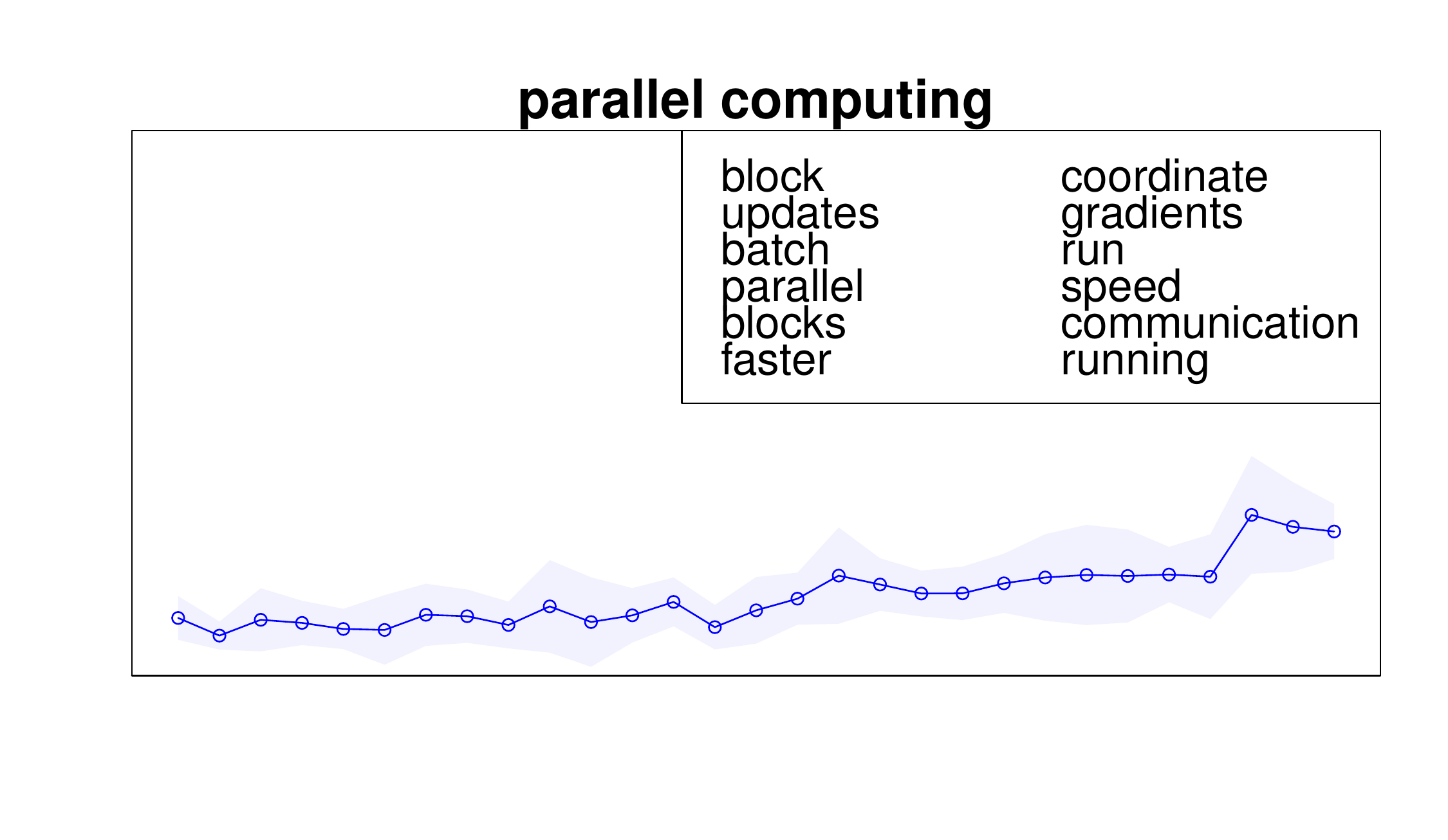}
\end{subfigure}
\begin{subfigure}{.3243\textwidth}
\centering
\includegraphics[height=4.4cm,width=1.18\textwidth]{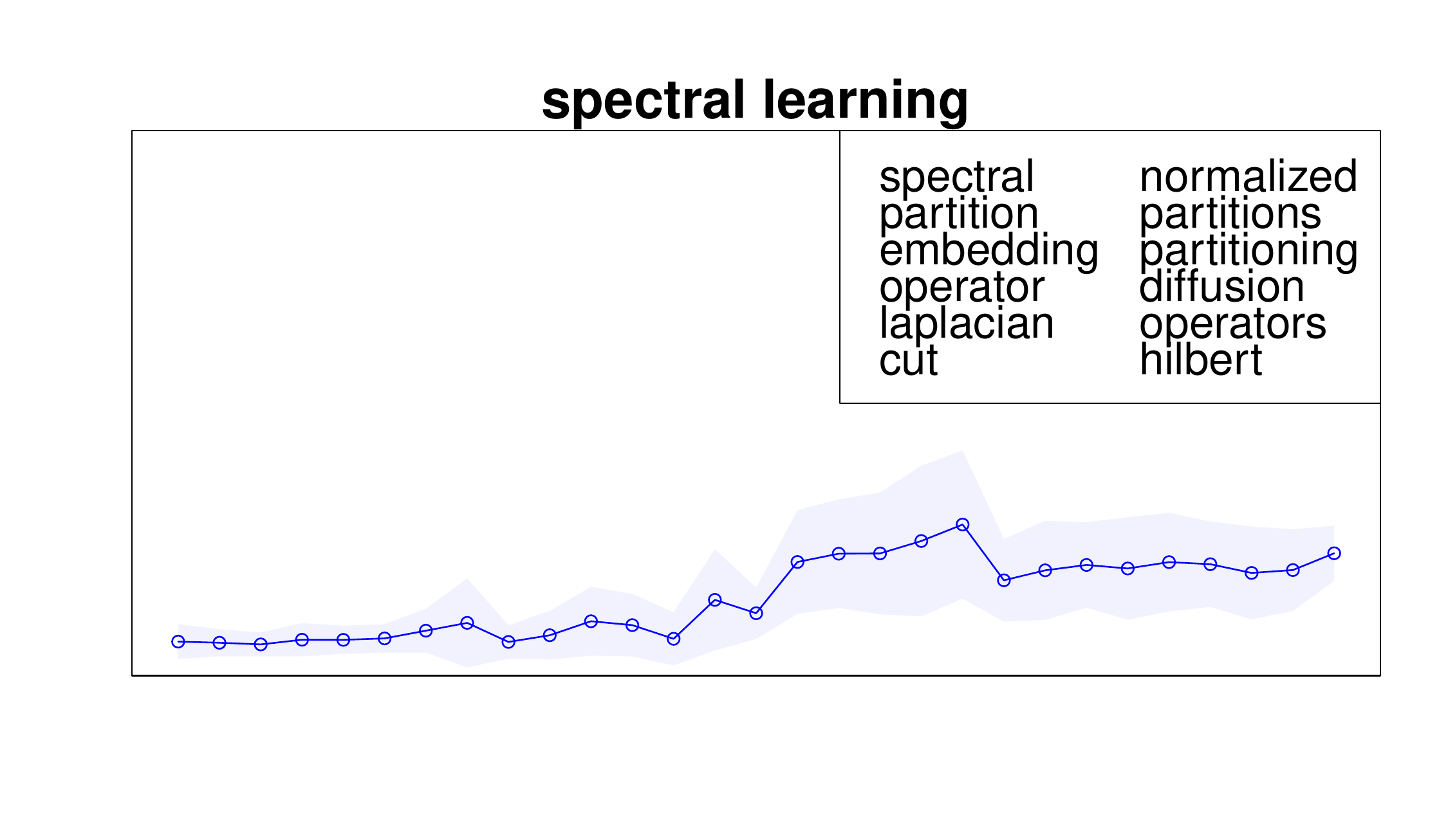}
\end{subfigure}

\vspace{-0.4in}

\begin{subfigure}{.3243\textwidth}
\centering
\includegraphics[height=4.4cm,width=1.18\textwidth]{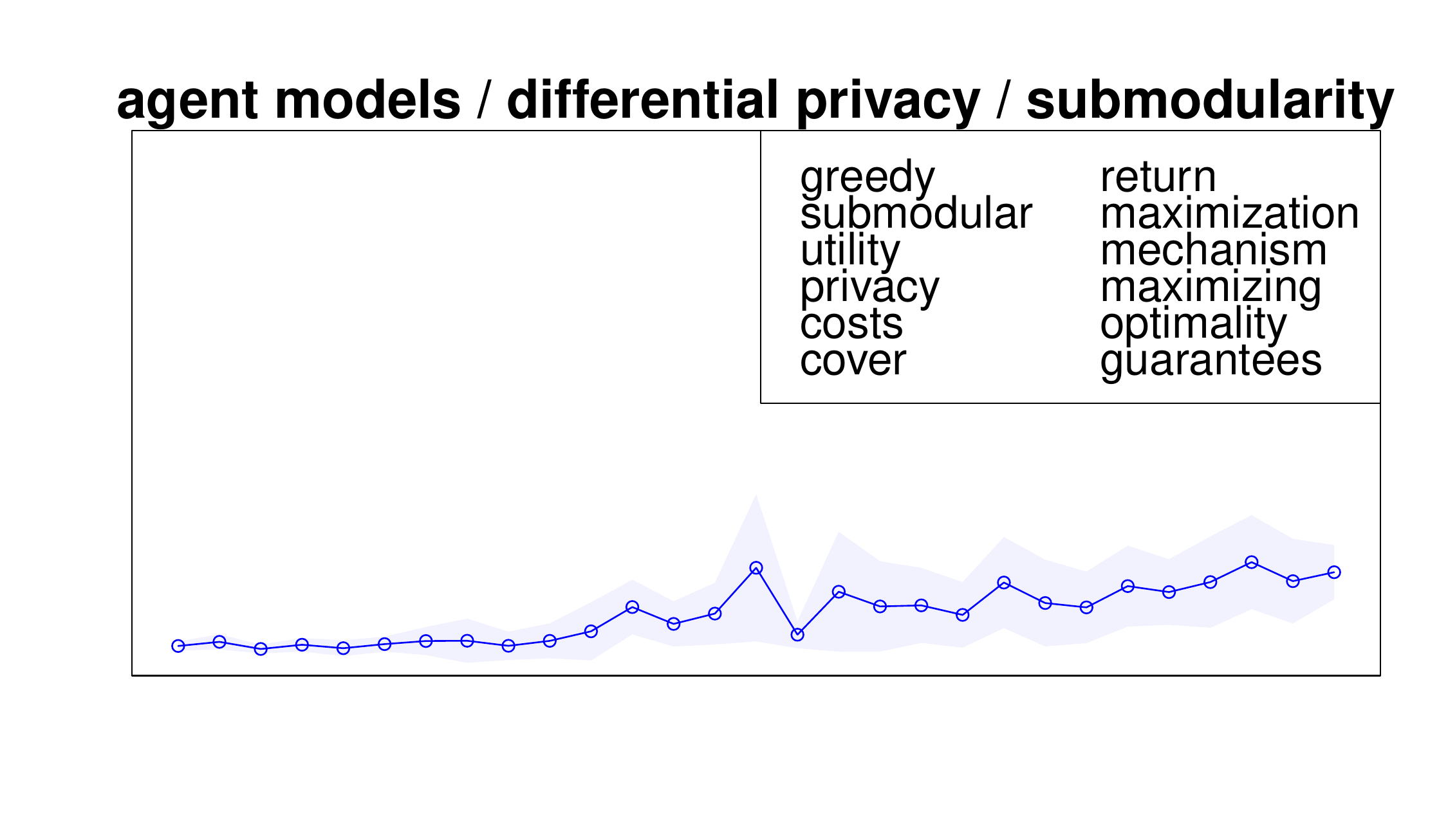}
\end{subfigure}
\begin{subfigure}{.3243\textwidth}
\centering
\includegraphics[height=4.4cm,width=1.18\textwidth]{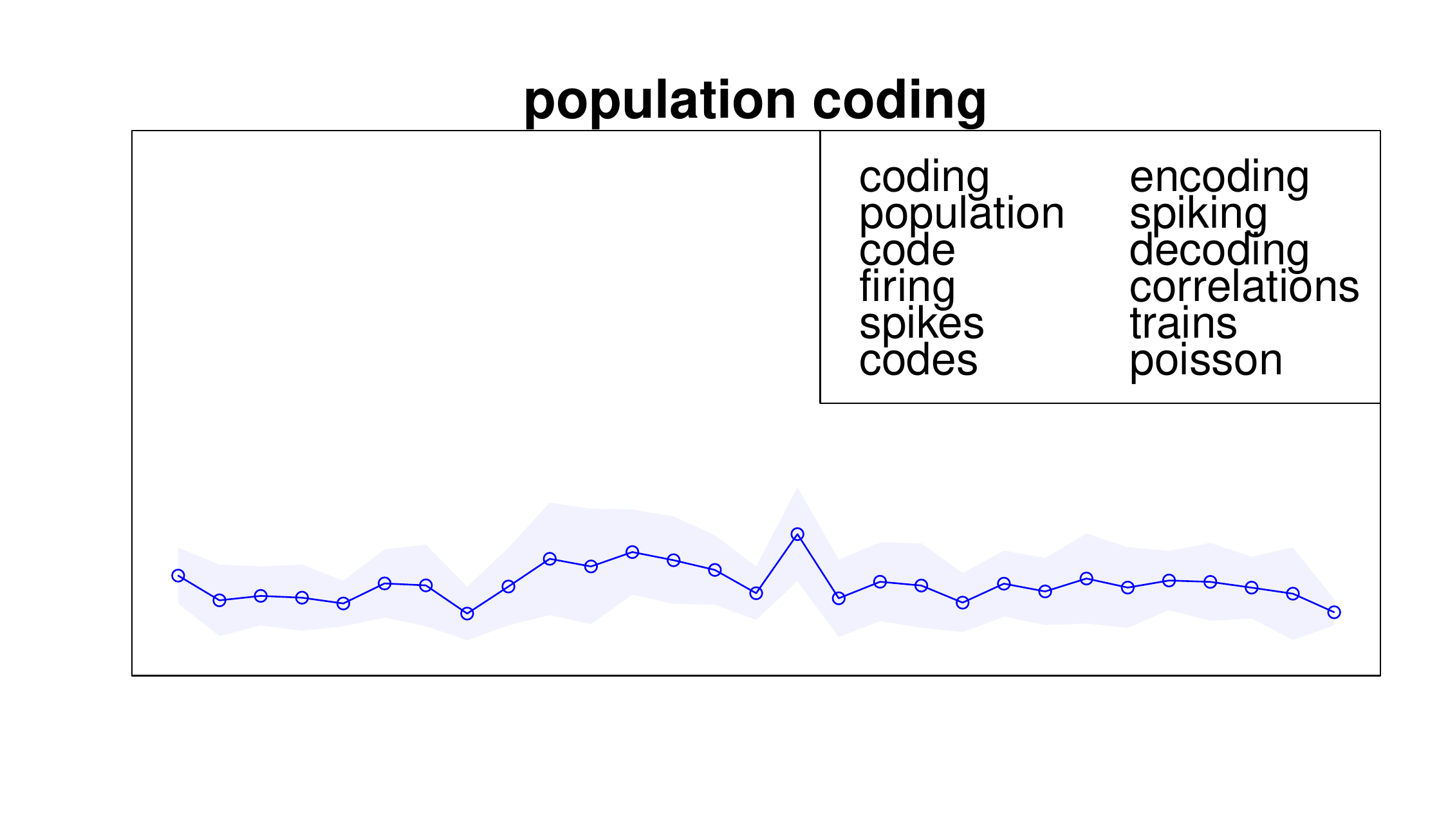}
\end{subfigure}
\begin{subfigure}{.3243\textwidth}
\centering
\includegraphics[height=4.4cm,width=1.18\textwidth]{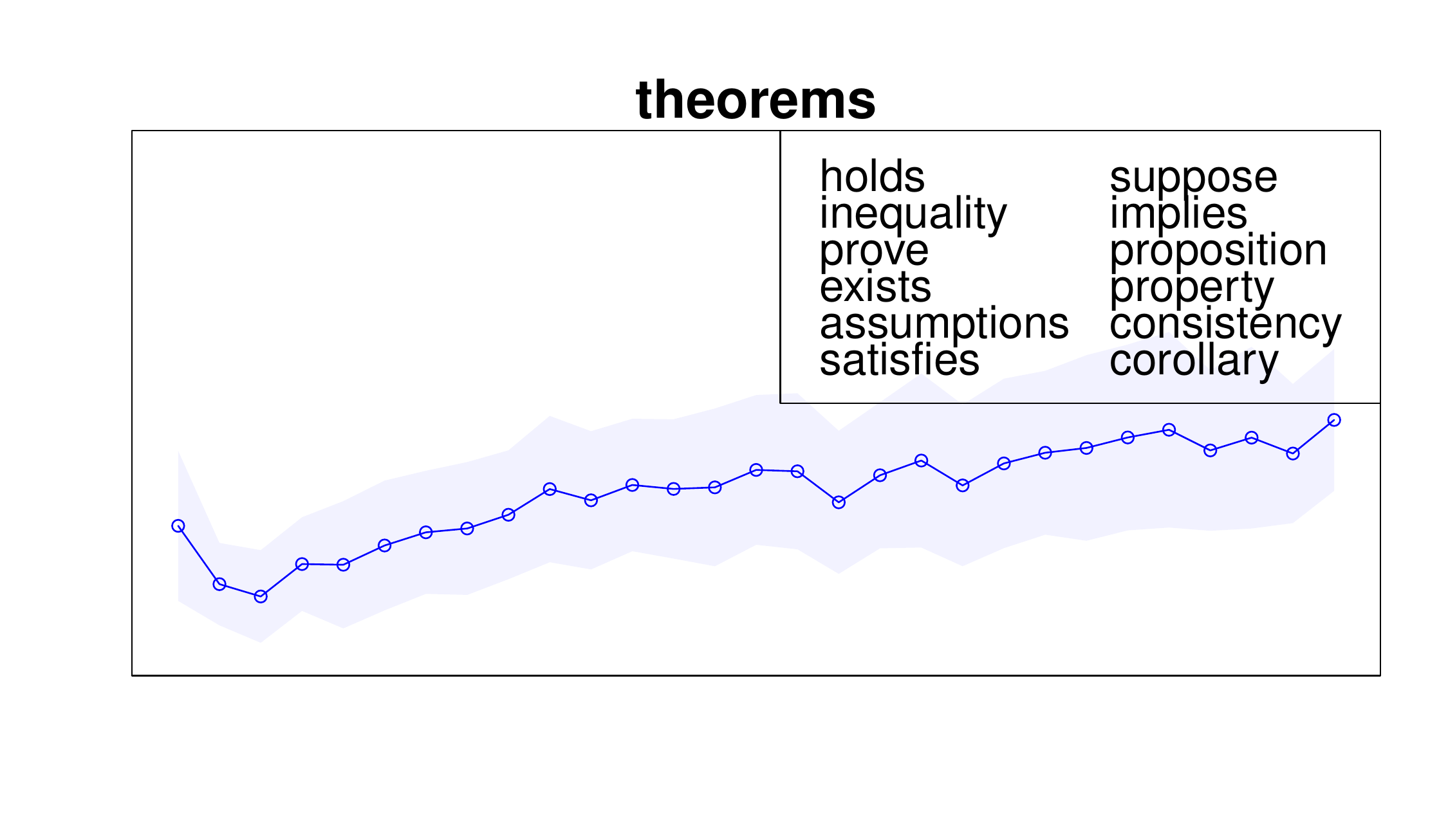}
\end{subfigure}

\vspace{-0.4in}

\begin{subfigure}{.3243\textwidth}
\centering
\includegraphics[height=4.4cm,width=1.18\textwidth]{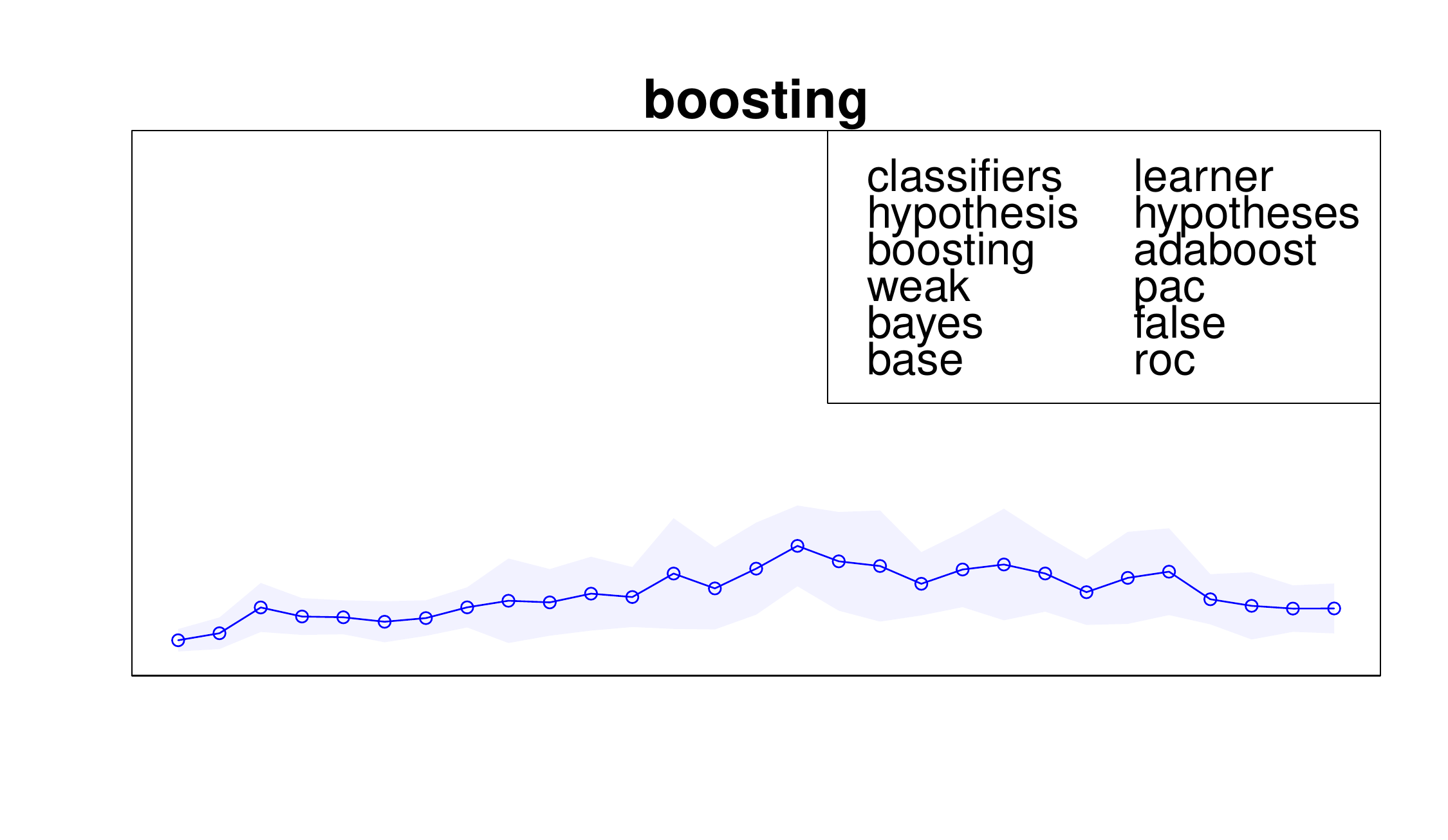}
\end{subfigure}
\begin{subfigure}{.3243\textwidth}
\centering
\includegraphics[height=4.4cm,width=1.18\textwidth]{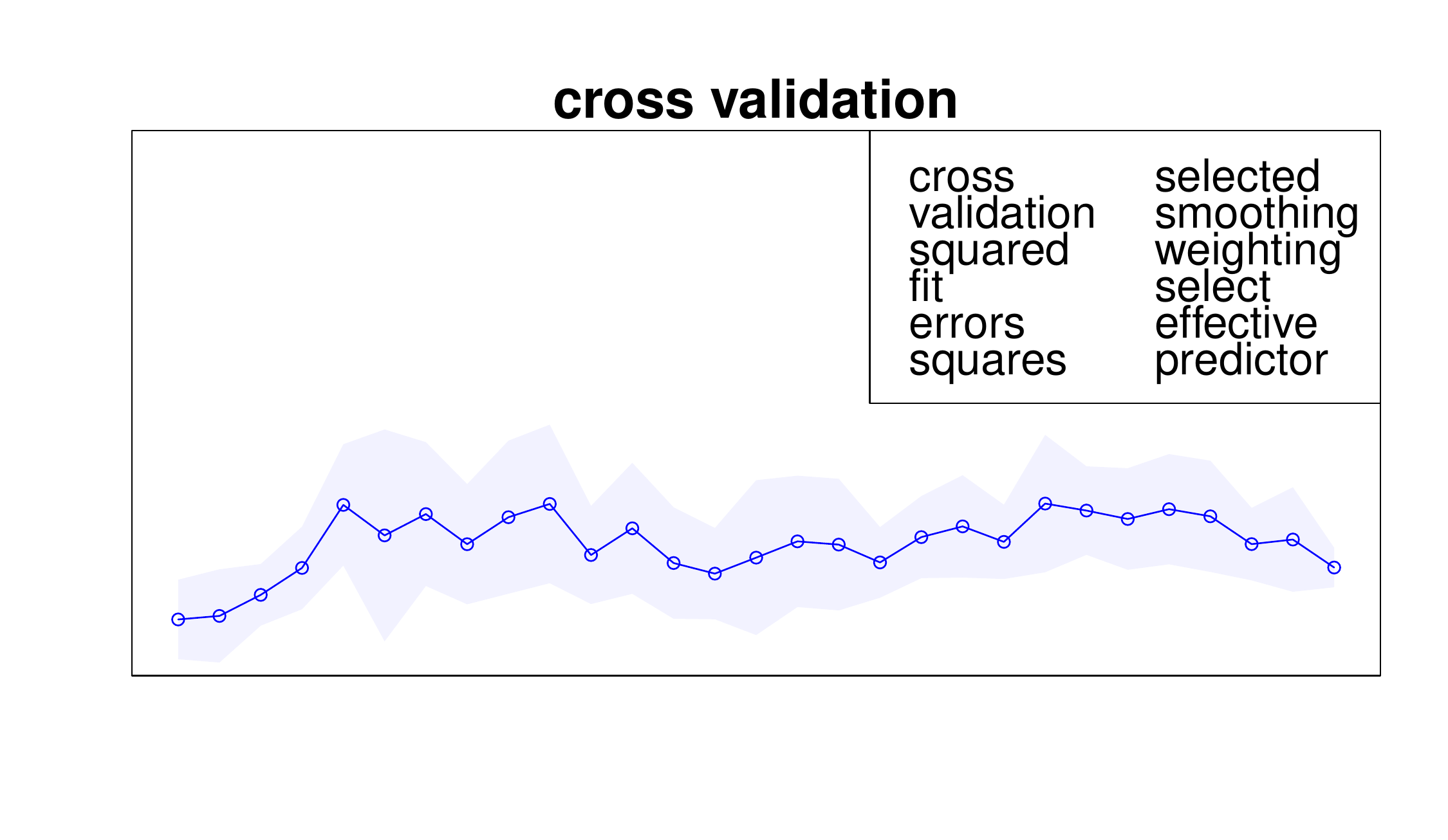}
\end{subfigure}
\begin{subfigure}{.3243\textwidth}
\centering
\includegraphics[height=4.4cm,width=1.18\textwidth]{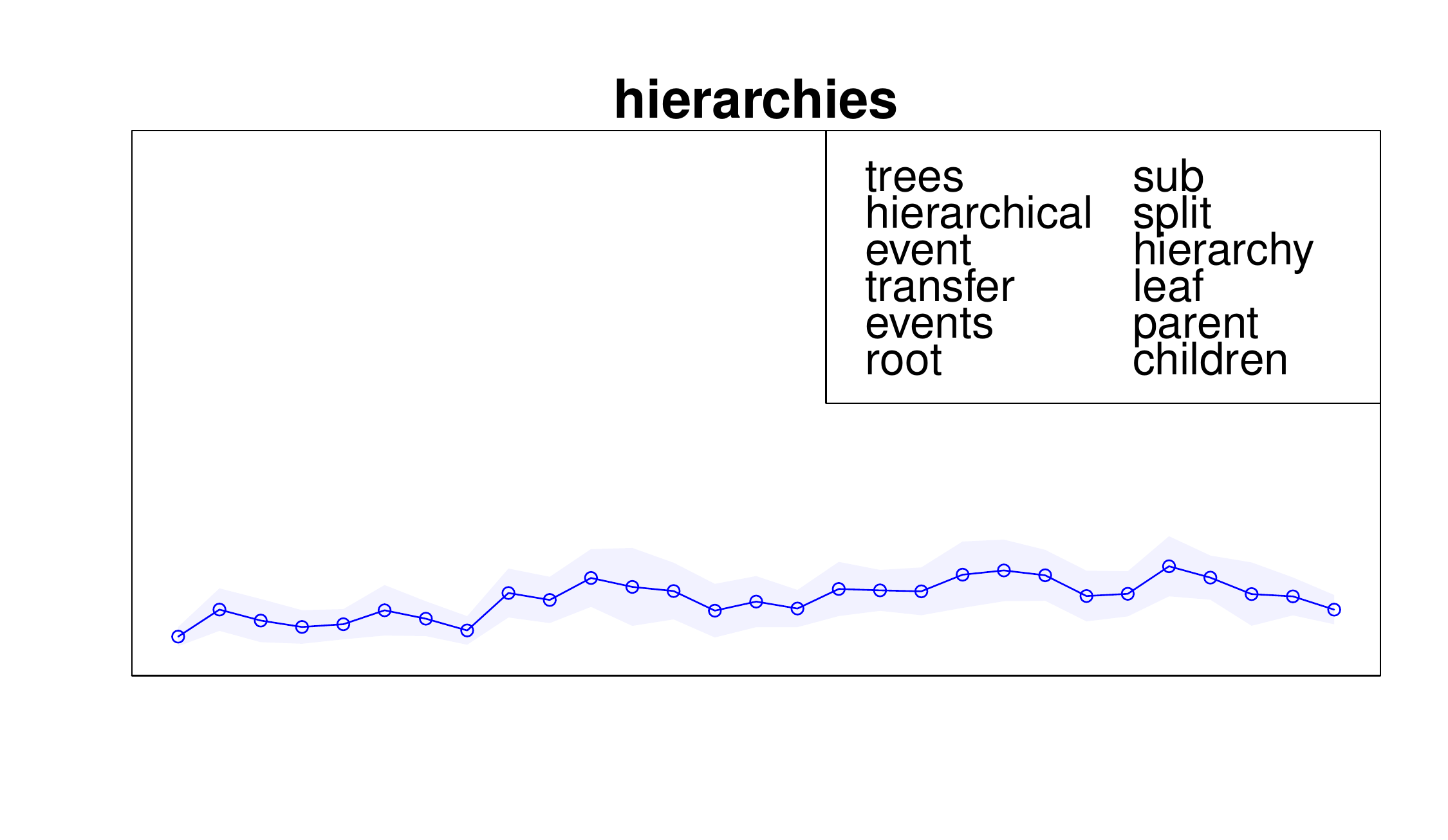}
\end{subfigure}

\vspace{-0.4in}

\begin{subfigure}{.3243\textwidth}
\centering
\includegraphics[height=4.4cm,width=1.18\textwidth]{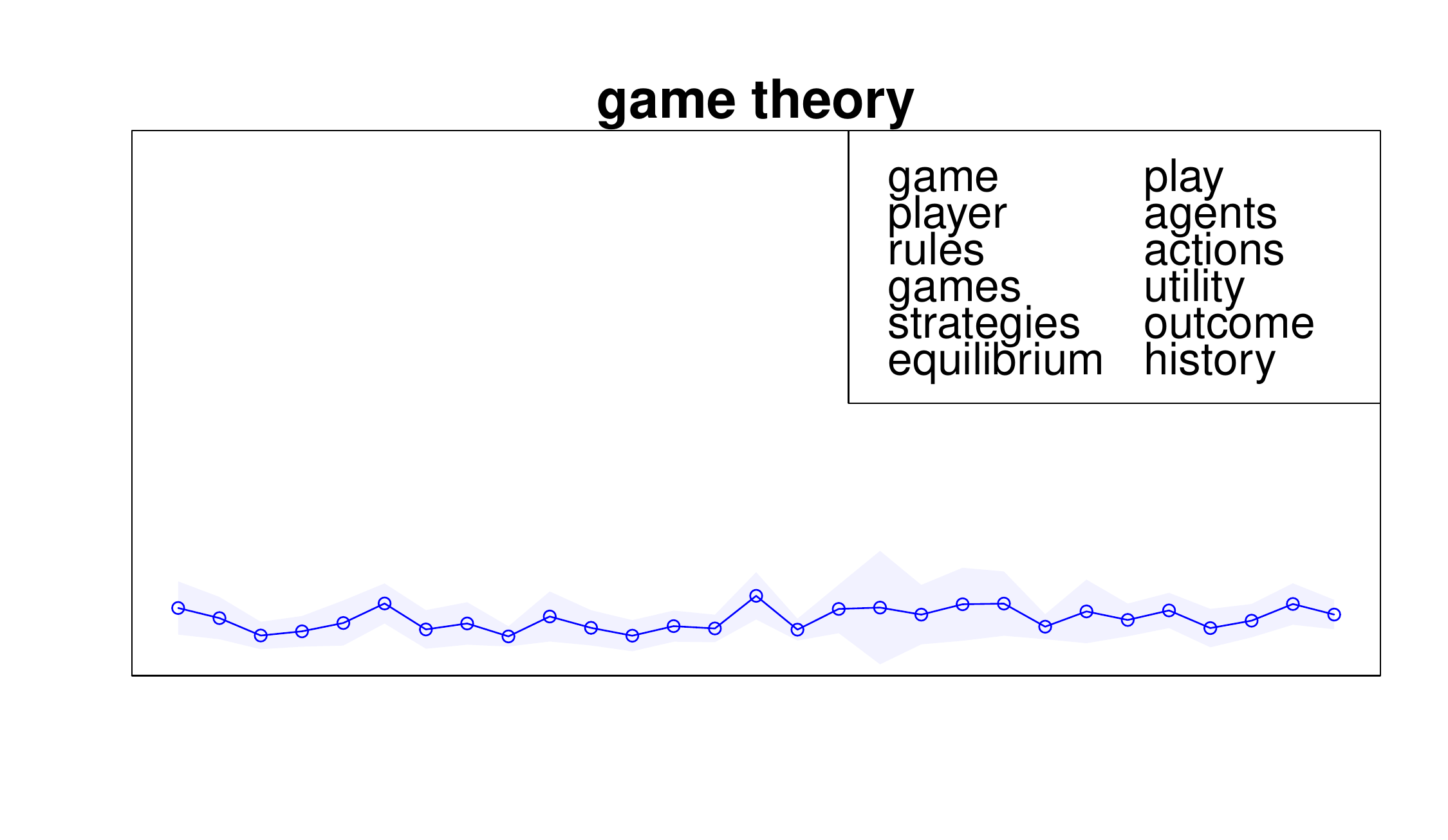}
\end{subfigure}
\begin{subfigure}{.3243\textwidth}
\centering
\includegraphics[height=4.4cm,width=1.18\textwidth]{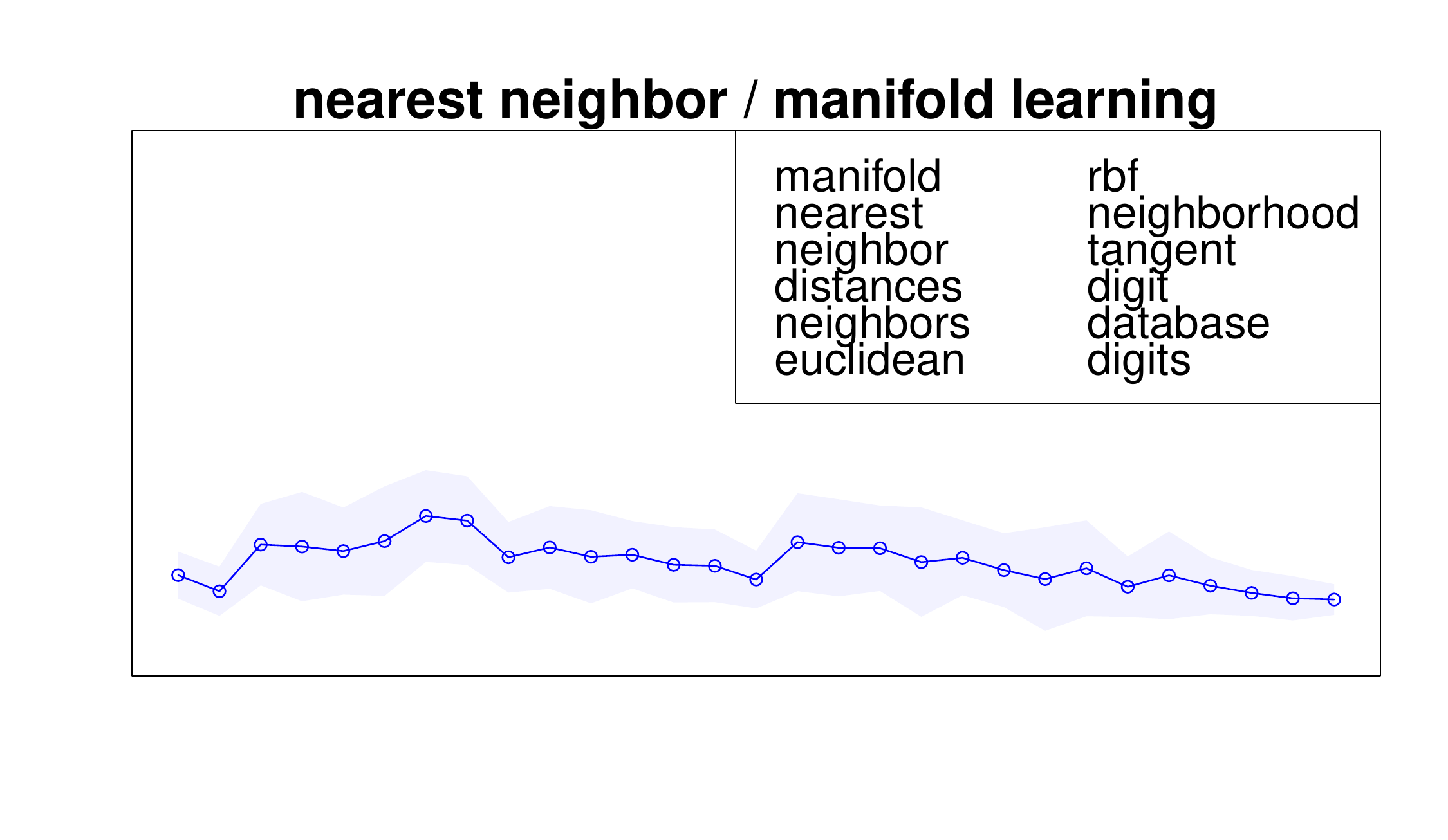}
\end{subfigure}
\begin{subfigure}{.3243\textwidth}
\centering
\includegraphics[height=4.4cm,width=1.18\textwidth]{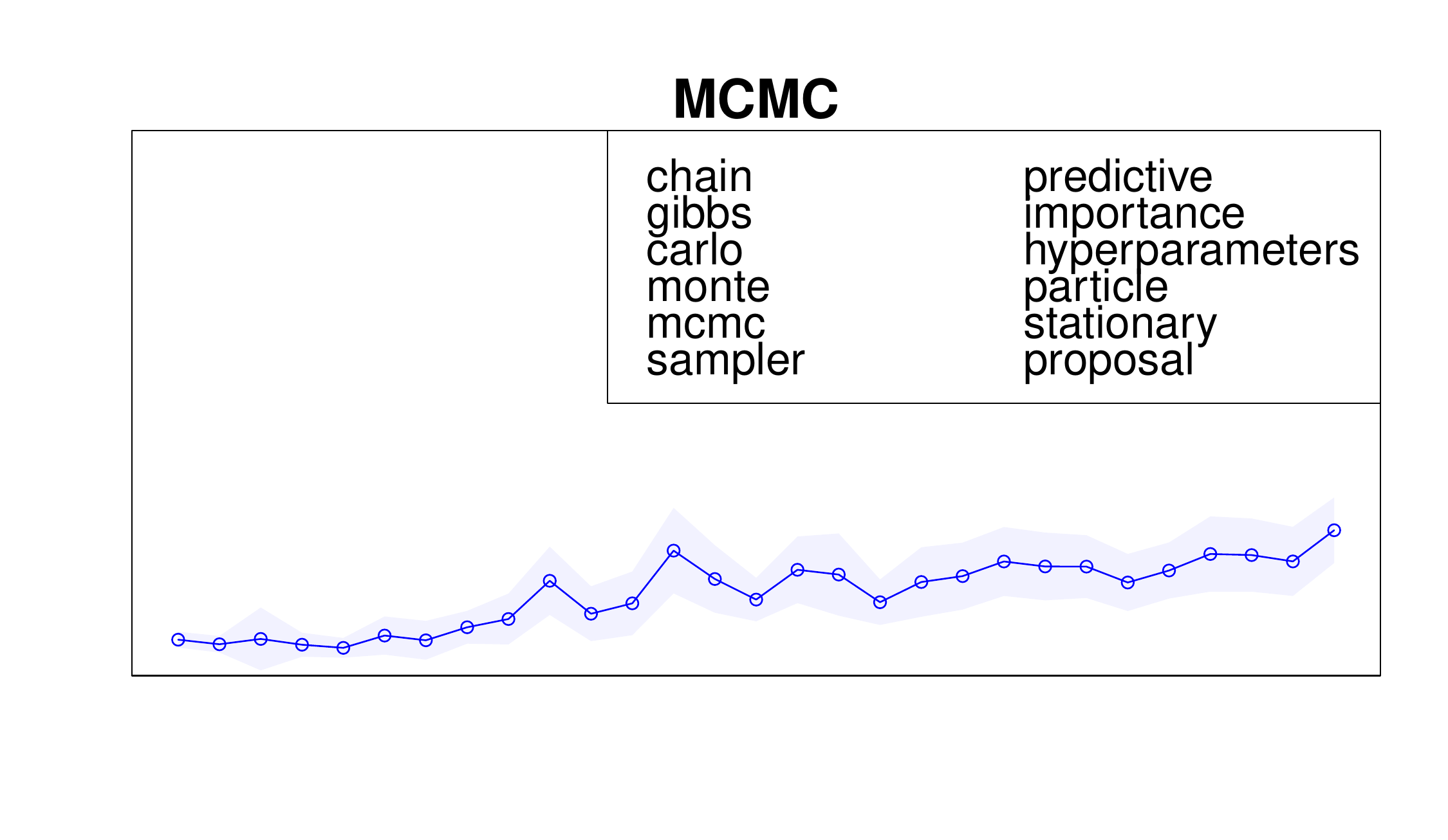}
\end{subfigure}

\vspace{-0.4in}

\begin{subfigure}{.3243\textwidth}
\centering
\includegraphics[height=4.4cm,width=1.18\textwidth]{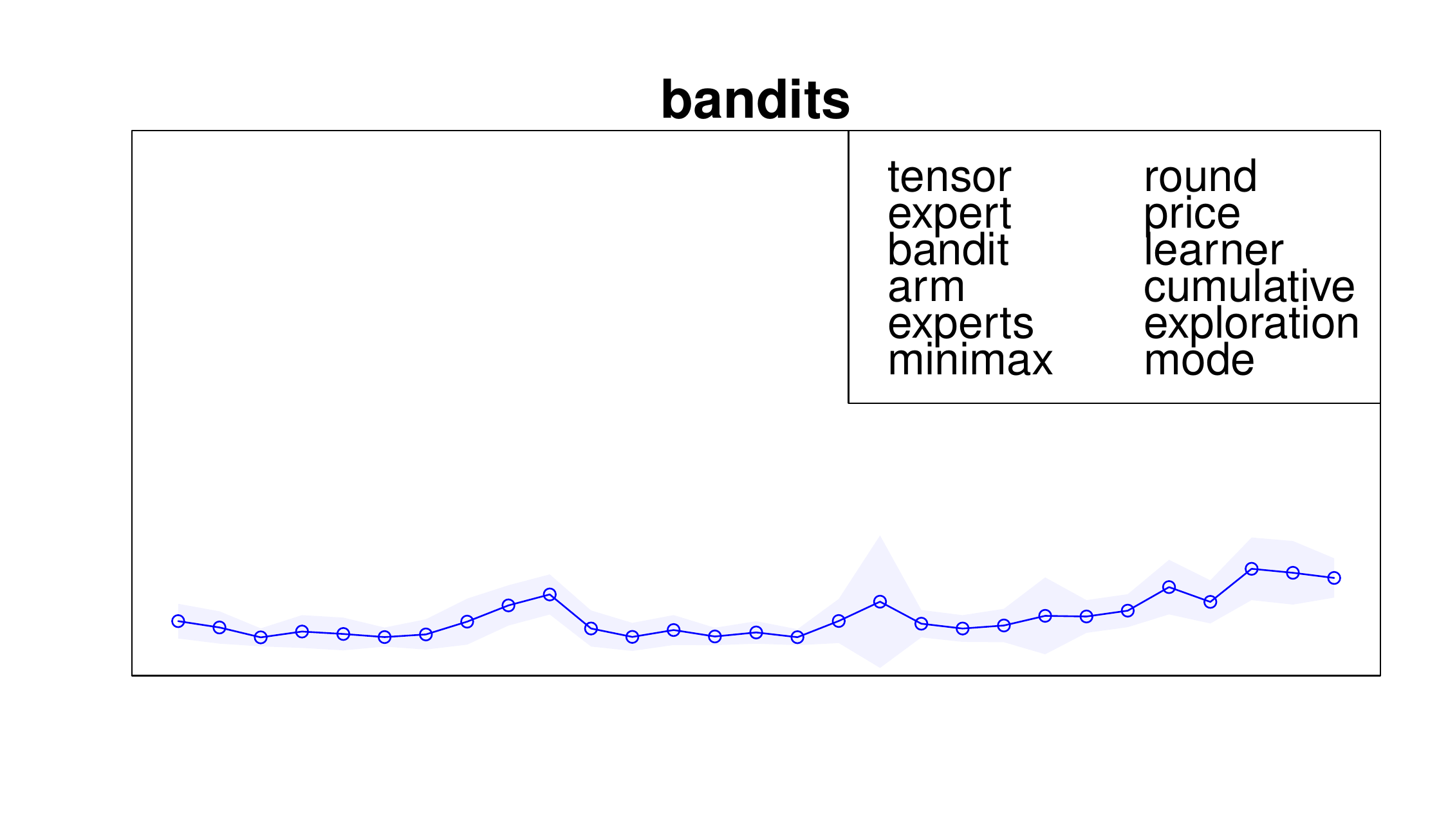}
\end{subfigure}
\begin{subfigure}{.3243\textwidth}
\centering
\includegraphics[height=4.4cm,width=1.18\textwidth]{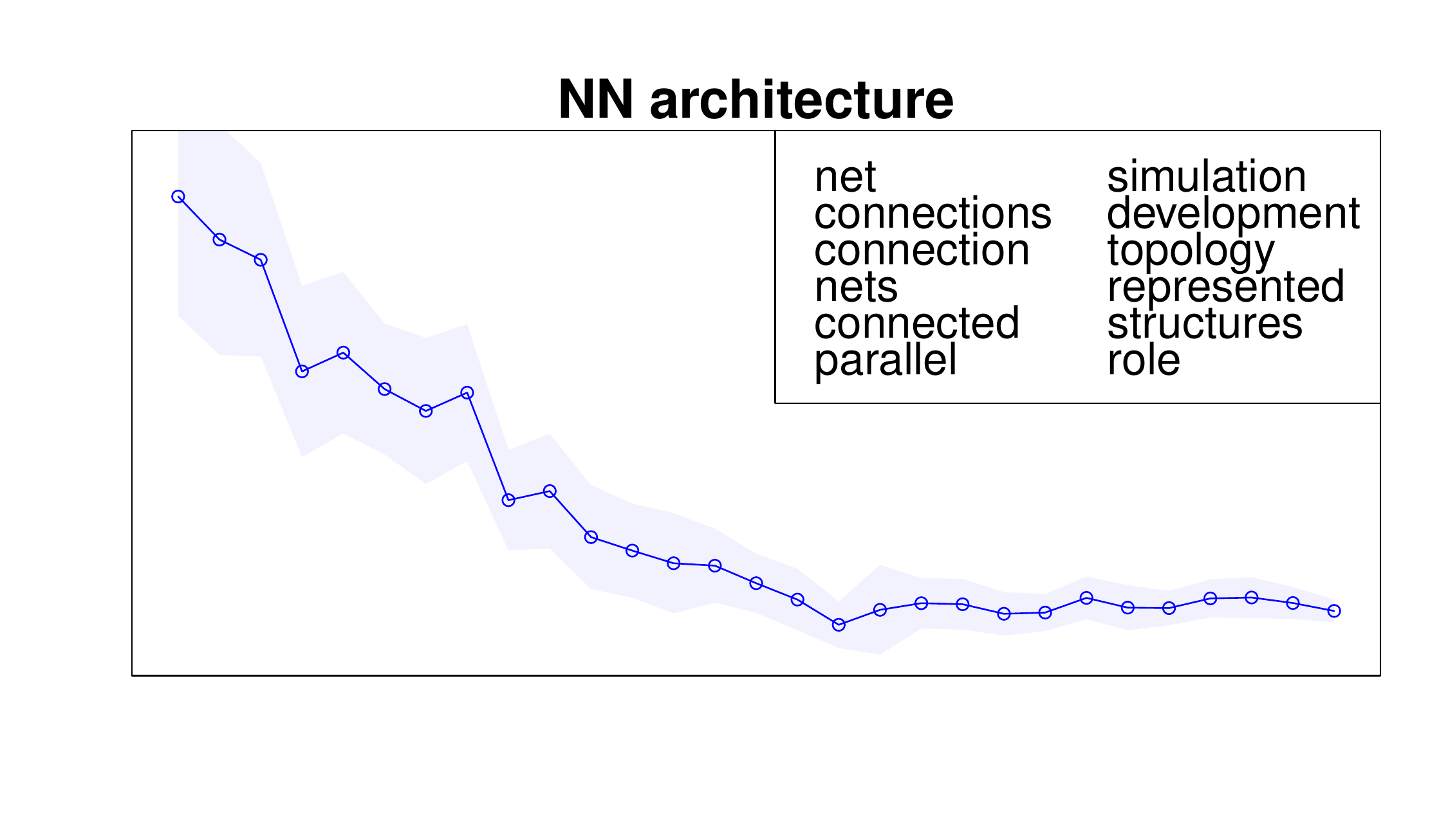}
\end{subfigure}
\begin{subfigure}{.3243\textwidth}
\centering
\includegraphics[height=4.4cm,width=1.18\textwidth]{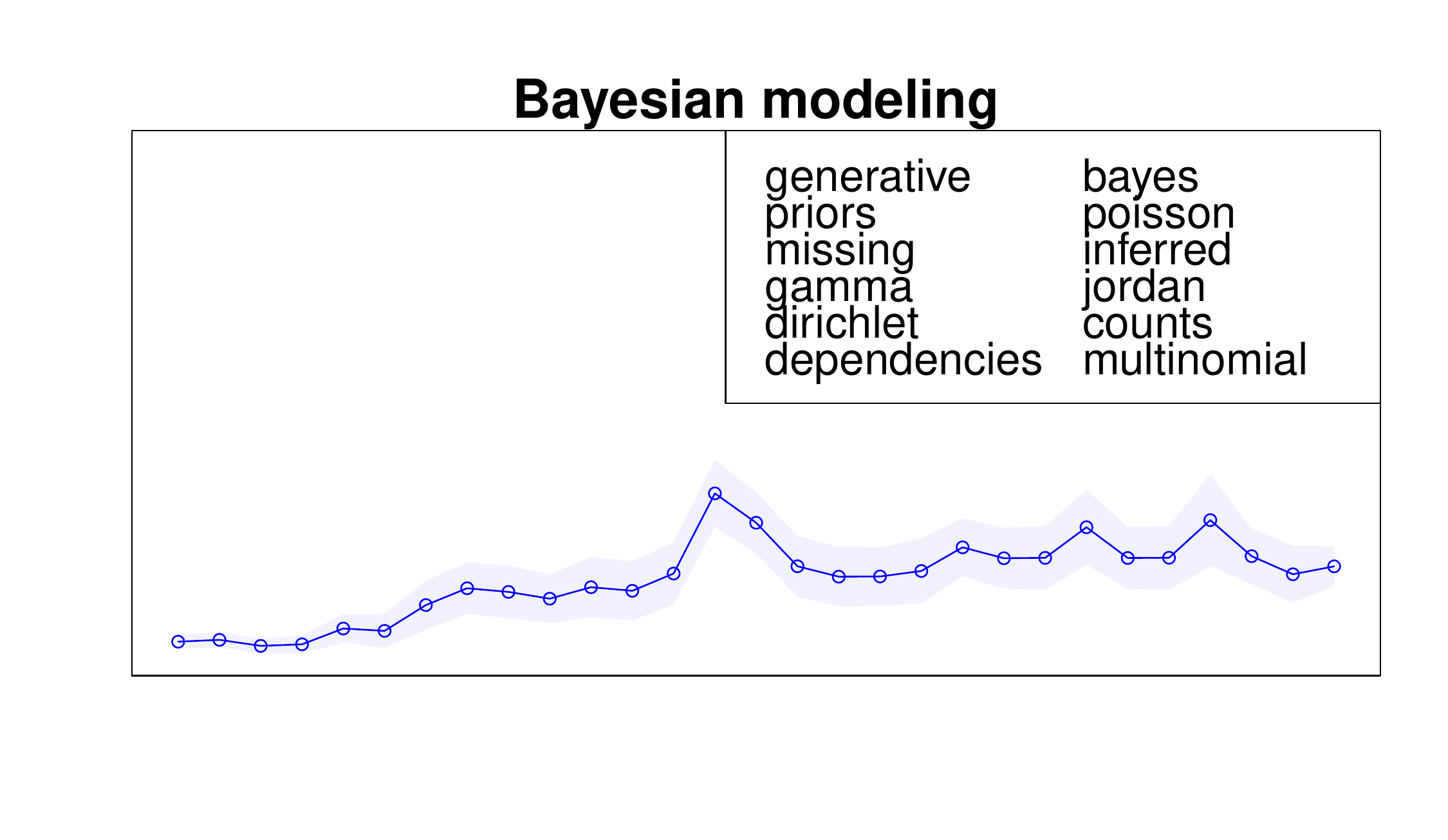}
\end{subfigure}

\caption{Posterior topic proportions over the years 1987-2015 and 12 most likely words for each topic (NIPS data set).}
\label{fig:individual_topics}
\end{figure*}

\begin{comment}
\begin{table}[ht!]
\centering
\caption{Ten most likely words for a subset of the topics found in the NIPS data set.}
\label{table:words}
\begin{tabular}{ | l | l | }
\hline
  & 10 most likely words \\ \hline
neural networks (NN) &unit - net - architecture - activation - inputs \\ 
& recurrent - sequence - back - propagation - trained \\ \hline
probabilistic modeling &distribution - gaussian - probability - parameters - density \\ 
& likelihood - mixture - log - variables - bayesian \\ \hline
neuroscience &neurons - neuron - cell - activity - synaptic  \\ 
& response - firing - spike - cells - stimulus \\ \hline
circuits &analog - parallel - circuit - hopfield - threshold  \\ 
& bit - voltage - current - design - circuits \\ \hline
speech &recognition - speech - classification - classifier - word  \\ 
& test - words - trained - classifiers - rbf\\ \hline
image &image - feature - features - object - images  \\ 
& objects - map - recognition - representation - view \\ \hline

\end{tabular}
\end{table}
\end{comment}

\section{Conclusion}
We have presented a new framework for generating dependent IBPs by means of a novel time-evolving beta process. The key insight has been building on the PRF from population genetics to derive a suitable model for the prevalence and evolution of features over continuous time. We have developed an interesting MCMC framework for exact posterior inference with this model and presented an alternative truncated scheme where the number of features is fixed.

As an application of the WF-IBP, we have described a new time-dependent nonparametric topic modeling scheme that builds on the model of \cite{ibpc}. The WF-IBP topic model allows for a flexible evolution of the popularity of topics over time, keeps inference simple and, compared to HDP-based models, has the benefit of decoupling the probability of topics and their proportion within documents. We have used our model to explore the data set consisting of the full text of NIPS conference papers from 1987 to 2015 and obtained an interesting visualization of how the popularity of the underlying topics evolved over these 29 years. In addition, test-set perplexity results have shown that incorporating time also improves on the predictive performance of the model.

A number of directions for future work are open. Firstly, as $K \to \infty$ the fixed-$K$ truncation marginally converges to the infinite model and simulations showed that their dynamics are remarkably similar. Further work could formally investigate how the dynamics of the fixed-$K$ truncation approximates the infinite model. Secondly, the current MCMC framework could be generalized to include inference of the hyperparameters $\alpha$, $\beta$ and the time step of the W-F diffusion.

Finally, an extension of this work could modify the PRF by letting features evolve according to a more general W-F diffusion with selection and recombination, which would allow for feature-specific drifts in popularity and the coupled evolution of different features, respectively. 

\section*{Acknowledgement}
Valerio Perrone is supported by EPSRC [EP/L016710/1]. Paul Jenkins is supported in part by EPSRC [EP/L018497/1]. Dario Span\`{o} is supported in part by CRiSM, an EPSRC-HEFCE UK grant. Yee Whye Teh is supported by EPSRC for research funding through grant EP/K009850/1, EP/N000188/1 and EP/K009362/1.

\newpage

\appendix
\section*{Appendix A. Proof of Theorem 1}
The mean density of a PRF with immigration parameter $\alpha$ is 
\begin{align*}
l(x)= \alpha m(x) dx,
\end{align*}
where 
\begin{align} \label{speed}
m(x) &=  \frac{e^{I(x)}}{\sigma^2(x)}, \quad \text{with } I(x) :=  \int^x_0  \frac{2 \gamma(y)}{\sigma^2(y)} dy,
\end{align}
is the speed density of the process \cite{griff}. $\gamma$ and $\sigma$ are the drift and diffusion terms as defined in (\ref{drift}) and (\ref{diffusion}). Plugging (\ref{drift}) with $\mu=0$ and (\ref{diffusion}) into the integral, we have
\begin{align*}
I(x) &= %\int^x_0  \frac{- \beta y}{y(1-y)} dy = \int^x_\eta   \frac{\beta}{y-1} dy =
\beta  \ln (1-x) % - \beta  \ln (1-\eta),
\end{align*}
so that
\begin{align*}
m(x) &= \frac{ (1-x)^{\beta}  %(1-\eta)^{-\beta}
}{x(1-x)} =  x^{-1} (1-x)^{\beta-1}. %(1-\eta)^{-\beta}.
\end{align*}

It follows that
\begin{align*}
l(x)= \alpha m(x) dx &=   \alpha x^{-1} (1-x)^{\beta-1} dx
\end{align*}
is the resulting mean density, which completes the proof.

\begin{remark}
When $\mu, \beta=0$ it is necessary to condition each diffusion on hitting the boundary $0$ before $1$ in reverse time, which leads to an extra term in the analogous result in \cite{sawyer1992}.
\end{remark}
\section*{Appendix B. Derivation of full conditionals}
As observed in \cite{focused}, in this topic modeling setting there are two equivalent ways of generating documents. Either the total number of words is sampled from a negative binomial NB($\sum_k z_{ikt} \phi_{kt}, 1/2)$ and then the topic and word assignments are drawn, or the number of words generated by each topic is drawn from NB($ z_{ikt}\phi_{kt}, 1/2)$ and then the word assignments are picked. Following \cite{focused} closely, we make use of the latter construction to derive the full conditional distributions for the Gibbs sampler of the WF-IBP topic model.
\paragraph{Full conditional of $a_{il}$}
Recall that $a_{ilt}=k$ indicates that the $l$th word in document $i$ at time $t$ is assigned to topic $k$. We have
\begin{align*}
 p(a_{ilt}=k  \mid  a_{-il},Z_t, w_{ilt}, \phi_t) \nonumber    
& \propto  p(w_{ilt}  \mid  a_{ilt}=k)  p(a_{ilt}=k  \mid  a_{-il},z_{ikt}, \phi_{kt}) \nonumber \\
& \propto  p(w_{ilt}  \mid  a_{ilt}=k) (n^i_{kt} + \phi_{kt} z_{ikt}) \nonumber, 
\end{align*}
where the last step is given by integrating out $\theta_{it}$, namely the distribution over topics in document $i$ at time $t$, and using the Dirichlet-Categorical conjugacy. Recall that $n^i_{kt}$ denotes the number of words assigned to topic $k$ in document $i$ at time $t$ and that $ n_{kt}$ denotes the total number of words assigned to topic $k$ at time $t$, both excluding the assignment $a_{il}$. Similarly, integrating out the parameter $\rho_k$ representing the distribution over words of topic $k$ and using the Dirichlet-Categorical conjugacy, we have that 
\begin{align*}
p(w_{ilt}  \mid  a_{ilt}=k)  =   \frac{(n^{w_{il}}_k + \eta)}{n_k + \eta D - 1},
\end{align*}
which, plugged into the previous equation, gives the desired full conditional.
\paragraph{Full conditional of $\phi_k$ and $\gamma$}
We have that
\begin{align}
 p(\phi_{kt},\gamma  \mid  n_{kt},X(t),Z_t)  \nonumber  
& \propto p(\phi_{kt},\gamma, n_{kt},X(t),Z_t)  \nonumber  \\ 
& \propto  p(\phi_{kt}  \mid  \gamma) P(\gamma) P (n_{kt}  \mid  Z_t, \phi_{kt} ),  \label{condit}
\end{align}
where 
\begin{align*}
 p(n_{kt}  \mid  Z_t, \phi_{kt} ) 
& =  \prod_{i=1}^{N_t} p(n^i_{kt}  \mid  Z_{ikt}, \phi_{kt})  =  \prod_{i=1}^{N_t} \text{NB}(n^i_{kt}; Z_{ikt} \phi_{kt}, 1/2).     
\end{align*}
Note that $p(\phi_{kt}  \mid  \gamma)$ is distributed according to its prior Gamma$(\gamma,1)$ and  $\gamma$ according to a chosen hyper-prior. The result follows immediately by plugging these three distributions into \eqref{condit}.
\paragraph{Full conditional of $Z_{ikt}$}
Recall that $n_{ikt}$ denotes the total number of words assigned to topic $k$ in document $i$ at time $t$. If $n_{ikt} > 0$, then the corresponding entry $Z_{ikt}$ is active with probability 1. If $n_{ikt} = 0$, we have
\begin{align*}
& p(Z_{ikt}=1  \mid  Z_{-(ik)t},n_{ikt}=0, X_k(t), \phi_{kt},S_t) =  \\
& \frac{p(Z_{ikt}=1, Z_{-(ik)t}, n_{ikt}=0, X_k(t), \phi_{kt},S_t)}{p(Z_{-(ik)t},n_{ikt}=0, X_k(t), \phi_{kt},S_t)} .
\end{align*}
The numerator is equal to
\begin{equation*}
\begin{multlined}
 p(n_{ikt}=0  \mid  Z_{ikt}=1, \phi_{kt})  p(S_t  \mid  Z_{ikt}=1, Z_{-(ik)t})  p(Z_{ikt}=1  \mid  X_k(t))  p(\phi_{kt},X_k(t), Z_{-(ik)t})  = \\ 
\text{NB}(0;\phi_{kt}, 1/2)  \frac{1}{x^*_1(t)} x_k(t)    p(\phi_{kt},X_k(t), Z_{-(ik)t})
\end{multlined} 
\end{equation*}
Denoting by $C$ the product of all terms not depending on $z_{ikt}$, we have
\begin{align} \label{previous1}
p(Z_{ikt}=1  \mid  Z_{-(ik)t},n_{ikt}=0, X_k(t), \phi_{kt},S_t) = C \frac{1}{2^{\phi_{kt}}}  \frac{1}{x^*_1(t)}  x_k(t).
\end{align}
By the same token, we have
\begin{align} \label{previous2}
p(Z_{ikt}=0  \mid Z_{-(ik)t},n_{ikt}=0, X_k(t), \phi_{kt},S_t) = C  \frac{1}{x^*_0(t)}(1- x_k(t)).
\end{align}
As the two probabilities must sum to 1, we have that
\begin{align*}
C = \frac{2^{\phi_{kt}}x^*_1(t)x^*_0(t)}{x^*_0(t)x_k(t) + 2^{\phi_{kt}}x^*_1(t)(1- x_k(t))},
\end{align*}
which, plugged into equations \ref{previous1} and \ref{previous2}, gives the result.
\vskip 0.2in

\newpage

\small
\bibliographystyle{unsrt}
\bibliography{ibp.bib}

\end{document}